\documentclass[11pt]{article}

\usepackage[margin=1in]{geometry}
\usepackage[utf8]{inputenc}
\usepackage[T1]{fontenc}
\usepackage{lmodern}
\usepackage[dvipsnames]{xcolor}     
\usepackage{pifont}   
\usepackage{multirow}
\usepackage{booktabs} 
\usepackage{siunitx}  
\usepackage{fancyvrb}
\usepackage{xcolor}
\definecolor{tiiPurple}{RGB}{122, 0, 255}

\usepackage{hyperref}
\usepackage{subcaption} 
\hypersetup{
    colorlinks=true,
    linkcolor=tiiPurple,
    urlcolor=tiiPurple,
    citecolor=tiiPurple
}

\newcommand{\purplecheck}{{\textcolor{tiiPurple}{\checkmark}}}
\newcommand{\purplecross}{{\textcolor{tiiPurple}{\ding{55}}}}
\usepackage[most]{tcolorbox}
\usepackage{graphicx}
\usepackage{tabularx}
\usepackage[useregional]{datetime2}
\DTMsetdatestyle{iso}  
\usepackage[normalem]{ulem}
\usepackage{titlesec}
\usepackage{adjustbox}
\usepackage{enumitem}
\usepackage{booktabs}
\usepackage{multirow}
\usepackage{array}
\usepackage{float}
\usepackage{caption}
\usepackage{amsfonts}       
\usepackage{amsmath, amssymb}       
\usepackage{nicefrac}       
\usepackage{hyperref}       
\usepackage{url}            
\usepackage[utf8]{inputenc} 
\usepackage[T1]{fontenc}    
\usepackage{natbib}
\usepackage{graphicx}
\usepackage{enumitem}
\usepackage{makecell}
\definecolor{bestcolor}{RGB}{220,255,220}
\usepackage{CJKutf8}
\usepackage[para]{threeparttable}

\titleformat{\section}
  {\normalfont\Large\bfseries}{\thesection.}{1em}{}
\usepackage{fancyhdr}
\begin{document}
\noindent
\begin{minipage}[t]{0.49\textwidth}
    \vspace*{-4.2em}
    \includegraphics[height=1.3cm,width=2.5cm]{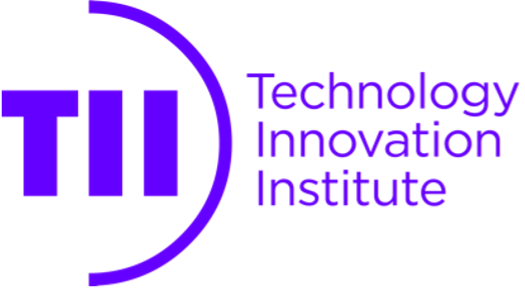}  
\end{minipage}%
\hfill
\begin{minipage}[t]{0.49\textwidth}
    \vspace*{-1.5em}
    \raggedleft
    \today
\end{minipage}
\vspace*{-0.5em}
\hrule
\vspace{1.2em}

\begin{center}
    {\Large \textbf{\textcolor{tiiPurple}{Falcon-H1: A Family of Hybrid-Head Language Models Redefining Efficiency and Performance}}}
\end{center}

\noindent

\begin{center}

\textbf{Falcon LLM Team} \\
\vspace{1em}

\begin{tabular}{@{}l l@{}}
    \raisebox{-0.15\height}{\includegraphics[width=0.4cm]{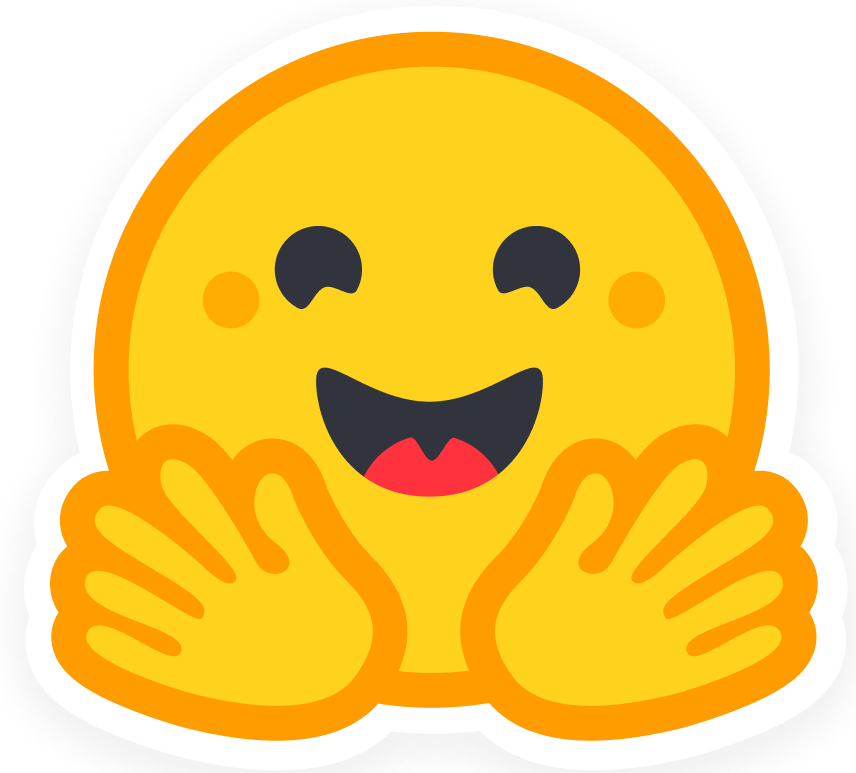}} &
    \href{https://huggingface.co/tiiuae}{https://huggingface.co/tiiuae} \\
    \raisebox{-0.15\height}{\includegraphics[width=0.4cm]{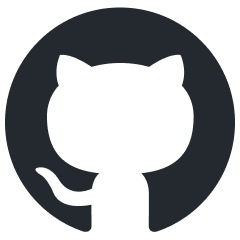}} &
    \href{https://github.com/tiiuae/falcon-h1}{https://github.com/tiiuae/falcon-h1}
\end{tabular}

\end{center}


\vspace{1.5em}
\noindent\textbf{Abstract:}
In this report, we introduce \text{Falcon-H1}, a new series of large language models (LLMs) featuring novel hybrid architecture designs that are optimized for both high performance and efficiency across a broad spectrum of use cases. Unlike previous Falcon models, which were built solely on either Transformer or Mamba architectures, the \text{Falcon-H1} series is based on a parallel hybrid architecture that combines the strengths of the Transformer-based attention mechanism with State Space Models (SSMs), known for their superior long-context memory and computational efficiency. We also systematically revisited nearly every aspect of model design, data strategy, and training dynamics—challenging several conventional practices in the domain. 
To support a wide range of deployment scenarios, the \text{Falcon-H1} series is released in a rich set of configurations, including both base and instruction-tuned models at 0.5B, 1.5B, 1.5B-deep, 3B, 7B, and 34B parameter scales. Quantized versions of the instruction-tuned models are also available. In total, over 30 model checkpoints can be accessed via Hugging Face Hub.

Our comprehensive evaluations demonstrate that Falcon-H1 models consistently set new performance benchmarks through exceptional parameter and training efficiency. The flagship Falcon-H1-34B-Instruct rivals or outperforms leading models up to the 70B scale, such as Qwen3-32B, Qwen2.5-72B and Llama3.3-70B, despite being approximately half the size and trained on a fraction of the data. This parameter efficiency is even more pronounced at smaller scales, where our 1.5B-Deep model achieves performance competitive with state-of-the-art 7B–10B models, Falcon-H1-0.5B delivers performance on par with typical 7B models from 2024. These models demonstrate leadership across a wide range of tasks, including reasoning, mathematics, multilingual, instruction-following, and scientific knowledge. Combined with support for extended context windows of up to 256K tokens and multilingual coverage across 18 languages, Falcon-H1 models are well-suited for a wide array of applications.
All Falcon-H1 models are released under a permissive open-source license\footnote{\url{https://falconllm.tii.ae/falcon-terms-and-conditions.html}}, reinforcing our commitment to accessible, high-impact AI research and development.

\vspace{1.5em}

\tableofcontents

\section{Introduction}
The rapid progress in the development of large foundation models—particularly large language models (LLMs)—has been driven by advances in architectural design, scaling strategies, and training paradigms. Beginning with Transformer-based architectures~\citep{vaswani2017attention} and expanding through massive-scale pretraining and techniques such as supervised fine-tuning (SFT) and reinforcement learning from human feedback (RLHF)\citep{ouyang2022training}.

A key limitation of the vanilla Transformer lies in its quadratic complexity with respect to input sequence length. To address this, recent research has explored more efficient alternatives to traditional attention mechanisms, such as Multi-head Latent Attention (MLA), featured in the DeepSeek series~\citep{liu2024deepseekv2,liu2024deepseek}. In parallel, several novel architectures have been proposed to move beyond Transformers entirely, including Griffin~\citep{de2024griffin}, RWKV~\citep{peng2023rwkv}, Titans~\citep{behrouz2024titans}, and Mamba~\citep{gu2023mamba}, which offer comparable or superior performance in certain tasks with greater computational or memory efficiency.
These advances have given rise to a new class of hybrid models that combine attention mechanisms with state-space models (SSMs), taking advantage of their complementary strengths: attention excels at modeling long-range dependencies, while SSMs provide efficient sequence mixing. This hybrid paradigm has gained popularity in models such as Jamba~\citep{lieber2024jamba,team2024jamba}, Samba~\citep{ren2024samba}, Zamba~\citep{glorioso2024zamba}, and Hymba~\citep{dong2024hymba}.

Building on these insights, we introduce Falcon-H1—an innovative series of large language models that feature a novel parallel hybrid architecture integrating Transformer-style attention with Mamba-based state-space models (SSMs).
Distinct from other leading open-weight LLM series—including LLaMA~\citep{grattafiori2024llama}, Mistral~\citep{jiang2023mistral}, Qwen~\citep{yang2024qwen2,yang2024qwen2_5}, and DeepSeek~\citep{liu2024deepseekv2,liu2024deepseek}, Falcon-H1 is explicitly architected around this hybrid design, harnessing the complementary strengths of both mechanisms to deliver faster inference, lower memory usage, and state-of-the-art performance across a wide array of benchmarks.
The series includes both pre-trained and instruction-tuned variants across seven scales: 0.5B, 1.5B, 1.5B-deep, 3B, 7B, and 34B parameters. In addition to bfloat16-precision models, we also provide quantized versions in multiple precisions to support efficient deployment across diverse hardware environments. Notably, our flagship model, Falcon-H1-34B-Instruct, achieves competitive or superior results compared to the strongest open-weight models to date, such as Qwen3-32B, Qwen2.5-72B-Instruct and LLaMA3.3-70B-Instruct—despite being approximately half the size. 

Below, we show the key features of Falcon-H1: 
\begin{itemize}[leftmargin=*]

    \item \textbf{Innovative Hybrid Architecture}: We combine attention and Mamba-2 heads in parallel within our hybrid mixer block. Importantly, the amount of attention and mamba heads can be adjusted independently, allowing for an optimal attention and SSM ratio. This hybrid design enables faster inference, lower memory usage, and strong generalization across tasks.

    \item \textbf{Wide Range of Model Sizes}: The Falcon-H1 family includes base, instruction-tuned and quantized variants in various sizes—0.5B, 1.5B, 1.5B-deep, 3B, 7B and 34B—designed to meet the needs of diverse usages and deployment scenarios, from edge devices to large-scale systems.
    
    \item \textbf{Multilingual by Design}: Supports 18 languages out of the box, including Arabic (ar), Czech (cs), German (de), English (en), Spanish (es), French (fr), Hindi (hi), Italian (it), Japanese (ja), Korean (ko), Dutch (nl), Polish (pl), Portuguese (pt), Romanian (ro), Russian (ru), Swedish (sv), Urdu (ur), and Chinese (zh) — with scalability to 100+ languages, thanks to our multilingual tokenizer trained on diverse language datasets.

    \item \textbf{Compact Models, Big Performance}: Falcon-H1-0.5B delivers performance on par with typical 7B models from 2024, while Falcon-H1-1.5B-Deep rivals many of the current leading 7B–10B models. Each Falcon-H1 model is designed to match or exceed the performance of models at least twice its size, making them ideal for low-resource and edge deployments without compromising on capability.

    \item \textbf{256K Context Support}: Falcon-H1 models support up to 256K context length, enabling applications in long-document processing, multi-turn dialogue, and long-range reasoning, with exceptional long context performance and greater computational and memory efficiency, Falcon-H1 provides a great balance between performance and resource cost. 

    \item \textbf{Robust Data and Training Strategy}: Falcon-H1 employs a redesigned training approach that maximizes the value of high-quality but limited data. Additionally, the training process scales smoothly across model sizes through a customized \textit{Maximal Update Parametrization} ($\mu P$) recipe, specifically adapted for this novel architecture.
    
\end{itemize}

\section{Architecture}



\begin{figure}[ht] 
    \centering 
    \includegraphics[width=0.7\textwidth]{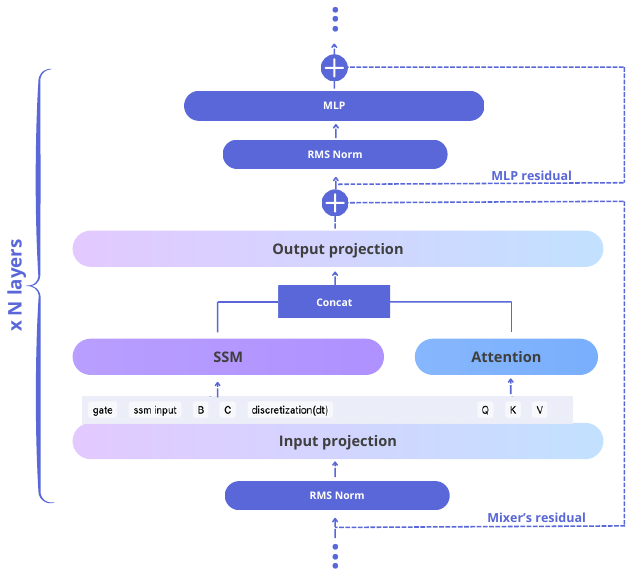} 
    \caption{Falcon-H1 architecture. Attention and SSM run in parallel within each block; their outputs are concatenated before the block’s output projection. The number of SSM/Attention heads can be flexibly tuned.}
    \label{fig:final-arch} 
\end{figure}

Hybrid models combining SSM and attention mechanisms have emerged recently as a promising direction. The classical approach to integrating these components was through \textbf{sequential} designs~\citep{team2024jamba, ren2024samba, glorioso2024zamba}, where one module feeds into the other in series across layers. More recently, \textbf{parallel} designs~\citep{dong2024hymba} have emerged as an alternative integration strategy, where both modules see the same input and their outputs are fused/concatenated before the block projection. For Falcon-H1, we adopt the parallel formulation as shown in Figure~\ref{fig:final-arch}. Our parallel hybrid design has the freedom to choose the ratio of attention and SSM channels, and we are able to keep a small share of attention heads for precision while SSMs handle most of the work.

However, since Mamba architecture is relatively new, their architectural hyper-parameters remain under-explored compared to well-established transformer designs. This necessitates a systematic investigation of design choices to optimize performance. We therefore perform detailed ablations and coarse grid searches on 300M to 1.5B parameter proxy models to understand the impact of key architectural decisions. We sweep critical settings and record both training loss and throughput metrics. These experiments inform the final configuration summarized in Table~\ref{tab:model_architecture} and provide insights for the broader SSM research community.



\begin{table}[!htbp]
\centering
\begin{adjustbox}{max width=\textwidth}
\begin{tabular}{lccccccccc}
\toprule
\textbf{Model} & \textbf{Params (B)} & \textbf{Layers} & \# \textbf{Vocab} & $\boldsymbol{d_{\text{model}}}$ & \textbf{Heads (Q/KV, SSM)} & $\boldsymbol{d_{\text{head}}}$ (Attn/SSM) & $\boldsymbol{d_{\text{state}}}$ & \textbf{Context Len.} & \# \textbf{Tokens} \\
\midrule
\texttt{Falcon-H1-0.5B}        & 0.52 & 36  & 32,778   & 1024  & 8/2, 24     & 64/64     & 128  & 16K   & 2.5T   \\
\texttt{Falcon-H1-1.5B}        & 1.55 & 24  & 65,536   & 2048  & 8/2, 48     & 128/64    & 256  & 128K  & 3T     \\
\texttt{Falcon-H1-1.5B-Deep}   & 1.55 & 66  & 65,536   & 1280  & 6/2, 24     & 128/64    & 256  & 128K  & 3T     \\
\texttt{Falcon-H1-3B}          & 3.15 & 32  & 65,536   & 2560  & 10/2, 32    & 128/128   & 256  & 128K  & 2.5T   \\
\texttt{Falcon-H1-7B}          & 7.59 & 44  & 130,048  & 3072  & 12/2, 24    & 128/128   & 256  & 256K  & \textasciitilde12T \\
\texttt{Falcon-H1-34B}         & 33.6 & 72  & 261,120  & 5120  & 20/4, 32    & 128/128   & 256  & 256K  & \textasciitilde18T \\
\bottomrule
\end{tabular}
\end{adjustbox}
\caption{Model architecture details of the Falcon-H1 series. Embedding and projection layers are untied for all the models.}
\label{tab:model_architecture}
\end{table}

\vspace{0.5em}
\subsection{Channel Allocation}\label{sec:ablate_channels}
A key aspect of our hybrid design is the concatenation of attention and SSM channels, which introduces an additional degree of freedom: the ability to independently vary the number of attention and SSM channels within each hybrid layer. This flexibility has not been explored in previous hybrid designs. For instance,~\citep{dong2024hymba} averages the outputs of attention and SSM channels, which requires both to have identical dimensions. To fully leverage this channel allocation flexibility, we conducted a systematic study of various allocation strategies—including different configurations of MLP channels and the positioning of the MLP block relative to the mixer.

\paragraph{Experimental settings.} To provide a fair comparison across different channel allocation strategies, we adopted the following experimental design. Let $d_\mathrm{ssm},d_\mathrm{attn},d_\mathrm{MLP}$ denote the variable numbers of inner channels for the SSM, attention, and MLP blocks, respectively. The MLP channel dimension $d_\mathrm{MLP}$ can be varied freely without any constraints, as shown in Figure~\ref{fig:final-arch}. For the attention and SSM blocks, we vary the number of query heads while keeping the head dimension, number of key-value (KV) heads, and number of attention groups fixed. Then, we divide the total available channels into 8 chunks, which can be freely allocated across the SSM, attention, and MLP modules. This results in the following parameterization:
\begin{equation}\label{eq:channel_allocations}
d_\mathrm{ssm}=\alpha_S \times 4096, \quad d_\mathrm{attn}=\alpha_A \times 6144, \quad d_\mathrm{MLP}=\alpha_M\times4864, 
\end{equation}
where the chunk fractions $\alpha_S,\alpha_A,\alpha_M$ are chosen from
\begin{equation}\label{eq:SAM_partitions}
    \alpha_S,\alpha_A,\alpha_M\in\{\tfrac{1}{8},\tfrac{2}{8},\tfrac{3}{8},\tfrac{4}{8},\tfrac{5}{8},\tfrac{6}{8}\}, \quad \alpha_S+\alpha_A+\alpha_M=1.
\end{equation}
In this experiment, the base amount of channels per chunk in \eqref{eq:channel_allocations} is set differently for the SSM, attention, and MLP blocks, with a ratio $4096:6144:4864=2:3:2.375$. The reason behind this setting is to balance parameter count and efficiency of different allocations. Recall that the number of matrix layer parameters that scale with inner channels \footnote{For attention and MLP all parameters that scale with inner channels are located in the matrix layers. However, Mamba2 SSM block contains a few vector-like parameters that scale with $d_\mathrm{ssm}$, for example, the additional RMSnorm applied to inner channels after SSM computation. As the number of such extra parameters is tiny compared to the matrix layer ones, we neglect them when balancing parameter counts for different channel allocations.} are given by $3d_\mathrm{ssm}d, \; 2d_\mathrm{attn}d, \; 3d_\mathrm{MLP}d$ for the respective blocks, where $d$ is the hidden dimension of the model. Then, the ratio $2:3:2$ of base channels would keep the number of parameters constant for different blocks. From this fixed parameter ratio, we slightly increased MLP base channels to take into account the lower computational cost of MLP compared to token mixing attention and SSM. This led to the adjusted ratio of $2:3:2.375$, which we instantiate in practice as $4096:6144:4864$ in this experiment.

In addition to channel allocations, another choice we have to make when assembling the hybrid model block is how to position SSM, attention, and MLP blocks with respect to each other. Similarly to parallel and sequential transformer design~\citep{zhao2019museparallelmultiscaleattention,chowdhery2022palmscalinglanguagemodeling,he2024simplifying}, we have the same choice of whether each pair of blocks should be processed in parallel or one after another. We have tested 3 configurations: fully parallel (\texttt{SAM}), semi-parallel  (\texttt{SA\_M}), and fully sequential (\texttt{S\_A\_M}), with the respective forward passes of a single model block $l$ 
\begin{align}
    \label{eq:SAM_configuration}
    \texttt{SAM:}& \quad \mathbf{r}_{l+1}=\mathbf{r}_l + \mathcal{F}_l^\mathrm{MLP}(\mathcal{N}_l(\mathbf{r}_l)) +\mathcal{F}_l^\mathrm{attn}(\mathcal{N}_l(\mathbf{r}_l))+ \mathcal{F}_l^\mathrm{SSM}(\mathcal{N}_l(\mathbf{r}_l))\\ 
    \label{eq:SA_M_configuration}
    \texttt{SA\_M:}& \quad \mathbf{r}_{l+1}=\mathbf{r}_l' + \mathcal{F}_l^\mathrm{MLP}(\mathcal{N}_l'(\mathbf{r}_l')), \quad \mathbf{r}'_l = \mathbf{r}_l + \mathcal{F}_l^\mathrm{attn}(\mathcal{N}_l(\mathbf{r}_l))+ \mathcal{F}_l^\mathrm{SSM}(\mathcal{N}_l(\mathbf{r}_l))\\ 
    \label{eq:S_A_M_configuration}
    \texttt{S\_A\_M:}& \quad \mathbf{r}_{l+1}=\mathbf{r}_l'' + \mathcal{F}_l^\mathrm{MLP}(\mathcal{N}_l''(\mathbf{r}_l'')), \quad \mathbf{r}''_l = \mathbf{r}_l' + \mathcal{F}_l^\mathrm{attn}(\mathcal{N}_l'(\mathbf{r}'_l)), \quad \mathbf{r}'_l=\mathbf{r}_l+ \mathcal{F}_l^\mathrm{SSM}(\mathcal{N}_l(\mathbf{r}_l))
\end{align}
Here $\mathbf{r}_l$ is the residual at the beginning of $l$'th model block, $\mathbf{r}_l',\mathbf{r}_l''$ are intermediate residuals in the middle of sequential model blocks; $\mathcal{F}_l^\mathrm{SSM},\mathcal{F}_l^\mathrm{attn},\mathcal{F}_l^\mathrm{MLP}$ denote SSM, attention and MLP forward passes; and $\mathcal{N}_l,\mathcal{N}_l',\mathcal{N}_l''$ are RMSnorms applied at the beginning of each block and shared for the blocks arranged in parallel to each other. Switching between these 3 block configurations almost does not change the parameter count in the model because SSM/attention/MLP blocks are rearranged as a whole without modifying the insides of each block. The only extra parameters come from the necessity to add new RMSnorm layers $\mathcal{N}_l',\mathcal{N}_l''$ for each additional sequential computation within the model block. 

We have compared all the configurations described above for a relatively deep model with $L=60$ layers, hidden dimension $d=1280$, resulting in approximately $1.2$B parameters. All other training and architecture hyperparameters were identical, and we measured the loss after 70GT of training. 

\begin{figure}[ht]
  \centering
  \includegraphics[width=\linewidth]{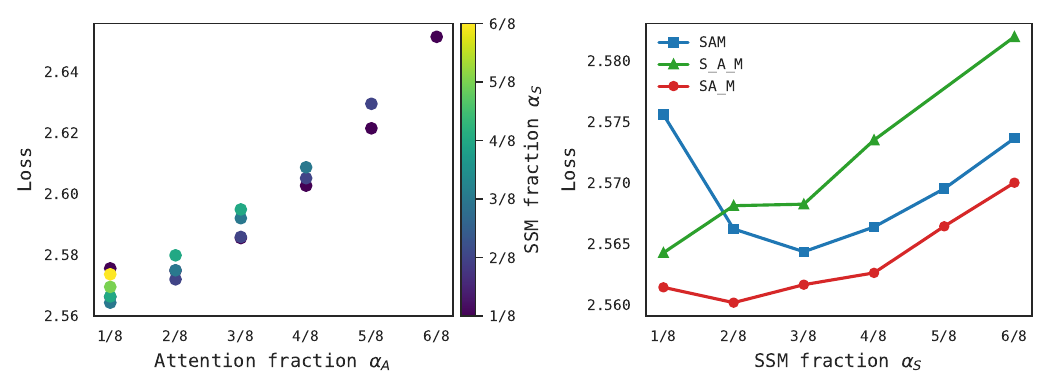}
  \caption{
  \textbf{(Left)}: The loss of fully parallel \texttt{SAM} hybrid block configuration for all possible $(\alpha_S,\alpha_A,\alpha_M)$ channel allocations according to \eqref{eq:channel_allocations},\eqref{eq:SAM_partitions}. 
  \textbf{(Right)} The loss of all 3 considered block configurations \texttt{SAM} \eqref{eq:SAM_configuration}, \texttt{SA\_M} \eqref{eq:SA_M_configuration}, and \texttt{S\_A\_M}\eqref{eq:S_A_M_configuration} for fixed optimal attention allocation $\alpha_A=\frac{1}{8}$ and varied SSM/MLP channel allocation.}
  \label{fig:channel_ablations}
\end{figure}

\paragraph{The results.} Our channel allocation experiments have two parts, focusing on channel allocations and then on block arrangement.  

First, for fully parallel \texttt{SAM} blocks arrangement we examined all 21 admissible $(\alpha_S,\alpha_A,\alpha_M)$ partitions, according to \eqref{eq:SAM_partitions}. The resulting loss values are plotted on figure \ref{fig:channel_ablations} (left). We see a clear separation of magnitude between the impact of the number of attention channels and SSM $\leftrightarrow$ MLP channel switching. Having more attention channels significantly degrades the performance, while SSM $\leftrightarrow$ MLP channel switching has a noticeable but much weaker effect. We have also confirmed similar behavior for the \texttt{SA\_M} block arrangement on a smaller number of runs. 

For the second part of the experiment, we compare all three \texttt{SAM}, \texttt{SA\_M}, \texttt{S\_A\_M} block arrangements while fixing attention channels to the minimum $\alpha_A=\frac{1}{8}$ and varying SSM/MLP channels so that $\alpha_S+\alpha_M=\frac{7}{8}$. The resulting loss values are plotted on figure \ref{fig:channel_ablations} (right). We see that semi-parallel \texttt{SA\_M} configuration provides the best results, $(\alpha_S,\alpha_A,\alpha_M)=(\tfrac{2}{8},\frac{1}{8}, \frac{5}{8})$ being the optimal channel allocation. However, the dependence on SSM/MLP allocations is flat near optimum. Interestingly, as block configuration becomes more sequential $\texttt{SAM}\to\texttt{SA\_M}\to\texttt{S\_A\_M}$, the optimal SSM fraction reduces as $\frac{3}{8}\to\frac{2}{8}\to\frac{1}{8}$. At the moment, we don't have an explanation of this behavior.

Based on the experiment results above, for Falcon-H1 models, we adopted \texttt{SA\_M} block configuration with channel allocations roughly following $2:1:5$ ratio, with slight deviation for different model sizes, which is possible thanks to flat dependence on SSM/MLP allocations near optimum.  

\subsection{SSM-Specific Parameters Ablations}
\label{sec:ssm_params}
We start by revisiting the Mamba2 design~\citep{dao2024transformers} of SSM block we employ for Falcon-H1 models. At the core of the block is a token mixing mechanism that maps an input sequence $x\in \mathbb{R}^{L}$ of length $L$ to $y=\operatorname{SSM}(x,\mathbf{B},\mathbf{C},dt| A_\mathrm{log}, D)\in \mathbb{R}^{L}$ via a recurrent mechanism involving hidden state $\mathbf{h}\in \mathbb{R}^{d_\mathrm{state}\times L}$ 
\begin{equation}
\label{eq:siso-ss}
\mathbf{h}_{t+1} =\overline{A}_t\,\mathbf{h}_t+\mathbf{B}_{t} dt_t x_t, 
\qquad
y_t=\mathbf{C}_{t}^\top\,\mathbf{h}_t+D x_t.
\end{equation}
Here $\mathbf{B}_t$ determines the vector $\mathbf{B}_t dt_t x_t$ to be written into the SSM hidden state, $\overline{A}_t\in\mathbb{R}$ determines the fraction of previous hidden state to be forgotten, $\mathbf{C}_t$ is a reading projector, and the scalar $D\in\mathbb{R}$ controls the direct connection between input and output sequences, by passing the hidden state recurrent computation. The ``time step'' $dt_t\in\mathbb{R}$ controls both the writing intensity via $\mathbf{B}_t dt_t x_t$, and forgetting intensity via the parametrization of $\overline{A}_t=\exp\!\bigl[-e^{A_{\log}}\,dt_t\bigr]$. 

It is often convenient to view SSM operation \eqref{eq:siso-ss} as implementing an linear sequence transformation with a casual ``attention'' matrix $M_{ts}$
\begin{equation}
\label{eq:attention-form}
y_t
   = \sum_{s \le t} M_{ts}\,x_{s},
\qquad
M_{ts}
   = \mathbf{C}_{t}^{\!\top}\,\mathbf{B}_{s}\,dt_{s}
     \prod_{i=s+1}^{t} \overline{A}_i
     + D\,\delta_{ts},
\qquad
\prod_{i=t+1}^{t} := 1.
\end{equation}

The recurrence \eqref{eq:siso-ss} describes the sequence transformation of a single SSM channel. Mamba2 organizes all the $d_\mathrm{ssm}=d_\mathrm{head}n_h$ channels first into $n_h$ heads of dimension $d_\mathrm{head}$, and then further unites the heads into $n_g$ groups with some of the parameters shared within each group. Such organization into heads and groups is similar to grouped query attention (GQA)~\citep{ainslie2023gqa}. Specifically, all the parameters $\mathbf{B}_t,\mathbf{C}_t,dt_t,\overline{A}_t,D$ are broadcasted along the head dimension, and $\mathbf{B}_t,\mathbf{C}_t$ are further shared within each group. 

The key distinction of Mamba2 block compared to early SSM designs is that parameters defining recursion \eqref{eq:siso-ss} are input dependent, allowing for much stronger expressivity. First, the input to the block $\mathbf{u}\in\mathbb{R}^{d\times L}$ goes through a linear transformation, then selectively through a causal depthwise $1$D convolution $\mathrm{conv1d}(\cdot)$, and finally through the SiLU activation function. Denoting concatenation with merged letters, we write the full input transformation as
\begin{align}
\widetilde{\mathbf{x}}\widetilde{\mathbf{z}}\widetilde{\mathbf{B}}\widetilde{\mathbf{C}}\widetilde{\mathbf{dt}} = W_{xzBCdt} \mathbf{u}, \qquad  W_{xzBCdt}\in\mathbb{R}^{(2d_\mathrm{ssm}+2n_gd_\mathrm{state}+n_h)\times d},\\
    \mathbf{x}\mathbf{B}\mathbf{C} = \operatorname{SiLU}\big(\mathrm{conv1d}\big(\widetilde{\mathbf{x}}\widetilde{\mathbf{B}}\widetilde{\mathbf{C}}\big)\big), \quad \mathbf{dt}=\operatorname{Softplus}(\widetilde{\mathbf{dt}}+\mathbf{b}), \quad \mathbf{z}=\operatorname{SiLU}(\widetilde{\mathbf{z}}).
\end{align}
Here, $\mathbf{b}\in\mathbb{R}^{n_h}$ is a head-wise bias, and $\mathbf{z}$ is the gate multiplied element-wise with SSM output $\mathbf{y}_g=\mathbf{y}\odot\mathbf{z}$, similarly to the gated MLP. Finally, $\mathbf{y}_g$ goes into grouped RMSnorm\footnote{grouped RMS  normalization layer is required here to enable the usage of tensor parallelism (TP), which splits SSM channels across different devices.}, followed by the output projection layers $W_o\in\mathbb{R}^{d\times d_\mathrm{ssm}}$ to produce the output of Mamba2 block. We note that parameters $A_\mathrm{log},D,\mathbf{b}$ are, however, static learnable weights that are not input dependent.   

In the remaining paragraph, we describe ablations for various dimensions of the Mamba2 block described above: head dimension $d_\mathrm{head} = d_\mathrm{ssm}/n_h$, the number of groups $n_g$, the recurrent state dimension $d_\mathrm{state}$, the depthwise 1-D convolution kernel size, and the scan chunk size used by SSD algorithm~\citep{dao2024transformers} implementing the recursion \eqref{eq:siso-ss}. 


\paragraph{State dimension \emph{vs.} group count.}
Prior work shows that enlarging the SSM state size \(d_{\text{state}}\) consistently boosts accuracy, but at a non‑trivial efficiency cost~\citep{gu2023mamba,Liu2024VMamba,Stan2024BinaryS4D,Mitra2025CharSSM}. Systematic studies of the accuracy–efficiency frontier remain sparse.

We therefore ran a two–dimensional grid search over the state dimension \(d_{\text{state}}\) and the number of groups \(n_g\).  
To disentangle their effects from model size, we can fix the total number of parameters by fixing budget  
\(B = d_{\text{state}}\!\times\!n_g\) while varying \((d_{\text{state}}, n_g)\). We swept five budgets  
\(B\in\{4,\,16,\,64,\,256,\,1024\}\). 

\begin{figure}[ht]               
  \centering
  \begin{subfigure}[b]{0.48\linewidth}
    \centering
    \includegraphics[width=\linewidth]{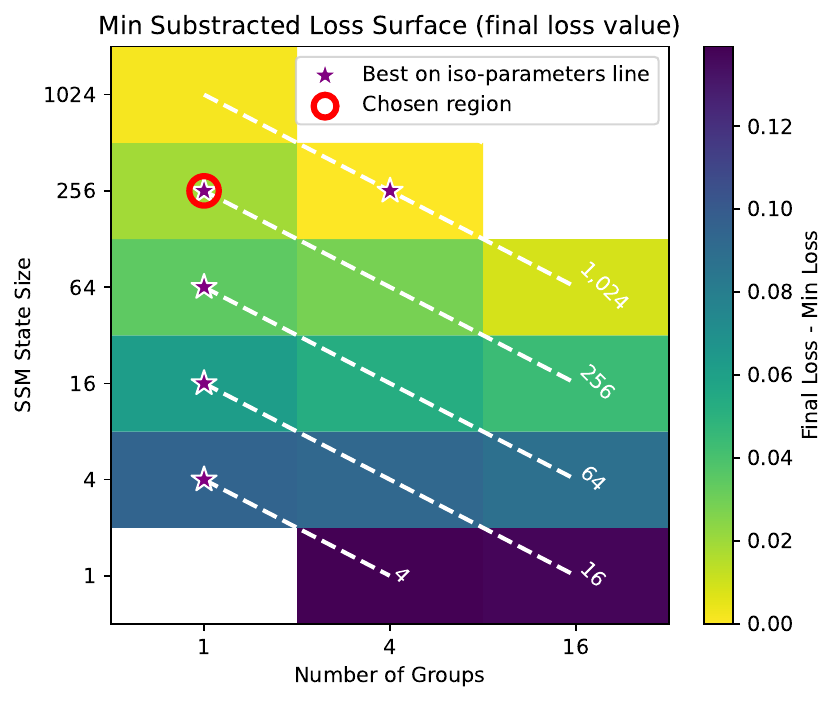}
    \caption{}
  \end{subfigure}
  \hfill
  \begin{subfigure}[b]{0.48\linewidth}
    \centering
    \includegraphics[width=\linewidth]{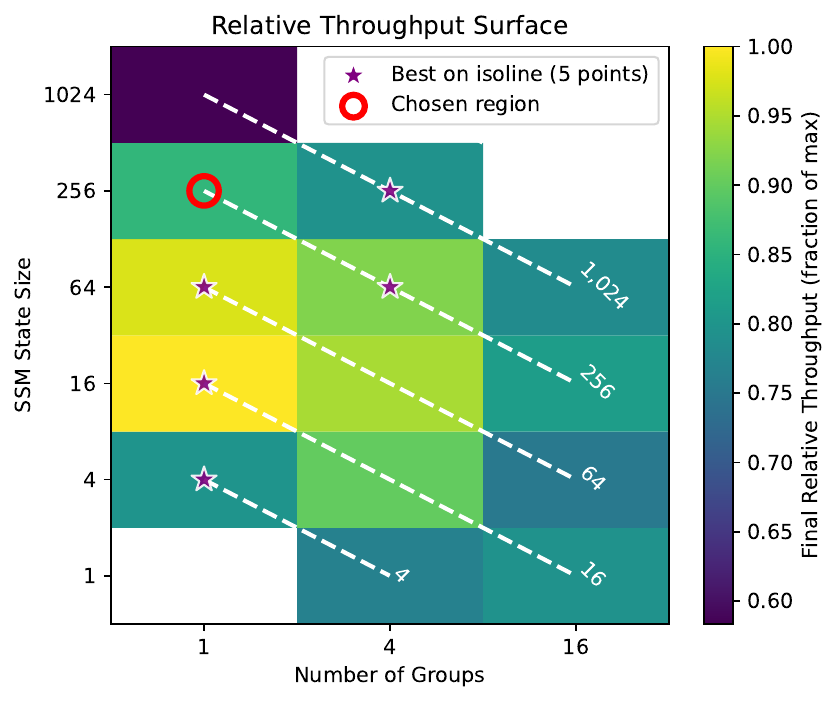}
    \caption{}
  \end{subfigure}

  \caption{Hyperparameter optimization landscapes for SSM number of groups and state dimension size. \textbf{(a)} Loss surface showing performance relative to global minimum across number of groups and d\_state size. \textbf{(b)} Relative throughput surface as fraction of maximum performance. Dashed lines indicate iso-parameter curves (ng $\times$ ds = constant), implying constant total parameter count. Red stars mark optimal configurations for each computational budget, revealing distinct trade-offs between model quality and efficiency.}
  \label{fig:ng_ds_combined}
\end{figure}

As shown in Fig.\,\ref{fig:ng_ds_combined}, validation accuracy rises almost exclusively with larger \(d_{\text{state}}\); varying \(n_g\) has only a marginal impact. Thus, the best configuration within any budget uses the smallest feasible \(n_g\) and the largest possible \(d_{\text{state}}\).
Conversely, training throughput deteriorates with increasing \(d_{\text{state}}\), with a pronounced efficiency apex around \(d_{\text{state}}=16\).

Note that all experiments were conducted at a sequence length of 2048.  
Because longer sequences require a larger state to retain historical information, we take \((n_g, d_{\text{state}}) = (1,\,256)\) for the final models as the best compromise. Since \textsc{Falcon-H1-34B} was trained with tensor parallelism (TP) = 4 and mixer parallelism (MP; see~\S\ref{subsubsec:mixer_parallelism}), we fixed the number of groups to \(n_g = 2\) so that it is divisible by TP/2.



\paragraph{Head dimension $d_\mathrm{head}$.}  

\begin{figure}[ht]               
  \centering
  \begin{subfigure}[b]{0.48\linewidth}
    \centering
    \includegraphics[width=\linewidth]{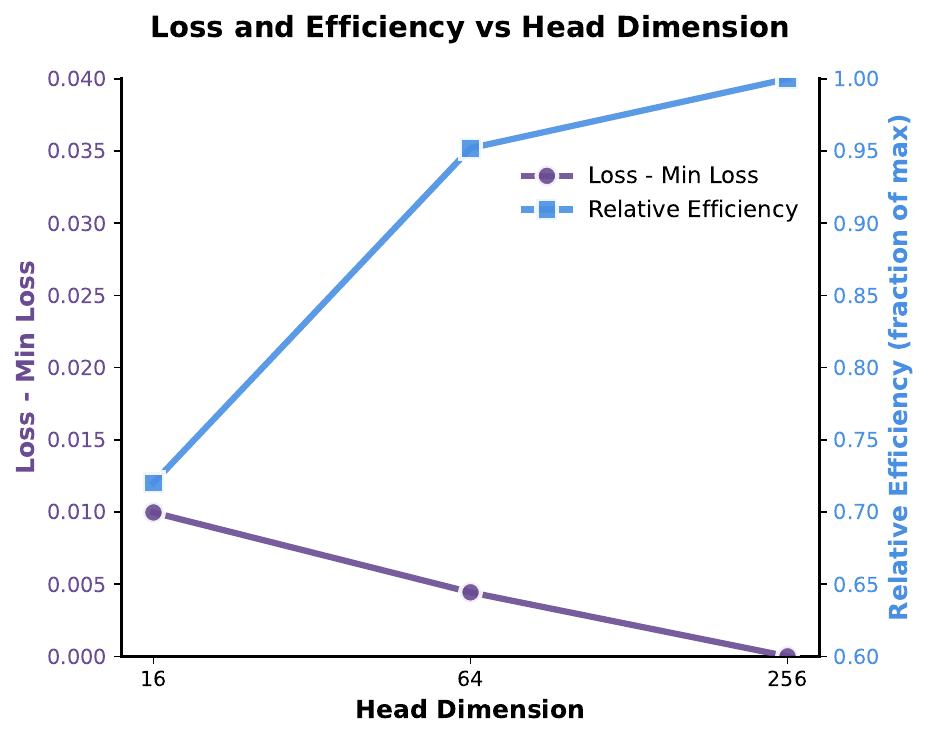}
    \caption{}
    \label{fig:hd}
  \end{subfigure}
  \hfill
  \begin{subfigure}[b]{0.48\linewidth}
    \centering
    \includegraphics[width=\linewidth]{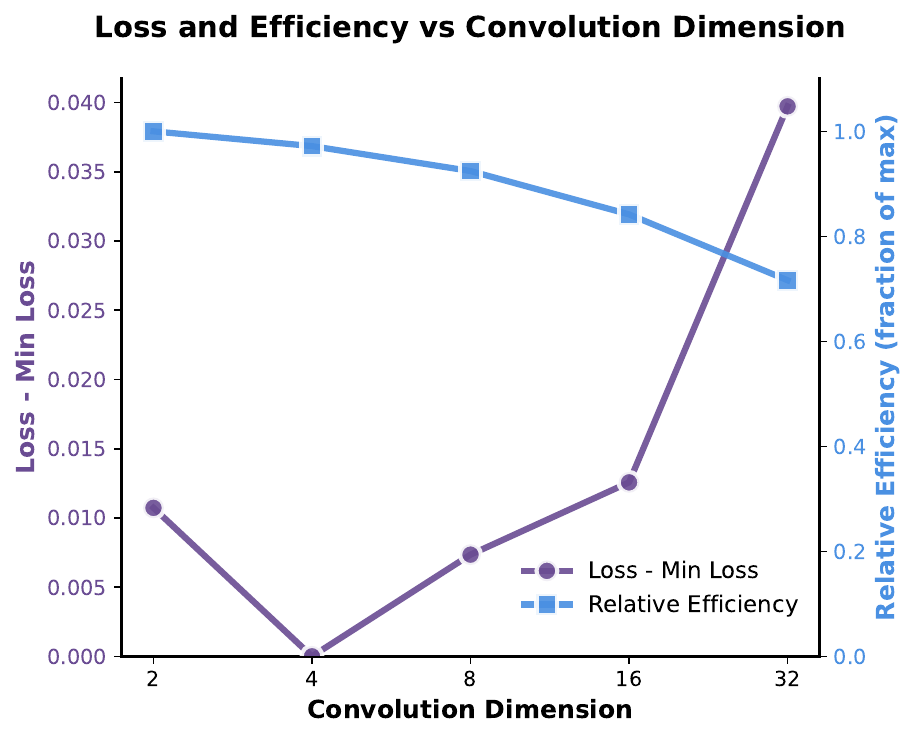}
    \caption{}
    \label{fig:cv}
  \end{subfigure}

  \caption{Model accuracy and computational efficiency across architectural dimensions. (a) Similar analysis for attention head dimensions, showing that larger head dimensions provide both better computational efficiency and lower loss. (b) Loss (purple line, left y-axis) and relative computational efficiency (blue line, right y-axis) as functions of convolution dimension. The analysis reveals an optimal trade-off at dimension 4, where the model achieves minimal loss while maintaining high efficiency. }
  \label{fig:hd_cv_combined}
\end{figure}

To isolate the effect of the SSM head size \(d_{\text{head}}\), we trained variants with
\(d_{\text{head}}\in\{16,64,256\}\) while keeping \(d_{\mathrm{ssm}}=d_{\text{head}}n_h\) so the parameter count stays roughly constant.
We observe a \(\le10^{-2}\) change in training cross-entropy, with a clear gain at larger heads (see Fig.\,\ref{fig:hd}).
Throughput is more sensitive: \(d_{\text{head}}<32\) reduced GPU utilization, whereas \(d_{\text{head}}\ge64\) maintained optimal efficiency.
To our knowledge, no prior work reports an explicit ablation of \(d_{\text{head}}\) for SSMs; existing sources simply pick values (e.g., 64 or 128) and discuss kernel limits~\citep{Dao2024Mamba2Part1,dao2023flashattention2}.
  
Larger head dimensions therefore yield a more favorable accuracy and efficiency.

\paragraph{Depthwise causal 1-D convolution (\texttt{convdim}).}
To our knowledge, no prior work reports an ablation over the depthwise causal Conv1d kernel size inside Mamba-style SSM blocks; existing papers simply fix \(k=4\)~\citep{Chen2025TensorSSM,SEMamba2024}. The reference \texttt{causal\_conv1d} CUDA kernel itself only supports \(\{2,3,4\}\)~\citep{DaoAILabCausalConv1d,GitHubMambaKernels523}. 
We therefore re-implemented the kernel to handle sizes up to \(32\) and swept \(\{2,4,8,16,32\}\). Kernel size \(4\) indeed minimized validation loss, whereas both smaller and larger filters degraded accuracy, so we keep \(\text{kernel\_size}=4\) in the final models (see Fig.\,\ref{fig:cv}).

\paragraph{Chunk size (\texttt{cs}).}
The SSD kernel processes a long sequence in blocks of \texttt{cs} tokens.  
The efficiency scales favorably with \texttt{cs} until two limits emerge:

\begin{enumerate}
    \item \emph{Launch overhead.}  Very small chunks (\(\texttt{cs}<64\)) trigger
          many kernel launches and under-utilise the GPU.
    \item \emph{Memory pressure.}  When \(\texttt{cs}>256\) the cross-chunk
          prefix-sum kernel (\texttt{dA\_cumsum}) no longer fits in
          on-chip SRAM and becomes memory-bound, lowering the efficiency.
\end{enumerate}

A broad plateau therefore appears at
\(\texttt{cs}\in\{128,\,256\}\).  This matches the implementation guidance in the
Mamba official codebase where the Triton kernels are tuned for power-of-two
values and default to \(\texttt{cs}=256\).  We fix
\(\texttt{cs}=256\) for all subsequent experiments to maximize throughput while
retaining numerical stability.





\paragraph{Hidden State Resetting in \textsc{Mamba}.} 
When several documents are concatenated inside one long sequence,
the tail of document \(k\) (\(0\!\le s<T_k\)) leaks into
the head of document \(k{+}1\) through the product of \(\bar A\)
in \eqref{eq:attention-form}.  
This \emph{cross-doc leakage} violates the
independence assumption of language-model training and especially causes semantic contamination between unrelated contexts. For attention, a simple fix is a block‑diagonal (segmented) mask that zeroes attention scores across document boundaries—often called \emph{cross-document masking}~\citep{grattafiori2024llama, team2025gemma3}.

For recurrent architectures, such cross-document bleeding can be avoided by resetting the hidden state at document boundaries.
Let \(r_t = 1\) signify that token \(t\) is the first token of a new document in the sequence.  
This binary indicator is derived from the data loader’s position tensor and marks document boundaries within a packed sequence.

To reset hidden-state on document boundary we need to force $\bar A_t = 0$ when $t$ is at the document edge, we do so by injecting a large negative value \(-80\) \emph{channel-wise} into the SSM parameter vector $\bar A$ before exponentiation at positions where \(r_t = 1\):

\[
\bar A_i
= \exp\!\bigl[-e^{A_{\log}}\,\tilde{dt}_i + r_i \cdot (-80)\bigr]
\approx
\begin{cases}
\mathbf{0}, & r_i = 1, \\
\bar A, & r_i = 0.
\end{cases}
\]

Thus, at a boundary,  
\(
\mathbf{h}_{t+1} = \mathbf{0} \cdot \mathbf{h}_t + \bar B\,x_t = \bar B\,x_t,
\)  
perfectly \emph{resetting} the hidden state in a single step with no additional compute.  
Subsequent tokens use the standard transition weights \(\bar A\).
This resetting scheme does not add any compute or memory overhead while remaining gradient-safe because the \(-80\) bias is constant, so gradients propagate normally through \(\exp\) and \(\widetilde{\bar A}_t\). Moreover, it is numerically stable: \(\exp(-80)\!\approx\!10^{-35}\) lies above the FP16/BF16 underflow threshold (\(\sim10^{-45}\)) yet empirically zeros the hidden state without training instabilities.

\vspace{0.5em}

\subsection{Challenging Conventional Components}
\subsubsection{RoPE Base Frequency}  

For Falcon-H1 models, we have used an unconventionally high value $b=10^{11}$ of the RoPE base frequency. Below, we present our reasoning process to arrive at this value. 

We have started the training of 7B and 34B models with a standard value $b=10^4$ and sequence length $L_\mathrm{seq}=8192$. In the middle of the training, we increased sequence length to $L_\mathrm{seq}=16384$ but observed a drop in the model evaluations and revoked the change. This drop was unexpected since we did not change the data mixture, and the only effect of $L_\mathrm{seq}$ increase amounts to fewer training samples being cut. Suspecting RoPE base frequency as the main parameter that interacts with the training sequence length, we have increased the base frequency to $b=10^{6}$. This change resulted in an immediate boost in model evaluations, and another boost after increasing sequence length again to $L_\mathrm{seq}=16384$.    

\begin{figure}[ht]               
  \centering
    \begin{subfigure}[b]{0.48\linewidth}
    \centering
    \includegraphics[width=\linewidth]{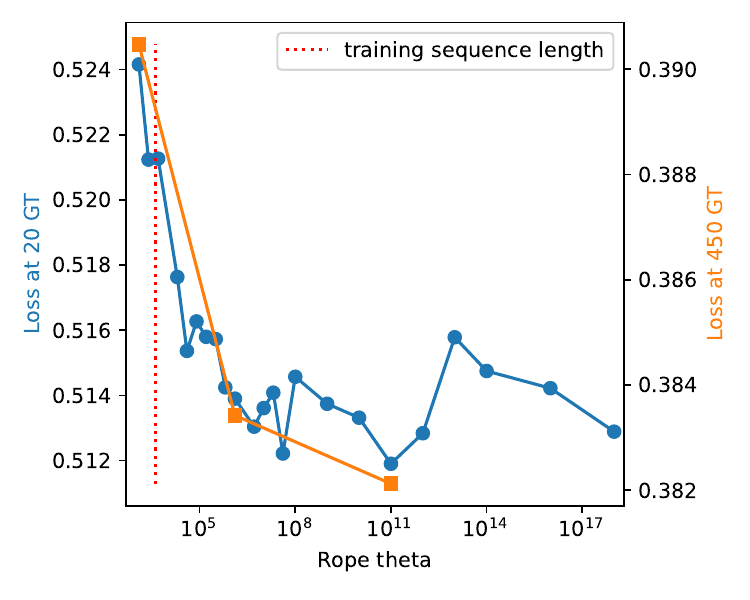}
    \caption{}
    \label{fig:roe}
  \end{subfigure}
  \hfill
  \begin{subfigure}[b]{0.5\linewidth}
    \centering
    \includegraphics[width=\linewidth]{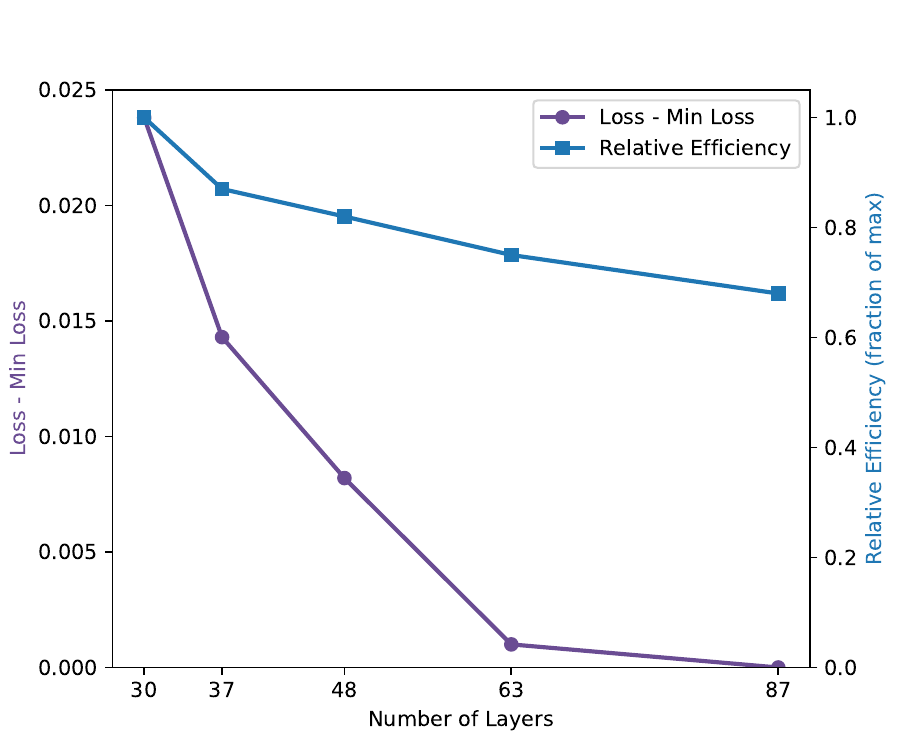}
    \caption{}
    \label{fig:ng_ds_loss}
  \end{subfigure}

  \caption{\textbf{(a)} Dependence of the training loss on RoPE base frequency $b$. The dotted line shows the training sequence length for a reference. First, we tried many base frequencies and measured the loss early in the training at 20GT. Then, we picked 3 characteristic base frequency values and measured the loss in a much later training stage at 450GT, with the results being roughly similar to the early measurement. \textbf{(b)} Dependence of the training loss on the number of layers for a fixed number of parameters.}
  \label{fig:layers_rope_combined}
\end{figure}

To further investigate the impact of base frequency $b$, we ran a $b$ sweep on Falcon-H1 0.5B model, depicted in Figure \,\ref{fig:roe}. At smaller $b$, a few orders of magnitude around the model sequence length, the training loss $L(b)$ steeply depends on $b$, with lower values being extremely suboptimal. Earlier drop of evaluations on 7B/34B models when increasing $L_\mathrm{seq}$ can be interpreted as moving up the $L(b)$ curve due to the decrease of $b$ relatively to $L_\mathrm{seq}$. At larger $b$ the curve $L(b)$ flattens and slowly increases, reflecting the NoPE (no positional embeddings). limit $b\to\infty$. Our chosen value $b=10^{11}$ roughly corresponds to the optimum of $L(b)$ curve. We stress, though, that optimal $b$ does not need to be estimated very accurately due to the flatness of $L(b)$ for large $b$. Applying $b=10^{11}$ to 7B/34B models resulted in another increase of evaluation scores. 

Finally, let us point out the advantage of using a large base frequency $b$ when scaling the sequence either during continual pretraining or inference. RoPE assigns different frequencies $\theta_k = b^{-2k / d_\mathrm{head}}$ to different dimensions $k$ within query and key vectors. When the value of $b$ is comparable to the original sequence length $L_\mathrm{seq}$, further increase of $L_\mathrm{seq}$ requires reassigning frequencies within QK dimensions to allocate some space for smaller $\theta$ (or larger wavelengths $\theta^{-1}$) needed to handle longer sequences. Different strategies of this reassigning, such as Position Interpolation~\citep{chen2023extendingcontextwindowlarge}, ``NTK-aware''~\citep{bloc97a} or ``NTK-by-parts''~\citep{bloc97b}, improve performance on larger sequences but still deform the original assignment $\theta_k$ the model has adapted to during training. 

Fortunately, using extremely large $b$ during training leaves many dimensions $k$ effectively unassigned, since the respective large sequences were never seen during training. In that case, no RoPE modifications are required when increasing sequence length beyond the training value, making sequence length extension for Falcon-H1 models extremely simple.     

An interesting question is whether such large $b$ values are optimal only for hybrid models, where SSM part can take care of short-range dependencies, or can also work for transformer models. 

\subsubsection{Width--Depth Trade-offs}
\label{sec:width_depth_tradeoff}
Increasing the \emph{depth} (number of layers) versus the \emph{width} (hidden dimensionality) of a decoder-only model presents distinct trade-offs in expressivity and efficiency. Depth provides more sequential composition of nonlinear transformations, enabling hierarchical feature learning and multi-step reasoning that a shallow-but-wide model might require exponentially more neurons to emulate~\citep{chen2024theoretical}. In contrast, width expands the model’s capacity per layer, allowing it to encode more features in parallel. Depth thus increases representational power by adding layers of compositionality, while additional width increases the representational richness at each layer without increasing the number of sequential transformations.

From an optimization and efficiency perspective, these choices have significant implications. Deep stacks are more sequential, which hampers parallelism and can slow down training and inference: each added layer introduces an additional serial step that cannot be parallelized in time, leading to higher latency. Wider layers, on the other hand, perform more computation in parallel within a single layer (e.g., larger matrix multiplications), which modern hardware can exploit efficiently. Memory constraints also differ: very deep models must store activations for many layers during backpropagation (increasing memory usage proportional to depth). By contrast, widening a layer adds no additional sequential operations: each layer can encode more information and shortens the longest gradient path, alleviating bottlenecks. The trade-off is a higher peak-memory footprint. From a training point of view, very deep networks demand careful architectural tweaks to remain trainable, whereas very wide networks may hit other limits such as memory bandwidth or diminishing returns in utilization of parameters.

In practice, state-of-the-art LLM architectures balance width and depth to leverage the benefits of both. Modern decoder-only LLMs are typically built with dozens to $\sim$100 layers, each with a very large hidden size, rather than choosing an extreme in one dimension. For example, LLaMA-3~\citep{grattafiori2024llama} scales from 8\,B to 70\,B parameters by increasing both depth and width layers with a 12\,288-dimensional hidden size. This co-scaling strategy ensures the model has sufficient depth to compose complex behaviors and sufficient width to maintain high capacity and parallel throughput. In fact, many design heuristics keep the network’s depth and width growing in tandem with total size.

For the \textsc{Falcon-H1} series, we revisited the width–depth trade-off under a fixed 1.5\,B parameter budget.  
We performed a joint sweep over hidden width $d_{\text{model}}$ and depth $L$, scaling the learning rate inversely with width following a simple $\mu P$ scaling~\citep{yang2022tensorprogramsvtuning} ($\eta \propto 1/d_{\text{model}}$) to stabilize training dynamics across configurations and omitted depth-scaling at this stage for simplicity.

We evaluated five architecture shapes: \texttt{W1536L87}, \texttt{W1792L63}, \texttt{W2048L48}, \texttt{W2304L37}, and \texttt{W2560L30}.  
Efficiency was measured in training giga-tokens per hour (gtok/h), and quality was assessed via pre-training cross-entropy.

As shown in Figure \ref{fig:ng_ds_loss}, greater depth yielded consistently higher overall quality. The 87-layer extreme \texttt{W1536L87} variant clearly outperformed the wider \texttt{W2560L30} one. From our empirical studies, it even matched/outperformed 3.0\,B and 7.0\,B reference models twice to 5 times its size. This accuracy boost came at a cost: training throughput dropped by 25–30 \% gtok/h and inference slowed by a comparable margin relative to the shallowest configuration. Because the depth–width balance remains under-explored, we are releasing two 1.5 B variants—\textsc{Falcon-H1-1.5B} (width-balanced with 24 layers) and \textsc{Falcon-H1-1.5B-Deep} (deeper with 66 layers) to foster further investigation of this trade-off.



\subsection{Tokenizer}
Considering the memory footprint and model performance acorss different model scales~\citep{tao2024scaling}, we decided to train multiple tokenizers with different vocabulary size. Basically, we increase gradually the vocabulary size regarding model size as shown in Table~\ref{tab:model_architecture}. In this section, we show detailed studies and experimental results when building Falcon-H1 tokenizers.

\subsubsection{Empirical Studies}

In the literature, a tokenizer with a high compression rate (i.e., low fertility score) across languages—including non-Latin scripts—is widely regarded as essential. By encoding equivalent text with fewer tokens, such tokenizers improve both training and inference efficiency, requiring fewer iterations to generate the same amount of text. However, compression alone does not fully capture the factors influencing end-to-end LLM performance.

To address this, we conducted a series of experiments to investigate how various tokenizer training strategies affect both proxy metrics and downstream outcomes. This was primarily evaluated using the fertility score (compression rate) and the average number of bytes per vocabulary token (expressiveness). Beyond these proxy metrics, we also performed full-scale training experiments to directly observe the impact of design choices that might not be evident through compression or expressiveness alone. Specifically, we explored the impact of scaling tokenizer training data, regex splitting patterns, handling of punctuation and digits, and inclusion of \LaTeX\ tokens.

\noindent\textbf{Will scaling training data impact tokenizer performance?}
To investigate how scaling training data  affects tokenizer performance, we conducted an experients with six tokenizers trained on English text. We varied two factors: the training corpus sizes (1GB, 14GB, and 40GB) and the vocabulary size (65k and 135k). All other training parameters were held constant to isolate the impact of these variables. The results are summarized in Table~\ref{tab:tokenizer_performance}.

\begin{table}[h!]
\centering
\sisetup{round-mode=places, round-precision=4} 
\begin{tabular}{
    c 
    c 
    S[table-format=1.4] 
    S[table-format=1.4] 
}
\toprule
\textbf{Vocab Size} & \textbf{Corpus Size} & {\textbf{Fertility Score} \(\downarrow\)} & {\textbf{Bytes per Token} \(\uparrow\)} \\
\midrule
\multirow{3}{*}{65k} 
    & ~1GB  & \textbf{1.4350} & 8.2435 \\
    & ~14GB & 1.4443         & \textbf{8.7274} \\
    & ~40GB & 1.4907         & 8.3748 \\
\cmidrule(lr){2-4}
\multirow{3}{*}{135k} 
    & ~1GB  & 1.3887         & 8.8912 \\
    & ~14GB & \textbf{1.3344} & \textbf{9.2688} \\
    & ~40GB & 1.3964         & 9.2577 \\
\bottomrule
\end{tabular}
\caption{Tokenizer performance by corpus and vocabulary size. Best results for each vocabulary size are in \textbf{bold}. The arrows indicate the desired direction for each metric.}
\label{tab:tokenizer_performance}
\end{table}

The relationship between corpus size and performance is non-monotonic and is conditioned on the vocabulary size. For a 65k vocabulary, optimal performance is achieved with smaller corpora: the 1GB corpus yields the best compression, while the 14GB corpus maximizes bytes per token. Performance degrades at 40GB. Conversely, for a 135k vocabulary, the 14GB corpus surpasses both smaller and larger corpora on both metrics. 

\begin{tcolorbox}[colback=yellow!10!white, colframe=orange!50!black, boxrule=0.8pt]
\textbf{Conclusion:} Simply increasing the training data volume does not guarantee a better tokenizer. Instead, there is an optimal range of data that depends on the vocabulary size.
\end{tcolorbox}

\noindent\textbf{How important is the Splitting RegEx?}
The Splitting RegEx is a pattern rule used to break raw text into preliminary word-like units before applying the actual tokenization algorithm. In other words, it defines how the tokenizer initially segments text, influencing how efficiently it can compress data into tokens. For this analysis, we compared three publicly available splitting regex patterns used by different tokenizers: GPT-4o, LLaMA-3, and GPT-2. We trained all tokenizers with a fixed vocabulary size of 131k on the same dataset, and then measured both the fertility score and the average number of bytes represented in the vocabulary.

\begin{table}[h!]
\centering
\sisetup{round-mode=places, round-precision=4} 
\begin{tabular}{
    c 
    S[table-format=1.4] 
    S[table-format=1.4] 
}
\toprule
\textbf{Splitting RegEx} & {\textbf{Fertility Score} \(\downarrow\)} & {\textbf{Bytes per Token} \(\uparrow\)} \\
\midrule
GPT-2            & 1.3466           & \textbf{9.1019} \\
GPT-4o              & \textbf{1.3209} & 8.7924 \\
LLaMA-3             & 1.3238           & 8.7003 \\
\bottomrule
\end{tabular}
\caption{Impact of different splitting regex patterns on fertility score and average bytes per token. Best scores for each metric are in \textbf{bold}; arrows indicate the desired optimization direction.}
\label{tab:split_regex_effect}
\end{table}

From the results shown in Table~\ref{tab:split_regex_effect}, we observe that while the choice of splitting regex does have a measurable impact on both the fertility score and the average bytes per token, the differences between recent regex patterns such as GPT-4o and LLaMA-3 remain relatively small.

\begin{tcolorbox}[colback=yellow!10!white, colframe=orange!50!black, boxrule=0.8pt]
\textbf{Conclusion:} The differences between splitting regex patterns used by modern tokenizers are minor. For practical purposes, it is advisable to adopt a splitting regex from a well-established and up-to-date tokenizer.
\end{tcolorbox}

\noindent\textbf{Whether to apply Punctuation and Digits Splitting?} Whether to apply punctuation and digit splitting remains an active topic of discussion in the community~\citep{singh2024tokenizationcountsimpacttokenization}, with notable impacts on domains such as mathematics and code. While metrics like fertility score and bytes-per-token are often used to evaluate tokenizers, we find they do not consistently predict downstream model performance. For example, as shown in Table~\ref{tab:digit_split_results}, we trained two tokenizers on identical data—one applying individual digit splitting and one without. Although the latter achieves a better (lower) fertility score, most recent studies~\citep{yang2024qwen2_5,grattafiori2024llama} adopt digit splitting as a standard practice, underscoring that such metrics alone may not fully capture tokenizer effectiveness. 


\begin{table}[ht]
\centering
\begin{tabular}{lccc}
    \toprule
    \textbf{Splitting Setting} & \textbf{Vocab Size} & {\textbf{Fertility Score} \(\downarrow\)} & {\textbf{Bytes per Token} \(\uparrow\)} \\
    \midrule
    Individual digits & 135k & 1.3209 & 8.7924 \\
    No digits splitting & 135k & \textbf{1.2884} & \textbf{8.9688} \\
    \bottomrule
\end{tabular}
\caption{Fertility Score (Compression rate, digits excluded on the test set) and average number of bytes for different splitting regular expressions.}
\label{tab:digit_split_results}
\end{table}

\begin{figure}[ht] 
    \centering 
    \includegraphics[width=0.7\textwidth]{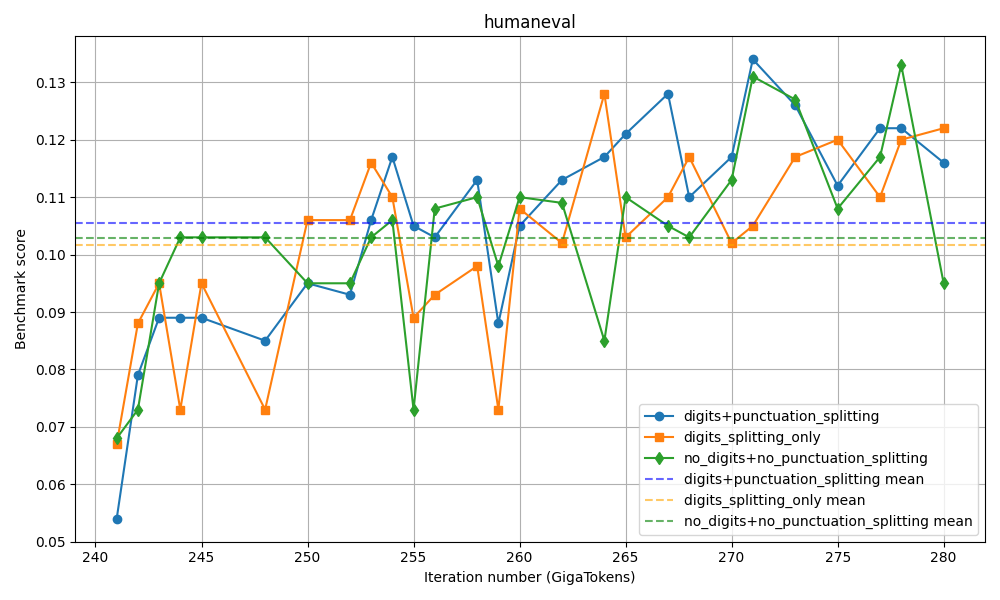}
    \caption{Model performance regarding different splitting strategies.}
    \label{fig:humaneval_results}
\end{figure}

To resolve this ambiguity, we conducted a controlled empirical study focused on downstream task performance. We trained three 1.8B Falcon-Mamba models~\citep{zuo2024falconmamba} on a shared 280GT dataset (including 40GT of decay stage). The models differed only in their tokenizer configurations: \textit{Splits both digits and punctuation}; \textit{Splits digits only}; \textit{No specialized splitting for digits or punctuation}.
As shown in Figure~\ref{fig:humaneval_results}, despite some score fluctuations inherent to training dynamics, the results on the HumanEval code benchmark~\citep{chen2021codex} indicate a clear trend: enabling both punctuation and digit splitting consistently leads to superior code generation performance.
The benefits of punctuation splitting are further supported by a qualitative analysis of tokenization in non-Latin languages. In languages like Chinese and Japanese, punctuation marks are often full-width characters that can be incorrectly merged with adjacent words if not explicitly separated.
Figure~\ref{fig:cjk_splitting_example} illustrates this phenomenon. Without punctuation splitting, the tokenizer merges characters with punctuation, producing semantically incoherent units.

\begin{figure}[ht]
    \begin{CJK*}{UTF8}{gbsn}
    \centering
    \setlength{\fboxsep}{1pt} 
    \small 
    \textbf{Without Punctuation Splitting (Falcon3 tokenizer)} \\
    \textit{Tokens are incorrectly merged with punctuation.} \\
    \vspace{2mm}
    \colorbox{Gray!30}{关闭} \colorbox{Gray!30}{此} \colorbox{Gray!30}{模式} \colorbox{Gray!30}{后} \colorbox{Orchid!30}{，您} \colorbox{Gray!30}{将} \colorbox{Gray!30}{无法} \colorbox{Gray!30}{再} \colorbox{Gray!30}{看到} \colorbox{Gray!30}{它} \colorbox{Orchid!30}{。如果您} \colorbox{Gray!30}{丢失} \colorbox{Gray!30}{了} \colorbox{Gray!30}{它} \colorbox{Orchid!30}{，则} \colorbox{Gray!30}{必须} \colorbox{Gray!30}{创建} \colorbox{Gray!30}{一个新的} \colorbox{Gray!30}{。}

    \vspace{0.7cm}

    \textbf{With Punctuation Splitting (Falcon-H1 tokenizer)} \\
    \small \textit{Punctuation is correctly isolated, preserving word boundaries.} \\
    \vspace{2mm}
    \centering
    \colorbox{Gray!30}{关闭} \colorbox{Gray!30}{此} \colorbox{Gray!30}{模式} \colorbox{Gray!30}{后} \colorbox{Orchid!30}{，} \colorbox{Salmon!30}{您} \colorbox{Gray!30}{将} \colorbox{Gray!30}{无法} \colorbox{Gray!30}{再} \colorbox{Gray!30}{看到} \colorbox{Gray!30}{它} \colorbox{LimeGreen!30}{。} \colorbox{Orchid!30}{如果您} \colorbox{Gray!30}{丢失} \colorbox{Gray!30}{了} \colorbox{Gray!30}{它} \colorbox{Orchid!30}{，} \colorbox{Turquoise!30}{则} \colorbox{Gray!30}{必须} \colorbox{Gray!30}{创建} \colorbox{Gray!30}{一个新的} \colorbox{Gray!30}{。}
    \end{CJK*}
    \caption{Qualitative comparison for a Chinese sentence, with or without punctuation splitting.}
    \label{fig:cjk_splitting_example}
\end{figure}

\begin{tcolorbox}[colback=yellow!10!white, colframe=orange!50!black, boxrule=0.8pt]
\textbf{Conclusion:} Splitting both digits and punctuation seems to be the most effective strategy. This approach enhances performance on code and math tasks and ensures more robust, semantically meaningful tokenization across diverse languages. Moreover, our results underscore that tokenizer design should be guided by the actual model performance, rather than relying solely on proxy metrics such as fertility score.
\end{tcolorbox}

\noindent\textbf{How important are the common \LaTeX\ tokens?}
Given the prevalence of \LaTeX\ syntax in scientific and mathematical documents, we hypothesized that incorporating common \LaTeX\ commands directly into the tokenizer's vocabulary would enhance a model's mathematical reasoning capabilities. The core principle is that representing frequent commands like \texttt{\textbackslash frac} or \texttt{\textbackslash sqrt} as single, atomic tokens simplifies the prediction task for the model, reducing the sequence length and compositional complexity of mathematical expressions.

\begin{figure}[ht]
    \centering
    \begin{subfigure}[b]{0.45\textwidth}
        \centering
        \includegraphics[width=\textwidth]{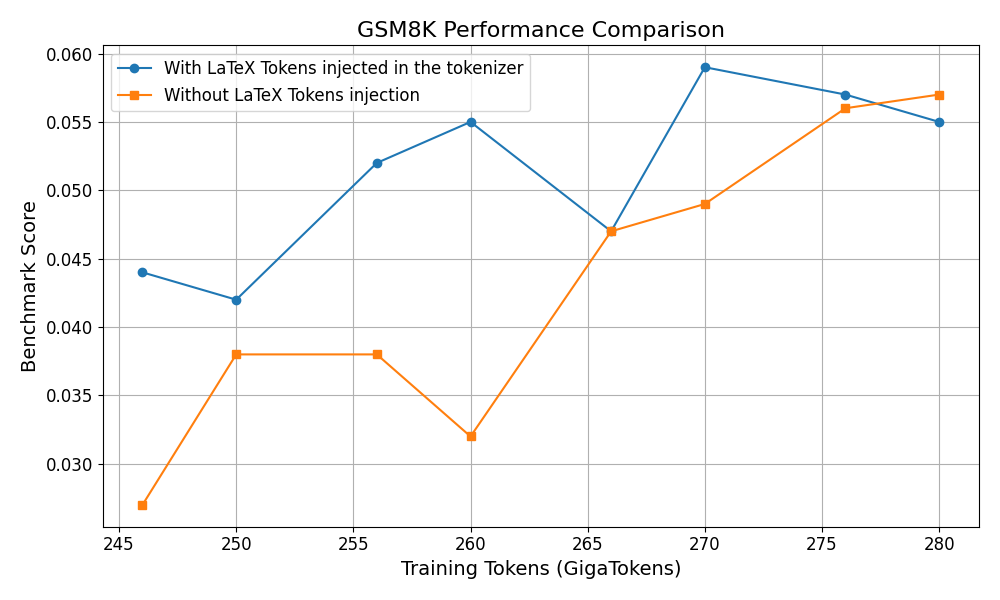}
        \label{fig:tokenizer-gsm8k}
    \end{subfigure}
    \hspace{1em} 
    \begin{subfigure}[b]{0.45\textwidth}
        \centering
        \includegraphics[width=\textwidth]{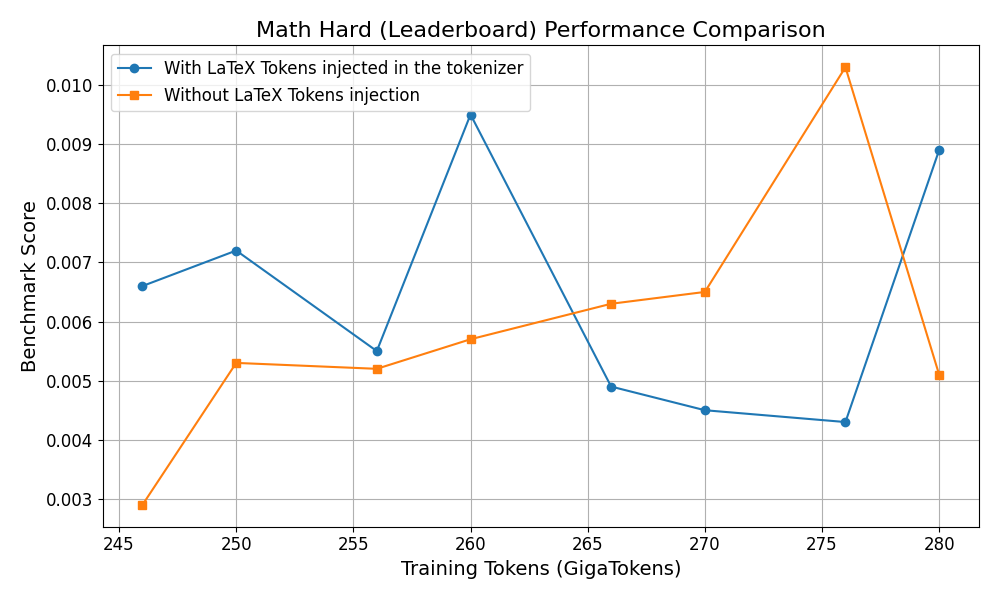}
        \label{fig:tokenizer-math-hard}
    \end{subfigure}
    
    \vspace{-1em}
    
    \begin{subfigure}[b]{0.45\textwidth}
        \centering
        \includegraphics[width=\textwidth]{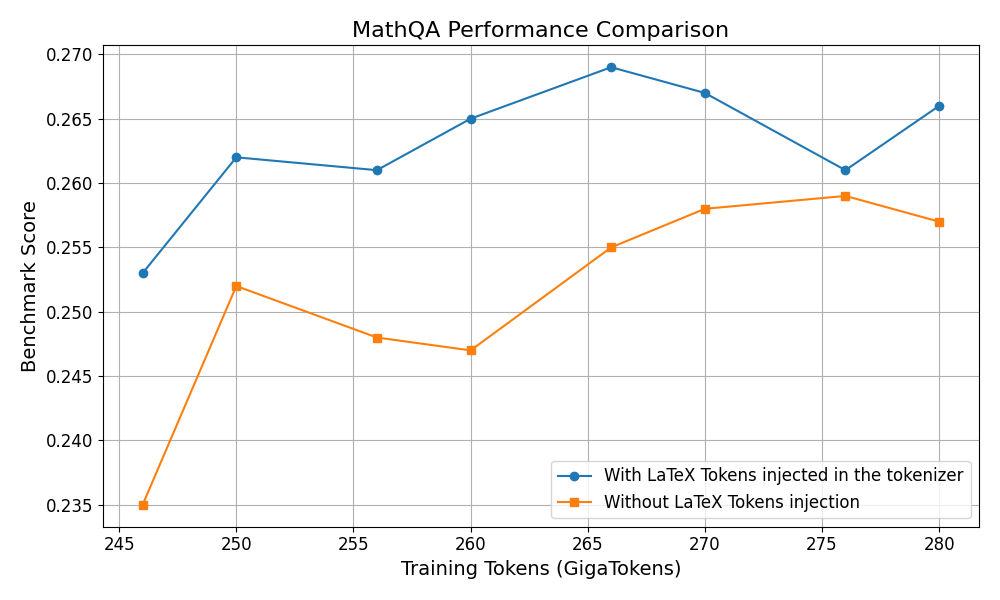}
        \label{fig:tokenizer-minerva-math}
    \end{subfigure}
    \hspace{1em} 
    \begin{subfigure}[b]{0.45\textwidth}
        \centering
        \includegraphics[width=\textwidth]{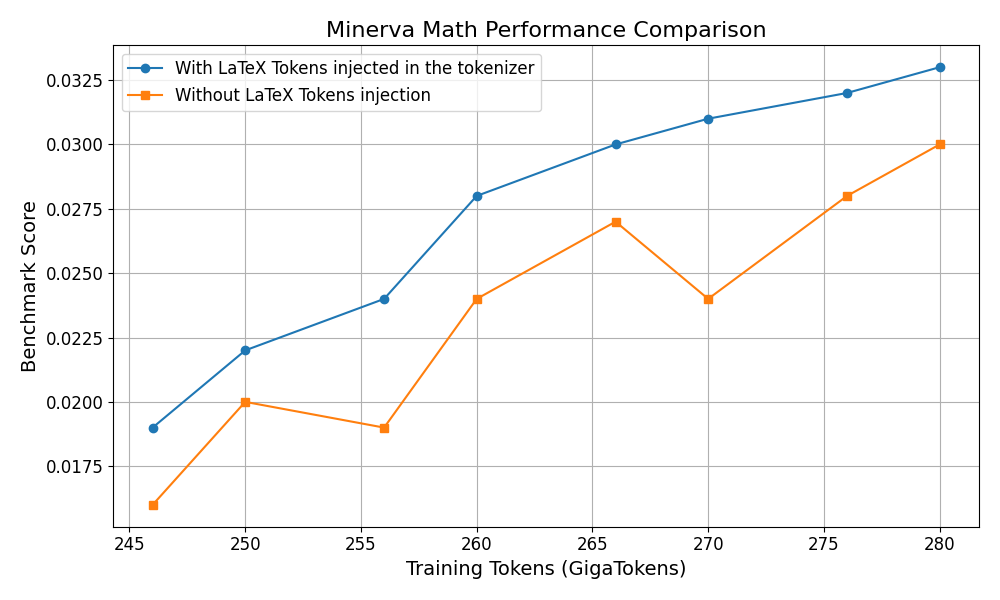}
        \label{fig:tokenizer-math-qa}
    \end{subfigure}
    
    \caption{Performance on mathematical benchmarks for two 1B models during the training decay stage. The model trained with a tokenizer augmented with specialized \LaTeX{} tokens (blue) consistently outperforms the baseline model (orange).}
    \label{fig:tokenizer-evals}
\end{figure}

To test this hypothesis, we curated a set of the most frequent \LaTeX\ commands, mainly from the Overleaf documentation to ensure comprehensive coverage of formatting, referencing, and mathematical functions. We then conducted a controlled experiment by training two 1B-parameter Falcon-Mamba models~\citep{zuo2024falconmamba} on an identical, math-heavy data mixture of 280GT (including a 40GT decay stage). One model used our baseline tokenizer, while the other used a version where unused vocabulary slots were replaced with our curated set of \LaTeX\ tokens.
The models were evaluated on four different math benchmarks: \texttt{MATH-Hard}~\citep{hendrycksmath2021}, \texttt{gsm8k}~\citep{cobbe2021gsm8k}, \texttt{math\_qa}~\citep{amini-etal-2019-mathqa} and~\texttt{minerva-math}. As shown in Figure \ref{fig:tokenizer-evals}, the model trained with the \LaTeX-augmented tokenizer demonstrated a consistent and notable performance improvement across most benchmarks.

\begin{tcolorbox}[colback=yellow!10!white, colframe=orange!50!black, boxrule=0.8pt]
\textbf{Conclusion:} Enriching the tokenizer's vocabulary with domain-specific tokens like \LaTeX\ is an effective strategy for boosting model performance on specialized tasks. By providing a native vocabulary for mathematical syntax, we directly facilitate the model's ability to process and reason about complex mathematical problems.
\end{tcolorbox}

\subsubsection{Final Tokenizer Design and Implementation}
Our final tokenization strategy employs the Byte Pair Encoding (BPE) algorithm~\citep{sennrich2016neuralmachinetranslationrare}, trained on an extensive multilingual corpus covering over 121 languages \ref{tab:language_codes} to ensure broad linguistic support. We developed a suite of tokenizers with vocabulary sizes of 32K, 65K, 130K, and 261K, each tailored to a specific model scale (Table~\ref{tab:model_architecture}). A primary design principle is to scale the vocabulary size in proportion to the model's overall architecture~\citep{tao2024scaling}. This balance is critical for preventing the embedding layer from becoming disproportionately large, thereby maintaining computational efficiency during fine-tuning and inference, particularly in resource-constrained environments.

The design of these tokenizers incorporates key findings from our empirical studies. Specifically, we implement both digit and punctuation splitting, which we found essential for improving performance on code-related tasks and ensuring accurate segmentation in non-Latin languages. Furthermore, based on our experiments demonstrating improved mathematical capabilities, we manually inject common \LaTeX{} commands directly into the vocabulary. To facilitate adaptability for downstream tasks, we reserve 1,024 special tokens across all tokenizers in the suite, providing end-users with the flexibility to customize the vocabulary for specific applications. A summary of the final trained tokenizers along with their vocabulary size is shown in the table \ref{tab:final_tokenizers}.

\begin{table}[ht]
\centering
\begin{tabular}{ccc}
    \toprule
    Tokenizer name & Vocabulary size & Model \\
    \midrule
    \texttt{Falcon-H1-32k} & 32,768 & \texttt{Falcon-H1-0.5B*} \\
    \texttt{Falcon-H1-65k} & 65,536 & \texttt{Falcon-H1-1.5B*}, \texttt{Falcon-H1-3B*} \\
    \texttt{Falcon-H1-131k} & 131,048 & \texttt{Falcon-H1-7B*} \\
    \texttt{Falcon-H1-262k} & 261,120 & \texttt{Falcon-H1-34B*} \\
    \bottomrule
\end{tabular}
\caption{List of available tokenizers trained for Falcon-H1}
\label{tab:final_tokenizers}
\end{table}

\section{Pretraining}

\subsection{Pretraining Data}

Capabilities of language models are known to come mainly from the training data, and that stays true for Falcon-H1 series. We have expanded our data corpus to more than 20 Teratokens, among which up to 18 Teratokens were used for training Falcon-H1 models. As shown in Table~\ref{tab:model_architecture}, each Falcon-H1 model comes with a different compute or token budget, considering their different learning capability and model capacity. The data mixtures are also designed differently across model scales. Noting that, although most Falcon-H1 models continued to show performance improvements toward the end of training, we chose to finalize training based on the allocated compute budget for each model.

The raw pretraining corpus was constructed from multiple sources, including web data, high-quality curated corpora, code, educational math content, and in-house synthetic data. We conducted an extensive evaluation and studies of the data mixture, performing exhaustive tests and applying data optimization strategies. Notably, beyond the domain-specific or knowledge-intensive data typically emphasized in many LLM training pipelines, we observed that the organization of knowledge within the dataset has a significant impact on model performance. A strong correlation was found between factors such as training epochs, the number of unique high-quality data tokens, and the proportion of each data type within a training epoch. However, isolating and analyzing these factors independently is impractical, as they are interdependent and would substantially increase the complexity of the analysis during pretraining.

\subsubsection{Data Sources}
\noindent \textbf{English Web data.} 
Starting from FineWeb~\citep{penedo2024fineweb}, we applied further quality filtering to improve knowledge density and training stability, using small language models as quality judges with carefully designed prompts. After processing the entire FineWeb dataset, we retained approximately 11T tokens.

\noindent \textbf{Multilingual data.}
Apart from English, Falcon-H1 models support 17 languages (except the 0.5B model, which is English-only): Czech (cs), German (de), Spanish (es), French (fr), Hindi (hi), Italian (it), Japanese (ja), Korean (ko), Dutch (nl), Polish (pl), Portuguese (pt), Romanian (ro), Russian (ru), Swedish (sv), Urdu (ur), and Chinese (zh). The multilingual data corpus draws from diverse sources—mainly Common Crawl and a range of curated datasets.

For multilingual web data from Common Crawl, language identification was first performed at the HTML level using \textit{pycld2}, then refined post-extraction with \textit{fasttext} and \textit{trafilatura}. The data was segmented into five similarly sized partitions (4–13 dumps each), covering 2012 to mid-2024. All dumps from 2022–2024 were included, with older dumps sampled randomly due to lower multilingual content. Processing followed the heuristics-based pipeline of~\citep{penedo2023refinedweb}, with language-specific tuning of Gopher Quality filtering, line-wise filtering, and stop words~\citep{malartic2024falcon2}.
A rule-based toxicity filter was applied using human-curated lists of offensive words. Native/proficient speakers rated each word (0: non-toxic, 1: context-dependent, 2: always toxic), and documents were filtered based on cumulative toxicity scores for which the formula was refined through human feedback.
The processed data was re-partitioned (two to five parts per language) and deduplicated via MinHash at the part level. For languages with limited web data from above-mentioned data sources, we supplemented with additional public datasets: CulturaY~\citep{nguyen2024culturay} (Hindi, Korean, Urdu) and FineWeb2~\citep{penedo2024fineweb-2} (Dutch, Romanian, Swedish, Korean, Urdu), processed through the same pipeline for consistency. The resulting multilingual web dataset for the 17 pre-trained languages totaled over 3,000 GigaTokens (GT), a portion of which was used during the pretraining phase.

To further enhance the multilingual data quality and diversity, we extract multilingual data from some highly curated data sources, including Wikipedia, academic preprints (arXiv, PubMed Central), online forums (Reddit, HackerNews, OpenSubtitles~\citep{tiedemann2016finding}, Ubuntu IRC, YouTube, StackOverflow), open-source books from Gutenberg~\citep{gutenberg}, public datasets like Europarl Corpus~\citep{europarl}, Gardian~\citep{gardian}, Makhzan~\citep{makhzan}, and proprietary data.

\noindent \textbf{Code data.}
Code data is widely considered as an important source for boosting a model's general and reasoning capabilities. We have internally curated an extensive code corpus that contains file-level, repository-level, and High Quality (HQ) code splits.  
For the file-level split, we used an internally scraped and curated dataset, 
spanning 67 programming languages (listed in Appendix~\ref{appendix:programming_langs}). The data was sourced from GitHub repositories created up to May 2024 and from notebooks within the Meta Kaggle Code dataset \footnote{\url{https://www.kaggle.com/datasets/kaggle/meta-kaggle-code}}. Notebooks were converted to scripts using JupyText~\footnote{\url{https://jupytext.readthedocs.io/}} after which all samples underwent heuristic filtering~\citep{lozhkov2024starcoder}. Language labeling followed the method in~\citep{penedo2023refinedweb}, retaining files classified under one of 18 target languages, with a relaxed acceptance threshold (0.15), given that non-English content can typically appear in comments.
Fuzzy deduplication was performed using MinHash and Local Sensitive Hashing (LSH)~\citep{MinHash}, computing 256 hashes per document with 5-grams and a Jaccard similarity threshold of 0.85. Personally Identifiable Information (PII), such as email addresses and IP addresses, was redacted using a pipeline inspired by DataTrove~\footnote{\url{https://github.com/huggingface/datatrove/blob/main/src/datatrove/pipeline/formatters/pii.py}} replacing tokens with standardized placeholders (e.g., \verb|>>EMAIL_ADDRESS<<|, \verb|>>IP_ADDRESS<<|).

The repository-level split containing long context code data was built with repository-level code samples~\citep{deepseekcoder,qwencoder}, which were constructed by concatenating all source files from a given repository, enabling cross-file code comprehension.
MinHash deduplication was performed at the repository level (prior to filtering), to preserve the logical structure of each repository. Files were concatenated in alphabetical order for effective duplicate detection, and later shuffled to mitigate bias and ensure a balanced representation. 

For both file-level and repository-level code data, we curated a high-quality (HQ) split by applying code quality classifiers.
For non-Python languages, we used a CodeBERT-based classifier~\footnote{\url{https://huggingface.co/devngho/code_edu_classifier_v2_microsoft_codebert-base}} covering 19 programming languages (see the full list in Appendix~\ref{appendix:code_classifier}), selecting samples that met a predefined quality threshold.
For Python, we applied both the CodeBERT-based classifier and a specialized Python scorer~\footnote{\url{https://huggingface.co/HuggingFaceTB/python-edu-scorer}}, retaining only samples that met the threshold for both classifiers. 
Additionally, we supplemented this corpus with the OpenCoder annealing corpus~\citep{huang2024opencoder}, which includes algorithmic data, synthetic QA, and synthetic code snippet datasets~\footnote{\url{https://huggingface.co/datasets/OpenCoder-LLM/opc-annealing-corpus}}. We further preprocessed those data following the same methodology used for our file-level corpus.

\noindent \textbf{Math data.}
We used a combination of open-source datasets and in-house crawled or retrieved math data from the Web. The open-source datasets include Proof-Pile-2~\citep{azerbayev2023llemma}, FineMath~\citep{liu2024finemath}, InfiMM-WebMath-40B~\citep{han2024infimm}, and OpenCoder FineWeb Math corpus~\citep{Huang2024OpenCoderTO}, largely consisting of math data extracted and filtered from Common Crawl.
For the in-house math data, following a similar approach to~\citep{shao2024deepseekmath}, we first used a \textit{fastText} classifier trained on OpenWebMath~\citep{paster2023openwebmath} to retrieve OpenWebMath-like pages. The classifier was then iteratively refined with extracted math data to expand coverage, targeting top math-related domains from Common Crawl.
All math data underwent a decontamination process to remove overlaps with popular math benchmarks such as GSM8K~\citep{cobbe2021gsm8k} and MATH~\citep{hendrycks2021measuring}.

\noindent \textbf{Synthetic data.} 
We used a combination of external open datasets and a large volume of in-house generated synthetic data. External sources include subsets of Nemotron-CC~\citep{su2024nemotroncctransformingcommoncrawl} (\texttt{diverse\_qa\_pairs}, \texttt{extract\_knowledge}) and Cosmopedia v2~\citep{benallal2024smollmcorpus}. 
From our experiments, we found that fully synthetic data generated without grounding on seed samples often lacks diversity and suffers from inconsistent quality. To address this, our in-house synthetic data was primarily created by rewriting curated raw data—including web, code, books, Wikipedia, arXiv, math sources, etc.
We employed a diverse set of internal and external open models across various scales and architectures, considering the compute cost, data quality, and content diversity. 
Instead of focusing only on question–answer pairs, our generation strategies included enhancing writing style, increasing knowledge density, filtering redundant tokens, and applying iterative quality control to the generated samples. This rewriting process helped structure and formalize the underlying knowledge, reduce noise, and ultimately improve training stability and efficiency. 

Apart from rewriting raw samples, we also generate raw question-answer pairs from DeepMind's mathematics dataset~\citep{dm_math}, which are further enhanced with context-enriched questions, and correct chain-of-thought solutions generated in a way similar to STaR~\citep{STaR}. To further increase the knowledge intensity of our pretraining corpus, we also generated synthetic textbooks using carefully constructed topic hierarchies extracted from Wikipedia, covering over 30K topics. Starting from 99 root categories (see Table~\ref{tab:wikipedia-categories-synthetic} in Appendix), we crawled the Wikipedia category graph up to a depth of 5 (or until reaching article nodes), followed by deduplication and pruning of irrelevant topics. For each hierarchy, we first generated a table of contents, then created structured content for each unit—including comprehensive explanations, relevant examples, and diverse exercises.

\noindent \textbf{Long context data.} 
Long-context data naturally appears in sources such as web pages, repository-level code, books, and academic papers (e.g., arXiv). To enhance the model’s ability to handle extended sequences, we applied a range of restructuring strategies across target lengths of 32K, 128K, and 256K tokens. These include Fill-in-the-Middle (FIM), where random sections are removed from documents, and section reordering, where segments are shuffled and the model is tasked with reconstructing the original order—requiring a combination of long-context reasoning, memory, and coherence understanding. Additionally, we created a small set of synthetic long-context samples with question–answer pairs to further improve capabilities such as in-context learning, complex pattern retrieval, etc.

\subsubsection{Data Strategy}
\noindent \textbf{Data validation.} 
We carefully inspect each data source prior to injecting it into the training pipeline to obtain a detailed understanding of its quality and its impact on domain-specific tasks as well as on overall model performance. 
More specifically, to derive strong, interpretable signals for each data source, we train multiple 0.5B-scale models from scratch, subsequently evaluating their performance on carefully selected domain benchmarks.
An alternative strategy involves training a single model from scratch and then modifying the data mixture during the learning rate decay stage to assess data quality with reduced computational costs~\citep{grattafiori2024llama}. However, in our experiments,  this approach sometimes failed to yield clear or robust conclusions regarding data quality. This is primarily due to additional confounding factors, e.g., data distribution shift between the stable and decay stage, correlations in data used across these stages, and data mixtures within each of these stages, etc. 
To this end, we systematically train 0.5B models from scratch using either individual data sources or well-studied combinations of them. This enables us to isolate extra factors and examine multiple dimensions: absolute data quality, relative quality in comparison to existing datasets, interactions and correlations between different data sources and formats, and their respective impacts on various domain-specific tasks. Once the data is validated at the 0.5B scale, it will be passed to the models at medium and large model scales, being retained or adjusted based on the observed improvements in model performance.



\noindent \textbf{Deterministic data loading and check-pointing.} 
When adopting multiple data sources into training, we implemented a deterministic dataloader that reads samples sequentially from each source rather than sampling them randomly. This design offers several advantages: a) deterministic training behavior, enabling precise comparisons across different runs; b) flexible continual pretraining through coordinated model and data checkpointing across multiple data sources, without disrupting the loading order; c) dynamic data mixture updates during pretraining without needing to restructure data files or data classes; and d) improved control over multi-epoch training, allowing custom epoch sizes across diverse data sources on the fly - we define \textit{one training epoch} as a full pass over all unique tokens from a given data source.

\noindent \textbf{Data mixture.}
Effective data organization strategies proved critical to the final performance of our models. These strategies include choices around the \textit{data mixture}, \textit{pre-training stages}, and \textit{multi-epoch training}. In practice, these factors are tightly interconnected and require joint optimization, taking into account compute budgets, model scales, data source quality, and data volumes.

One common concern when reducing the proportion of web data is its potential impact on model generalization and knowledge diversity. However, we found this concern might be somewhat overstated. While popular web datasets like FineWeb~\citep{penedo2024fineweb} or RefinedWeb~\citep{penedo2023refinedweb} offer broad topical diversity, their knowledge density is relatively low, even when enhanced by domain-specific filtering such as FineWeb-Edu~\citep{penedo2024fineweb} or FineFineWeb~\citep{FineFineWeb}. With highly knowledge-intensive data sources and rewritten web samples, the raw web data can be significantly reduced without impacting model generalization and knowledge diversity.

Through extensive experiments on data quality and validation, we iteratively refined and converged on an optimal data mixture across the pretraining process. Starting with an initial mixture validated on smaller models, we progressively adjusted the mixture throughout the training by incorporating newly optimized and carefully evaluated data at various model scales. This process led to the final data configurations for the 34B and 7B models (Table~\ref{tab:data_mix}), where web data accounted for only about 15\% and 12.35\%, respectively—substantially increasing the share of rewritten data (over raw web, code, curated sources, math data, etc.). For the smaller 3B, 1.5B, and 0.5B models, we maintained a constant data mixture throughout the whole pretraining stage, leveraging the dynamic data preparation carried out during the pretraining of the larger models.

\begin{table}[htbp]
\centering
\sisetup{table-format=2.2, table-space-text-post = \textsuperscript{*}}

\begin{tabular*}{\textwidth}{l @{\extracolsep{\fill}} *{7}{S}}
\toprule
& \multicolumn{2}{c}{\textbf{34B}} & \multicolumn{2}{c}{\textbf{7B}} & {\textbf{3B}} & {\textbf{1.5B}} & {\textbf{0.5B}} \\
\cmidrule(lr){2-3} \cmidrule(lr){4-5}
\textbf{Data Source} & {\makecell{Start}} & {\makecell{End}} & {\makecell{Start}} & {\makecell{End}} & {\makecell{Mix}} & {\makecell{Mix}} & {\makecell{Mix}} \\
\midrule
\textbf{Raw data}          & 99.47 & 43.45 & 81.07 & 42.26 & 39.70 & 23.20 & 11.50 \\
\quad Web                  & 40.00 & 14.60 & 25.00 & 12.35 & 11.60 & 10.20 &  6.50 \\
\quad Curated              & 25.00 & 15.93 & 26.00 & 16.47 & 11.68 &  4.75 &  0.00 \\
\quad Code                 & 20.00 & 10.05 & 20.00 & 10.74 & 14.00 &  8.00 &  5.00 \\
\quad Math                 & 14.47 &  2.87 & 10.07 &  2.70 &  2.42 &  0.25 &  0.00 \\
\midrule
\textbf{Rewritten data}    &  0.23 & 52.05 & 10.56 & 53.04 & 56.80 & 69.80 & 75.50 \\
\quad Web \& Curated       &  0.00 & 20.36 &  0.00 & 18.12 & 22.08 & 23.75 & 20.50 \\
\quad Code \& Math         &  0.23 & 31.69 & 10.56 & 34.92 & 34.72 & 46.05 & 55.00 \\
\midrule
\textbf{Synthetic data}\textsuperscript{*}  &  0.30 &  4.50 &  8.37 &  4.70 &  3.50 &  7.00 & 13.00 \\
\midrule
\textbf{Total} & 100.00 & 100.00 & 100.00 & 100.00 & 100.00 & 100.00 & 100.00 \\
\bottomrule
\end{tabular*}
\parbox{\textwidth}{\textsuperscript{*} \footnotesize{Fully synthetic samples, not derived from rewriting existing raw data.}}
\caption{Falcon-H1 data mixtures across model sizes. For the 34B and 7B models, both start and end-of-training mixtures are shown. For the 3B, 1.5B, and 0.5B models, a single, static data mix was used. All values are in percent (\%).}
\label{tab:data_mix}
\end{table}

To mitigate the risk of overfitting or bias towards specific domain tasks and to preserve a balanced performance across diverse skill areas, we perform frequent checkpoint evaluations over diverse domain tasks throughout the pretraining stage. Additionally, whenever the data mixture changed, we perform frequent \textit{vibe checks} on intermediate model checkpoints. This combined strategy ensures that the model maintains strong general capabilities while avoiding unintended specialization or biases toward particular domains.

\noindent \textbf{Data organization and scheduling.} 
A primary challenge in large-scale pretraining is the profound imbalance between the vast quantities of web-scale data and the relative scarcity of high-quality, curated datasets. Specialized corpora like mathematical texts, while crucial for model capability, comprise a small fraction of the total training data corpus. To ensure these high-quality sources maintain their influence, we employ an aggressive up-sampling strategy enabled by multi-epoch training. This approach decouples the effective training data size from the raw token count of our finite, high-quality corpora.

\begin{figure}[!ht] 
    \centering 
    \includegraphics[width=\textwidth]{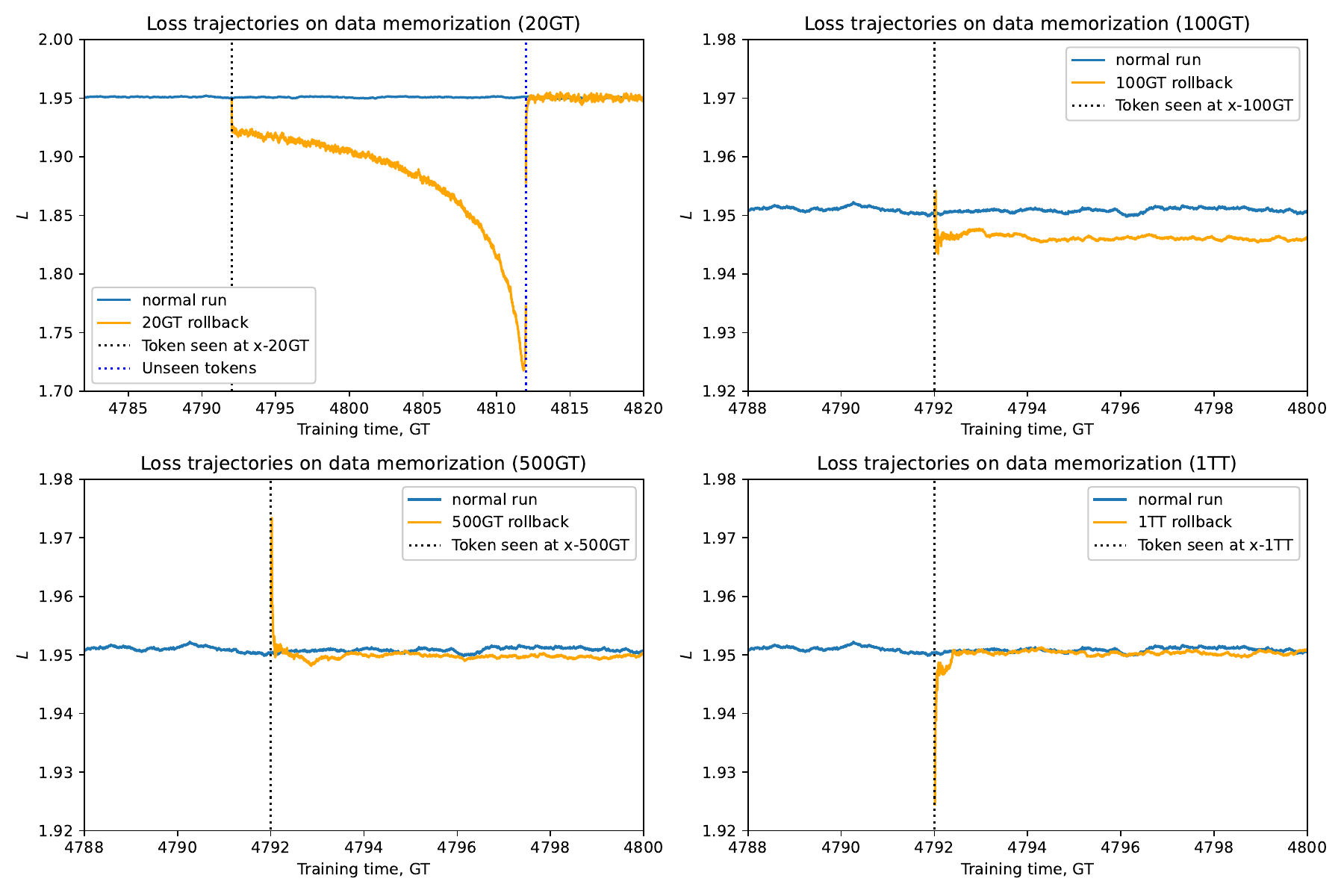} 
    \caption{Model's memorization window and loss trajectives} 
    \label{fig:memorization_window} 
\end{figure}

This extensive reuse of data challenges the common practice of single-epoch training, often adopted to mitigate the risk of model memorization. However, our empirical investigation suggests that concerns about memorization may be overstated in large-scale regimes. By estimating the model's \textit{memorization window}—the temporal span over which it retains specific training examples—we found it possible to reuse high-quality samples multiple times without compromising generalization. As illustrated in Figure~\ref{fig:memorization_window}, this window can be approximated by analyzing the training loss on tokens seen at various points in the past. This finding is particularly relevant as most literature on memorization and catastrophic forgetting focuses on smaller models or shorter training durations.

Beyond data composition, we found data scheduling to be a critical factor. Counter-intuitively, our experiments revealed that an "anti-curriculum" approach yielded superior results compared to conventional curriculum learning. In this paradigm, we introduce data of all complexity levels—from simple text to advanced mathematical problems—from the very beginning of training. We hypothesize that this early and continuous exposure provides the model with a more effective learning trajectory to develop the internal features required to master complex tasks.
Across models of various scales, our experiments confirmed that maintaining this optimized data mixture from the outset outperforms curriculum strategies that reserve high-quality data for later training stages. This finding holds when a sufficient volume of high-quality data is available.

\noindent \textbf{Long-context training.} 
We initiated pretraining with an 8K context window, with the majority of pretraining conducted at a 16K context length. An exception is Falcon-H1-0.5B, which was primarily trained at a 4K context due to its limited capacity for handling longer sequences. Thanks to our infrastructure optimizations, throughput degradation at 16K context was minimal. Moreover, we observed a performance boost compared to training with 4K or 8K context lengths.
For long-context extension, we maintained the overall data mixture largely unchanged, but slightly increased the proportion of long-context samples within each data source. Following the learning rate decay phase, we continued training with a fixed minimum learning rate during long-context extension. We empirically found that performing this extension after the decay phase yielded no significant performance difference compared to extending it before decay. However, the post-decay approach was substantially more compute-efficient due to the reduced training throughput of long sequences. For Falcon-H1-34B and Falcon-H1-7B, we applied 60GT at both 32K and 128K context, and 25GT at 256K context. For Falcon-H1-3B and Falcon-H1-1.5B, the same token counts were used for both the 32K and 128K context stages. 


\subsection{Training Dynamics}

\subsubsection{Training Stability}
During early experiments with Falcon-H1 we observed severe loss spikes from the beginning of the training.  
These spikes created two practical problems:
\begin{enumerate}[label=(\roman*)]
  \item Spiky loss curves distort ablation results: a variant may look inferior simply because a spike coincides with the learning-rate decay, reversing the true ordering; and
  \item continuing with such spikes would force us to choose learning rates well below the optimum, which would slow convergence.
\end{enumerate}
Eliminating the instability therefore became a prerequisite for any meaningful ablation.

Spike-like behaviour has been reported before in \textit{Falcon2}~\citep{malartic2024falcon2}, \textit{Falcon-Mamba}~\citep{zuo2024falconmamba}, \textit{Jamba}~\citep{lieber2024jamba}, and \textit{YuLan}~\citep{hu2024yulanmini}.  
Falcon2 attributed the issue to depth, whereas YuLan pointed to three potential triggers: exploding residual paths, unstable layer-norm statistics, and extreme attention scores.  

A common practice is to employ \emph{batch-skipping}: whenever the loss on a batch exceeds a user-defined multiple of the running median, that batch is dropped and the optimiser moments remain unchanged.  
This heuristic can suppress spikes caused by a handful of outlier examples, yet it is largely ineffective for \emph{dynamics-induced} instabilities and only only postpones the problem.

\paragraph*{Isolating the Source.}\label{sec:source_isolation}

Our initial experiments were conducted on a \emph{pure Mamba2} baseline.  
Loss spikes were already present in this setting, showing that the SSM pathway is \emph{sufficient} to trigger the instability—although it may not be the sole contributor.

To test whether the spikes were driven by width or depth, we compared two vanilla Mamba2 variants trained under identical conditions:

\begin{itemize}
    \item \textbf{Wide}: larger hidden dimension and more SSM heads, but fewer layers;
    \item \textbf{Deep}: smaller hidden dimension, comparable total parameter count, but more layers.
\end{itemize}

All other hyper\-parameters were kept fixed, and the learning rate was scaled as per $\mu$P scaling.

The wide model exhibited pronounced loss spikes, whereas the deep model trained smoothly.  
This finding implicates width\-related dynamics inside the SSM—specifically the larger number of heads—as a primary driver of the observed instability, and motivated the analysis reported in the next sections.

\paragraph*{Diagnosing the SSM Dynamics.}\label{sec:ssm_dynamics}

We logged parameter statistics throughout training and found that spiking runs displayed a wider range of \(\widetilde{dt}_t\) values. Recall from \eqref{eq:siso-ss} that $dt_t=\operatorname{softplus}(\widetilde{dt}_t+b)$ controls both forgetting of and writing to SSM hidden state, as can be seen from \eqref{eq:siso-ss}. Therefore, large positive \(\widetilde{dt}_t\) values have two antagonistic effects: linearly enlarging the information from the \emph{current} token written into the hidden state while exponentially forgetting the information from the \emph{previous} tokens.

Whenever the modeling objective requires both the recent token and its long-range context, gradient descent is pulled in opposite directions:
\begin{enumerate}[label=(\roman*)]
    \item increase \(\widetilde{dt}_s\) so that \(dt_s\) increases and amplifies the contribution of token \(s\) through $\mathbf{C}_t^{\!\top}\mathbf{B}_s\,dt_s$ in \eqref{eq:attention-form}.
    \item keep \(\prod_{j=s'}^{s}\overline{A}_j\) with \(s'\) < \(s\) from collapsing so that earlier information still reaches position \(t\) and propagates further.
\end{enumerate}

Because these antagonistic signals arrive at different scales and at different optimization steps, the parameter overshoots and then over-corrects, producing the characteristic loss spikes.


 
\paragraph*{Mitigation.}\label{sec:mitigation}

We tested three interventions:
\begin{itemize}
  \item Clip \(A_{\log}\): \textit{no effect.}
  \item Clip negative \(dt\): \textit{no effect.}
  \item Clip positive \(dt\): \textbf{completely removed spikes.}
\end{itemize}

This confirms our previous hypothesis: spikes are caused by writing to the hidden state. However, we recognize that such clipping may restrict expressiveness. 
A softer alternative is to multiply the \(dt\) activation by a constant \(0<\alpha<1\).  
This \emph{attenuation} preserves the full parameter range while preventing early excursions into the unstable regime.

With attenuation enabled we can train Falcon-H1 at relatively high learning rates without observing any loss spikes. The attenuation factor is a part of the $\mu P$ forward multipliers (see: \ref{subsubsec:enh_mup}) and will be tuned as well with other multipliers.

\subsubsection{Effective Learning Rate and Effective Weight Decay}\label{sec:param_norms_and_schedules}
\begin{figure}
    \hspace{-3mm}
    \includegraphics[scale=0.45, clip, trim=130 0 120 15]{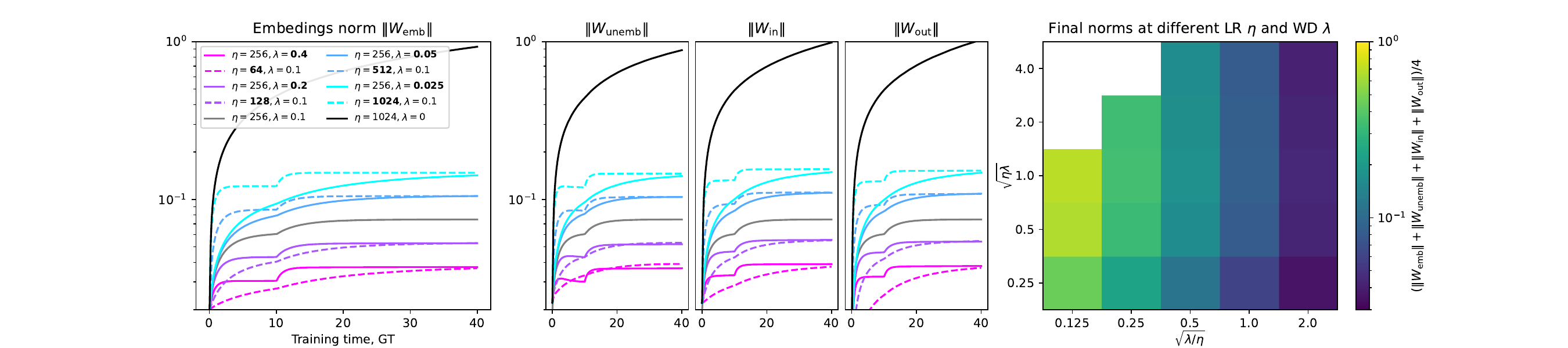}
    \caption{Behavior of weight norms of embedding $W_\mathrm{emb}$, unembedding $W_\mathrm{unemb}$, and input/output projections $W_\mathrm{in},W_\mathrm{out}$ layers of pure Mamba2 model. \textbf{(Left)} Evolution of weight norms during training of 1B model. In addition to no WD run, we show WD sweep (solid) and LR sweep (dashed), both of which display similar impact on weight norms. Learning rate is measured in the units of $10^{-5}$; and the jump of norms at 10GT corresponds to batch size doubling within rampup. \textbf{(Right)} Average weight norm across all model layers at two-dimensional $(\eta,\lambda)$ sweep on 300M model. The combinations $\sqrt{\frac{\lambda}{\eta}}$ and $\sqrt{\eta\lambda}$ are used as coordinate axes to demonstrate that weight norms are scaled as in \eqref{eq:weight_norm_scaling}. LR and WD are measured in the units of $\eta_0=256\times 10^{-5}$ and $\lambda_0=0.1$. Empty cells correspond to diverged runs.    
    }
    \label{fig:parameter_norms}
\end{figure}

\paragraph{Parameter norms.} To initiate the discussion of the joint effect of AdamW learning rate (LR) $\eta$ and weight decay (WD) $\lambda$ on the model training, we start with the behavior of parameter norms of matrix layers, to which WD was applied during training. Figure \ref{fig:parameter_norms} (left) shows that parameter norms grow indefinitely with no WD $\lambda=0$ while stabilizing at a constant level when $\lambda>0$. The value of parameter norms after stabilizing $||W||$ depends on LR, WD, and also batch size $B$, raising the question of the precise form of this dependence.   

Leaving the dependence of parameter norms on batch size to the future work, we have found that the dependence of $||W||$ on $(\eta,\lambda)$ is well described, for all matrix layers $W$, by a simple scaling rule
\begin{equation}\label{eq:weight_norm_scaling}
    ||W|| \propto \sqrt{\frac{\eta}{\lambda}}.
\end{equation}
Moreover, if the norm is normalized in a mean fashion as $\|W\|^2=\frac{1}{mn}\sum_{i=1}^n\sum_{j=1}^m W_{ij}^2$, its values for different layers $W$ in the architecture are extremely close to each other, see figure \ref{fig:parameter_norms} (left).  

In fact, the scaling \eqref{eq:weight_norm_scaling} is natural and is observed in a simple model of stochastic dynamics where Brownian expansion due to noise in the gradients dominates the attraction to the optimal parameter value. We provide such a toy model in section \ref{sec:toy_model_for_param_norms} and show that in the relevant regime of hyperparameters, the dependence weight norm on LR and WD has the functional form 
\begin{equation}\label{eq:toy_model_norm_scaling}
    ||W(\eta,\lambda)||^2 = C_1 \frac{\eta}{\lambda} + C_2 \frac{1}{\lambda^2},
\end{equation}
where the constants $C_1,C_2$ are determined by the variance of the gradient noise and steepness of the loss function near the optimal value of the parameters. Consider the region of $(\eta,\lambda)$ typically used in the model pertaining. One extreme scenario is large noise variance and a flat loss landscape, where the weight norm is mostly determined by the balance between weight decay contraction and Brownian expansion due to stochastic optimizer updates. Then, we have $C_1 \frac{\eta}{\lambda} \gg C_2 \frac{1}{\lambda^2}$ and the weight norms scaling is given by \eqref{eq:weight_norm_scaling}. The opposite extreme scenario is small noise variance and steep loss landscape, where the weight norm is mostly determined by the balance between weight decay contraction and attraction of the parameters to the optimal value $W^*$. Then, we would have $C_1 \frac{\eta}{\lambda} \ll C_2 \frac{1}{\lambda^2}$ and the weight norms scaling be $||W||\propto \lambda^{-1}$. 

To determine which of the two above scenarios better describes the behavior of the norms during the actual model training, we performed a 2d sweep over $(\eta,\lambda)$ values and measured the norms by the end of the constant LR stage of WSD schedule, just before LR decay. The results are depicted in the figure \ref{fig:parameter_norms} and show that the scaling \eqref{eq:weight_norm_scaling} indeed describes parameter norms very well, suggesting that the noise plays the dominant role in the dynamics of the matrix layers. 

\paragraph{Effective learning rate.} 
It is convenient to assign to both LR and WD a simple, intuitive role they play in the training process. Weight decay squeezes the learnable parameters and, therefore, controls their norms. Learning rate determines the speed of learning, and also the noise level in the presence of label noise~\citep{liu2025neuralthermodynamiclawslarge}. As a result, the final loss has a steep dependence on the LR, making it the most important parameter to be tuned. 

\begin{figure}
    \centering
    \includegraphics[scale=0.55, clip, trim=10 10 10 5]{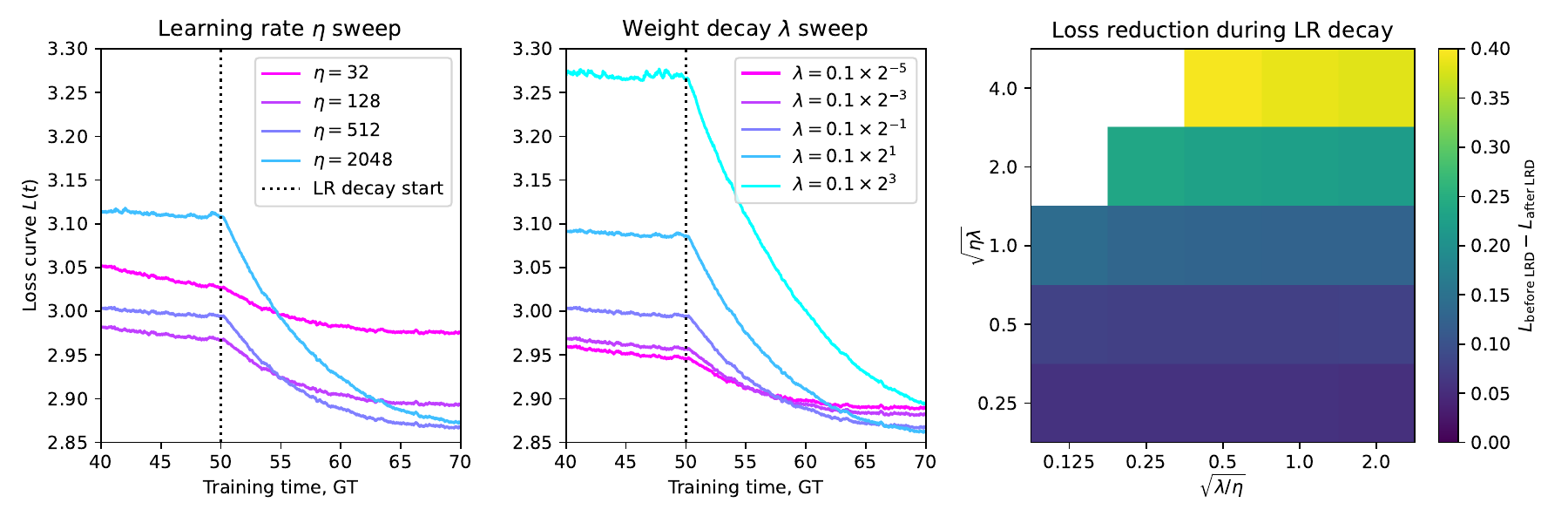}
    \caption{
    Effect of LR $\eta$ and WD $\lambda$ of the training loss curve of a 300M pure Mamba2 model. \textbf{(Left)} LR and WD sweeps on left and right subplots show that increasing (or decreasing) either LR or WD has a similar effect on the loss curve. \textbf{(Right)} Noise level measure as the loss gap before and after LR decay $L_\text{noise}=L_{\text{before LRD}}-L_{\text{after LRD}}$ at two-dimensional $(\eta,\lambda)$ sweep. Again, the use of ELR and EWD for coordinate axis shows that the noise is mostly determined by $\eta_\mathrm{eff}=\sqrt{\eta\lambda}$. 
    }
    \label{fig:effective_learning_rate}
\end{figure}

This simple intuition does not hold for weight decay as we saw in the previous section: the combination $\sqrt{\frac{\lambda}{\eta}}$ controls the parameter norms instead of WD $\lambda$ itself. Moreover, the LR intuition also does not hold, as we show in Figure \ref{fig:effective_learning_rate} (Left): changing $\lambda$ has a similar effect on the loss curve as changing $\eta$. We argue that simple LR and WD intuition can be kept if we switch original $\eta,\lambda$ with \emph{effective learning rate} (ELR) and \emph{effective weight decay} (EWD) defined as
\begin{equation}\label{eq:ELR_EWD_def}
    \text{(ELR)} \quad \eta_\mathrm{eff} = \sqrt{\eta \lambda}, \qquad \text{(EWD)} \quad \lambda_\mathrm{eff}=\sqrt{\frac{\lambda}{\eta}}.
\end{equation}

Essentially, the EWD definition was already verified in the previous section due to parameter norms scaling \eqref{eq:weight_norm_scaling}, equivalent to $||W||\propto \lambda_\mathrm{eff}^{-1}$. Here we take $||W||\propto \lambda^{-1}$ as an intuitive functional form of parameter norm scaling that can be observed, for example, for minimizers of $L_2$ regularized loss with strong regularization $\lambda$, or in the weak noise scenario of \eqref{eq:toy_model_norm_scaling}.

A heuristic way to obtain ELR definition \eqref{eq:ELR_EWD_def} is to look at an optimizer part $\delta W_t$ of a single step update $W_{t+1}=W_t-\eta \delta W_t-\lambda\eta W_t$ of a linear layer $W$, and define ELR as ``meaningful'' measure of the update strength associated with $\delta W_t$. We start with a proposition that relative magnitude of the parameters change $\frac{||\eta\delta W_t||}{||W_t||}$ is a more meaningful measure than the absolute magnitude $||\eta\delta W_t||$. Then, recall that for Adam the update is given by a ratio $\delta W_t=\frac{G_{1,t}}{\sqrt{G_{2,t}}+\epsilon}$ of gradient moments $G_{1,t},G_{2,t}$. An important empirical observation is that each component of the ratio has roughly the same magnitude around $0.1$, almost independent from LR and WD used in the training. As a result, we can treat $\|\delta W_t\|$ as a constant, leading to simplified update strength measure $\frac{||\eta\delta W_t||}{||W_t||}\to\frac{\eta}{\|W_t\|}$. Next, assume that the parameter norms have already stabilized at their stationary state given by the scaling $||W_t||\propto \sqrt{\frac{\eta}{\lambda}}$. This further changes update strength measure to $\frac{\eta}{\|W_t\|}\to\sqrt{\eta\lambda}$, which is exactly our ELR definition \eqref{eq:ELR_EWD_def}.

Now, let us examine the properties of ELR on the noise level during training, that we can roughly measure as a difference of the loss just before and after learning rate decay $L_{\text{before LRD}}-L_{\text{after LRD}}=L_\text{noise}$. In the figure \ref{fig:effective_learning_rate} (right) we plot $L_\text{noise}$ for different values of ELR $\eta_\mathrm{eff}$ and EWD $\lambda_\mathrm{eff}$. We see that keeping $\eta_\mathrm{eff}=\mathrm{const}$ and changing $\lambda_\mathrm{eff}$ has a weak effect on $L_\text{noise}$ while keeping $\lambda_\mathrm{eff}=\mathrm{const}$ and changing $\eta_\mathrm{eff}$ affects the noise level strongly. We can roughly interpret this observation as the noise level being the function of only the effective learning rate $L_\text{noise}(\eta,\lambda)=L_\text{noise}(\eta_\mathrm{eff})$. Essentially, this shows that ELR actually controls the noise level and closes our earlier statement that learning rate and weight decay intuition hold for EWD and ELR.  

Finally, let us point out to an additional property of EWD and ELR that we call \emph{log-scale orthogonality} and which is useful for $(\eta,\lambda)$ sweeps often done within empirical tuning of training hyperparameters. A robust strategy for such sweeps is to use a log-scaled grid, for example, with the powers of 2: $(\eta,\lambda)\in\{\eta_0 2^i, \lambda_0 2^j\}_{i,j=0}^n$, which ensures a good balance between precision and coverage. Such experiment design naturally corresponds to the Euclidean metric, but in log-coordinates $(\log \eta, \log \lambda)$. In this metric, the gradients of EWD and ELR turn out to be orthogonal, unlike in the standard Euclidean metric on the $(\eta, \lambda)$ plane.
\begin{equation}
    \nabla^{(\mathrm{log})}\eta_\mathrm{eff} \cdot \nabla^{(\mathrm{log})}\lambda_\mathrm{eff}= \frac{\partial \eta_\mathrm{eff}}{\partial \log \eta} \frac{\partial \lambda_\mathrm{eff}}{\partial \log \eta} + \frac{\partial \eta_\mathrm{eff}}{\partial \log \lambda} \frac{\partial \lambda_\mathrm{eff}}{\partial \log \lambda}=0. 
\end{equation}
With the above orthogonality property, the sweeps on log-scale grids contain directions that maximally change the noise level (via ELR) while keeping the parameter norms constant (via EWD), and vice versa. This makes possible to effectively measure the effect of noise level and parameter norms on training dynamics, and we will utilize it for our $\mu P$ sweeps in the next section.

To summarize, ELR and EWD defined in \eqref{eq:ELR_EWD_def} have the following useful properties 
\begin{enumerate}[label=(\Alph*)]
    \item \textbf{Parameter norms} $||W||$ are fully determined by EWD $\lambda_\mathrm{eff}$
    \item \textbf{Noise level} is fully determined by ELR $\eta_\mathrm{eff}$.
    \item \textbf{Log-scale orthogonality.} Directions of increase of ELR and EWD are orthogonal on $(\log\eta,\log\lambda)$ plane, allowing to separately and efficiently measure their effects within sweeps on log-scale grids. 
\end{enumerate}
We emphasize, however, that the above conclusions are ``rough'' approximations based on a limited set of experiments performed for Falcon-H1 training, and accurate investigation of these questions would be an exciting direction for future work.

\paragraph{Incorporating parameter norms scaling into the power scheduler.} Recent work~\citep{shen2024powerschedulerbatchsize,bjorck2025scaling} noticed that for longer training durations $T$, the optimal learning rate of the WSD scheduler scales as $\eta_\mathrm{opt}\propto T^{-b}$ with the exponent $b$ roughly in the range $[0.3,0.5]$. As a next step,~\citep{shen2024powerschedulerbatchsize} suggests to downscale the learning rate of the stable stages starting from specified activation time $t_0$ as $\eta(t)=\eta_0 \sqrt{\min(1,\frac{t_0}{t})}$, referred to as power scheduler (PS). Our internal tests support these conclusions with PS schedule achieving better loss than learning rate tuned WSD at different training token scales. 

However, we also decided to incorporate our insights on the role of weight decay into the schedule. Specifically, let us relate the PS scaling of learning rate and weight decay to the scaling of ELR and EWD 
\begin{equation}\label{eq:classical_power_scheduler_scalings}
    \begin{cases}
        \eta(t) \propto t^{-\frac{1}{2}}\\
        \lambda(t) \propto \mathrm{const}(t)
    \end{cases}
    \qquad \implies \qquad
    \begin{cases}
        \eta_\mathrm{eff}(t) \propto t^{-\frac{1}{4}}\\
        \lambda_\mathrm{eff}(t) \propto t^{\frac{1}{4}}
    \end{cases}
\end{equation}
Now, we combine two observations. On the one hand, we observed that the final loss is much more sensitive to ELR than to EWD. In that case, we can mostly attribute the effectiveness of the PS scaling \eqref{eq:classical_power_scheduler_scalings} due to the scaling of ELR rather than EWD. On the other hand, we still observed substantial gains from tuning weight decay $\mu P$ multipliers, suggesting that it is beneficial to have tuned parameter norms of all the layers in the architecture. Then, if we would like to keep parameter norms at the ``optimal'' level throughout the whole training, we should not scale EWD, which controls the parameter norms. This brings us to \emph{Effective Power Scheduler} (EPS) scaling 
\begin{equation}
    \begin{cases}
        \eta_\mathrm{eff}(t) \propto t^{-\frac{1}{4}}\\
        \lambda_\mathrm{eff}(t) \propto \mathrm{const}(t)
    \end{cases}
        \qquad \implies \qquad
    \begin{cases}
        \eta(t) \propto t^{-\frac{1}{4}}\\
        \lambda(t) \propto t^{-\frac{1}{4}}
    \end{cases}
\end{equation}

While we have observed an improved convergence speed of EPS over PS, a systematic study is required to properly incorporate weight norm control into the training schedule. In particular, EPS schedule we propose rests on the assumption that parameter norms should
not be scaled during long training runs, which is not guaranteed to be the optimal choice.   

\subsubsection{\texorpdfstring{Maximal Update Parametrization ($\mu P$)}{Maximal Update Parametrization (muP)} with Tunable Multipliers}
\label{subsubsec:enh_mup}
\begin{table}[h!]
    \centering
    \begin{adjustbox}{max width=0.9\textwidth}
    \begin{tabular}{cccc}
        \toprule
        \makecell{{\small Architecture}\\{\small part}} & {\small Forward pass} & {\small Shapes} & {\small Multiplier scaling} \\
        \midrule
        \makecell{{\small Full} \\ {\small model}} 
        & {\small $\textcolor{purple}{m_\mathrm{unemb}}W_\mathrm{unemb}\mathcal{N}^f((\mathcal{F}_L(\ldots\mathcal{F}_1(\textcolor{purple}{m_\mathrm{emb}}W_\mathrm{emb}X)\ldots)))$} 
        & \makecell{{\small $W_\mathrm{emb}\in\mathbb{R}^{d \times d_\mathrm{voc}}$}\\{\small$W_\mathrm{unemb}\in\mathbb{R}^{d_\mathrm{voc}\times d }$}} 
        & \makecell{{\small $\textcolor{purple}{m_\mathrm{emb}}\propto 1$}\\{\small $\textcolor{purple}{m_\mathrm{unemb}}\propto d ^{-1}$}}\\
        \midrule
        \makecell{{\small Hybrid} \\ {\small block}} 
        & \makecell{{\small $\mathcal{F}_l(\mathbf{r}_l)=\mathbf{r}_l'+\mathcal{F}_l^{\mathrm{MLP}}(\mathcal{N}_l'(\mathbf{r}_l'))$}\\{\small $\mathbf{r}_l'=\mathbf{r}_l+\mathcal{F}_l^{\mathrm{attn}}(\mathcal{N}_l(\mathbf{r}_l))+\mathcal{F}_l^{\mathrm{SSM}}(\mathcal{N}_l(\mathbf{r}_l))$}} 
        & {\small $\mathbf{r}_l,\mathbf{r}_l'\in \mathbb{R}^{d\times L_\mathrm{seq}}$} 
        &  \\
        \midrule
        \makecell{{\small SSM}\\{\scriptsize(Mamba2)}\\{\small block}} 
        & \makecell{
        {\small $\widetilde{\mathbf{x}}=\textcolor{purple}{m_x}W_x \mathbf{r}\quad \widetilde{\mathbf{B}}=\textcolor{purple}{m_B}W_B \mathbf{r}\quad\widetilde{\mathbf{C}}=\textcolor{purple}{m_C}W_C\mathbf{r}$} \\
        {\small $\mathbf{xBC}=\operatorname{SiLU(\mathrm{conv1d}(\widetilde{\mathbf{x}}\widetilde{\mathbf{B}}\widetilde{\mathbf{C}}))}$} \\
        {\small $\widetilde{\mathbf{z}}=\textcolor{purple}{m_z}W_z \mathbf{r} \quad \mathbf{dt}=\operatorname{Softplus}(\textcolor{purple}{m_{dt}}W_{dt} \mathbf{r}+b_{dt})$} \\
        {\small $\mathbf{y}_\mathrm{SSM} = \operatorname{SSM}(\mathbf{x},\mathbf{B},\mathbf{C},\mathbf{dt})\odot\operatorname{SiLU}(\widetilde{\mathbf{z}})$} \\
        {\small $\mathcal{F}^{\mathrm{SSM}}(\mathbf{r})=\textcolor{purple}{m_{\mathrm{SSM}}}W_\mathrm{SSM}\mathcal{N}^\mathrm{SSM}(\mathbf{y}_\mathrm{SSM})$}} 
        & \makecell{{\small $W_x\in\mathbb{R}^{d_h^{\mathrm{ssm}} n_h^{\mathrm{ssm}}\times d}$}\\{\small $W_z\in\mathbb{R}^{d_h^{\mathrm{ssm}} n_h^{\mathrm{ssm}}\times d}$}\\{\small $W_B\in\mathbb{R}^{d_\mathrm{state}^{\mathrm{ssm}} n_g^{\mathrm{ssm}}\times d}$}\\{\small $W_C\in\mathbb{R}^{d_\mathrm{state}^{\mathrm{ssm}} n_g^{\mathrm{ssm}}\times d}$}\\{\small $W_{dt}\in\mathbb{R}^{n_h^{\mathrm{ssm}}\times d}$}\\{\small $W_{\mathrm{SSM}}\in\mathbb{R}^{d\times d_h^{\mathrm{ssm}} n_h^{\mathrm{ssm}}}$}} 
        & \makecell{{\small $\textcolor{purple}{m_x}\propto d^{-1}$}\\{\small $\textcolor{purple}{m_z}\propto d^{-1}$}\\{\small $\textcolor{purple}{m_B}\propto (d_\mathrm{state}^{\mathrm{ssm}} n_g^{\mathrm{ssm}} d)^{-1}$}\\{\small $\textcolor{purple}{m_C}\propto d^{-1}$}\\{\small $\textcolor{purple}{m_{dt}}\propto d^{-1}$}\\{\small $\textcolor{purple}{m_\mathrm{SSM}}\propto (d_h^{\mathrm{ssm}} n_h^{\mathrm{ssm}})^{-1}$}} \\
        \midrule
        \makecell{{\small Attention}\\{\small block}} 
        & \makecell{{\small $\mathbf{Q}=W_Q \mathbf{r} \quad \mathbf{K}=\textcolor{purple}{m_\mathrm{key}} W_K \mathbf{r} \quad \mathbf{V}=W_V \mathbf{r}$}\\{\small $\mathcal{F}_l^{\mathrm{attn}}(\mathbf{r})=\textcolor{purple}{m_\mathrm{attn}}W_\mathrm{attn}\operatorname{GQA}(\mathbf{Q},\mathbf{K},\mathbf{V})$}} 
        & \makecell{{\small $W_\mathrm{attn}\in\mathbb{R}^{d\times d_h^{\mathrm{attn}} n_h^{\mathrm{attn}}}$} \\ {\small $W_Q\in\mathbb{R}^{d_h^{\mathrm{attn}} n_h^{\mathrm{attn}}\times d}$} \\ {\small $W_K\in\mathbb{R}^{d_h^{\mathrm{attn}} n_g^{\mathrm{attn}}\times d}$} \\ {\small $W_V\in\mathbb{R}^{d_h^{\mathrm{attn}} n_g^{\mathrm{attn}}\times d}$}} 
        & \makecell{{\small $\textcolor{purple}{m_\mathrm{attn}}\propto (d_h^{\mathrm{attn}} n_h^{\mathrm{attn}}d)^{-1}$} \\ {\small $\textcolor{purple}{m_\mathrm{key}}\propto d^{-2}(d_h^{\mathrm{attn}})^{-\frac{1}{2}}$}} \\
        \midrule
        \makecell{{\small MLP} \\ {\small block}} 
        & \makecell{{\small $\mathbf{y}_{\mathrm{MLP}}=\operatorname{SiLU}(\textcolor{purple}{m_\mathrm{gate}}W_\mathrm{gate}\mathbf{r})\odot(W_\mathrm{up}\mathbf{r})$}\\{\small $\mathcal{F}^\mathrm{MLP}(\mathbf{r})=\textcolor{purple}{m_\mathrm{MLP}}W_\mathrm{down}\mathbf{y}_{\mathrm{MLP}}$}} 
        & \makecell{{\small $W_\mathrm{up}\in\mathbb{R}^{d_\mathrm{MLP}\times d }$} \\ {\small $W_\mathrm{gate}\in\mathbb{R}^{d_\mathrm{MLP}\times d }$} \\ {\small $W_\mathrm{down}\in\mathbb{R}^{d \times d_\mathrm{MLP}}$}} 
        & \makecell{{\small $\textcolor{purple}{m_\mathrm{MLP}}\propto (d_\mathrm{MLP}d)^{-1}$} \\ {\small $\textcolor{purple}{m_\mathrm{gate}}\propto d^{-1}$}}
        \\
        \bottomrule
    \end{tabular}
    \end{adjustbox}
    \caption{This table summarizes the location of all the forward $\mu$P multipliers used in Falcon-H1 models, the rules to scale multipliers with model size, and the shapes of the relevant parameters or activations required for the scaling. Specific forms of scaling rule in the last column can be straightforwardly derived from $\mu$P Desideratum, see, for example, ~\citep{yang2024spectralconditionfeaturelearning}. $\operatorname{GQA}(\mathbf{Q},\mathbf{K},\mathbf{V})$ is Grouped Query Attention; $\operatorname{SSM}(\mathbf{x},\mathbf{B},\mathbf{C},\mathbf{dt})$ is recurrent sequence transformation described in section \ref{sec:ssm_params}; $\mathcal{N},\mathcal{N}',\mathcal{N}^{\mathrm{SSM}},\mathcal{N}^{\mathrm{f}}$ are RMS normalization layers.}
    \label{tab:forward_multipliers}
\end{table}

\paragraph{Background.}
Maximal update parametrization ($\mu P$) was proposed in~\citep{yang2022featurelearninginfinitewidthneural}. It combines several ideas of how different factors affect the ability of the model to learn non-trivial internal representations as the model's width $d$ increases. One such factor is parameterizing the weight matrix $W$ with a scalar multiplier $m$, so that the forward pass becomes $\mathbf{y}=mW\mathbf{x}$. Another is initialization variance and learning rate of different layers in the model's architecture. The impact of these factors is formulated rigorously in the infinite width limit $d\to\infty$, where a unique scaling of main parameters with $d$ is required to ensure nontrivial feature learning. The multiplier $m$, initialization variance $\sigma^2$, learning rate $\eta$, and weight decay $\lambda$ must scale with a certain powers of width $d$:
\begin{equation}\label{eq:mup_scaling_general}
    m = m_0 d^{-a}, \quad \sigma=\sigma_0 d^{-b}, \quad \eta=\eta_0 d^{-c}, \quad \lambda=\lambda_0 d^{-e},
\end{equation}
with specific values of scaling exponents $(a,b,c,e)$ depending on the location of the layer $W$ in the model architecture and the type of optimizer used, e.g. SGD or Adam. For example, for AdamW optimizer, $\mu$P scaling of hidden layers are $(a,b,c,e)=(0,\frac{1}{2},1,-1)$ while for output(LM head) layer $(a,b,c,e)=(0, 1, 1, -1)$. The subsequent work~\citep{yang2023tensorprogramsvifeature,bordelon2023depthwisehyperparametertransferresidual} also explores the scaling with model depth $L$ to enable feature learning in the limit $L\to\infty$. 

The main practical implication of $\mu P$ is zero-shot hyperparameter (HP) transfer (\emph{$\mu$Transfer})~\citep{yang2022tensorprogramsvtuning}. Essentially, if one finds the optimal hyperparameters at a reference, typically small, model size $d_\mathrm{ref}$, the optimal HPs at larger target size $d$ could be roughly obtained from reference model values by using scaling rules \eqref{eq:mup_scaling_general}. $\mu$Transfer were previously applied for tuning and transferring LLM HPs in~\citep{yang2022tensorprogramsvtuning,dey2023cerebrasgptopencomputeoptimallanguage,hu2024minicpm,dey2025dontlazycompletepenables, liu2023llm360fullytransparentopensource}.      

An important aspect of model parametrization is the presence of an exact symmetry transformation that rescales $(m,\sigma,\eta,\lambda)$ such that both the forward pass and the optimizer update stay unchanged. Specifically, for AdamW optimizer (with $\varepsilon=0$), such scaling  transformation with a parameter $p$ is
\begin{equation}\label{eq:scaling_symmetry}
    m\rightarrow p^{-1}m, \quad  \sigma \rightarrow p\sigma, \quad \eta \rightarrow p \eta, \quad \lambda \rightarrow p^{-1}\lambda.   
\end{equation}
This symmetry has a practical implication for $\mu$Transfer as it allows to remove in \eqref{eq:mup_scaling_general} the scaling of either learning rate or forward multiplier (by choosing $p=(d/d_\mathrm{ref})^{-a}$ or $p=(d/d_\mathrm{ref})^c$). The motivation to remove one of the scalings comes from training infrastructure: multiplier $m$ must be implemented directly in the forward pass of the model, while scaling of $\eta,\lambda$ is typically implemented via optimizer parameter groups. 

For Falcon-H1 models, we relocate the $\mu$P scaling \eqref{eq:mup_scaling_general} from learning rate/weight decay to forward multipliers. For example, the original scaling $(a,b,c,e)=(0,\frac{1}{2},1,-1)$ for the hidden layers becomes $(a,b,c,e)=(1,-\frac{1}{2},0,0)$. As a result, all the models in the series can be fine-tuned or continuously pretrained with the same learning rate and weight decay parameters.

\begin{table}[h!]
\centering
\renewcommand{\arraystretch}{1.2}
\begin{tabular}{
    l !{\vrule width 1.5pt}
    r|l!{\vrule width 1.5pt}
    r|l!{\vrule width 1.5pt}
    r|l!{\vrule width 1.5pt}
    r|l
}
\makecell{{Base model}\\{sizes}} & 
\multicolumn{2}{c!{\vrule width 1.5pt}}{\makecell{{Forward}\\{multipliers}}} &
\multicolumn{2}{c!{\vrule width 1.5pt}}{\makecell{{Matrix LR}\\{multipliers}}} &
\multicolumn{2}{c!{\vrule width 1.5pt}}{\makecell{{Matrix WD}\\{multipliers}}} &
\multicolumn{2}{c}{\makecell{{Vector LR}\\{multipliers}}} \\
\toprule
$L=66$ & $m_\mathrm{emb}$ & $2^{2.5}$ & $W_\mathrm{emb}$ & $2^{2}$ & $W_\mathrm{emb}$ & $2^{-3}$ & $\mathcal{N}^f$ & $2^{1.5}$ \\
$d=1280$ & $m_\mathrm{unemb}$ & $2^{-5}$ & $W_\mathrm{unemb}$ & $2^{0}$ & $W_\mathrm{unemb}$ & $2^{-2}$ & $\mathcal{N}^\mathrm{Mixer}$ & $2^{2}$ \\
$d^\mathrm{ssm}_h=64$  & $m_\mathrm{MLP}$ & $2^{-2}$ & $W_\mathrm{in}$ & $2^{-0.5}$ & $W_\mathrm{in}$ & $2^{0.5}$ & $\mathcal{N}^\mathrm{MLP}$ & $2^{1.5}$ \\
$n^\mathrm{ssm}_h=16$ & $m_\mathrm{attn}$ & $2^{-1}$ & $W_\mathrm{out}$ & $2^{-2}$ & $W_\mathrm{out}$ & $2^{2}$ & $\mathcal{N}^\mathrm{SSM}$ & $2^{1}$ \\
$d^\mathrm{ssm}_\mathrm{state}=128$ & $m_\mathrm{SSM}$ & $2^{-1.5}$ & $W_\mathrm{up}$ & $2^{-0.5}$ & $W_\mathrm{up}$ & $2^{-0.5}$ & $W_\mathrm{conv1d}$ & $2^{2.5}$ \\
$d^\mathrm{attn}_h=64$ & $m_\mathrm{gate}$ & $2^{-0.5}$ & $W_\mathrm{gate}$ & $2^{0.5}$ & $W_\mathrm{gate}$ & $2^{0}$ & $b_\mathrm{conv1d}$ & $2^{1}$ \\
$n^\mathrm{attn}_h=12$ & $m_\mathrm{key}$ & $2^{-2}$ & $W_\mathrm{down}$ & $2^{-0.5}$ & $W_\mathrm{down}$ & $2^{-0.5}$ & $b_{dt}$ & $2^{1.5}$ \\
$d_\mathrm{MLP}=3840$ & $m_x$ & $2^{-2}$ &  &  &  &  & $A_\mathrm{log}$ & $2^{1.5}$ \\
& $m_z$ & $2^{-1.5}$  &  &  &  &  & $D$ & $2^{3}$ \\
& $m_B$ & $2^{-1.5}$ &  &  &  &  &  & \\
& $m_C$& $2^{-1}$ &  &  &  &  &  & \\
& $m_{dt}$ & $2^{-1.5}$ &  &  &  &  &  &  \\
\end{tabular}
\caption{Base model shapes and its $\mu$P multipliers after tuning procedure described in section \ref{sec:mup_tuning}. The notations for forward multipliers, as well as notations for most of the learnable parameters, were introduced in table \ref{tab:forward_multipliers}. $W_\mathrm{in}=W_{xzBCdt|QKV}$ is a merged weight matrix combining all the initial projections of both SSM and attention mixers, and $W_\mathrm{out}=W_{\mathrm{SSM}|\mathrm{attn}}$ is similarly merged weight matrix that combines output projections of SSM and attention. We have merged these matrices in the context of $\mu$P multipliers to reduce the total number of multipliers to be tuned. Finally, $D$ and $A_\mathrm{log}$ are learnable parameters of Mamba2 sequence transformation.}
\label{tab:tuned_multipliers}
\end{table}

\paragraph{Our approach.} Our core idea is to augment $\mu$Transfer scaling for transferring HPs across model sizes with tuning the $\mu$P multipliers at the base model size. 

To motivate the approach, let us rewrite the general scaling \eqref{eq:mup_scaling_general} for a model parameter $W^{(i)}$ from the point of view of the base model with width $d_\mathrm{ref}$
\begin{equation}
    m^{(i)} = m_\mathrm{ref}^{(i)} \left(\frac{d}{d_\mathrm{ref}}\right)^{-a}, \; \sigma^{(i)} = \sigma_\mathrm{ref}^{(i)} \left(\frac{d}{d_\mathrm{ref}}\right)^{-b}, \; \eta^{(i)} = \eta_\mathrm{ref}^{(i)} \left(\frac{d}{d_\mathrm{ref}}\right)^{-c}, \; \lambda^{(i)} = \lambda_\mathrm{ref}^{(i)} \left(\frac{d}{d_\mathrm{ref}}\right)^{-e}.
\end{equation}
The classical $\mu$Transfer uses the Standard parametrization at the base model size, corresponding to no forward multipliers $m_\mathrm{ref}^{(i)}=1$ and use of global LR $\eta$ and WD $\lambda$ for all the layers, $\eta_\mathrm{ref}^{(i)}=\eta$ and $\lambda_\mathrm{ref}^{(i)}=\lambda$. However, this strategy is somewhat contradictory: it uses specialized HPs at target size $d>d_\mathrm{ref}$ that respect the limiting $d\to\infty$ feature learning scaling, while using global HPs for the base model as if $d_\mathrm{ref}$ were tiny and base model was very far from limiting $d\to\infty$ behavior. This assumption does not look reasonable in practice. For Falcon-H1 series, we have used the base model with $d_\mathrm{ref}=1280$ while the largest 34B model has $d=5120$, just an $\times 4$ factor from the base model. 

With such a premise, we have decided to individually tune HPs of the different layers to respect the limiting $d\to\infty$ tendency that could be affecting the model at $d_\mathrm{ref}$. First, we group model parameters according to their role in the architecture, e.g. LM head or MLP down projection, to tune LR and WD multipliers $\eta_\mathrm{ref}^{(i)}/\eta$ and $\lambda_\mathrm{ref}^{(i)}/\lambda$. Second, we introduce a minimal set of forward multipliers that produce all possible transformations of the activations throughout the whole forward pass of Falcon-H1 architecture\footnote{Formally, to have a fully complete set of forward multipliers we would also need to add multipliers to SSM inputs $\widetilde{\mathbf{B}}$ and $\widetilde{\mathbf{dt}}$. However, these multipliers are effectively taken care of by learnable parameters of SSM $D$ and $A_\mathrm{log}$ that can adapt to the right scale during training, unlike matrix parameters with non-adaptive norms as described in section \ref{sec:param_norms_and_schedules}.}. The minimal property of our set removes redundant multipliers that otherwise would be tuned in vain. For example, having both key and query multipliers $\mathbf{K}\to m_K \mathbf{K}$ and $\mathbf{Q}\to m_Q \mathbf{Q}$ is redundant because only their scalar product $\mathbf{Q}^\top\mathbf{K}\to m_Q m_K \mathbf{Q}^\top\mathbf{K}$ is used in attention and having only $m_K$ is sufficient to cover all possible scalings of attention scores. Finally, we don't tune initialization variances $\sigma^{(i)}$ because the symmetry \eqref{eq:scaling_symmetry} reduces one degree of freedom in the HP set $(m,\sigma,\eta,\lambda)$, and our preliminary sweeps on initialization variances have shown weak dependence of the loss on $\sigma^{(i)}$. This resulted in 35 multipliers to be tuned for Falcon-H1 architecture.

\begin{figure}
    \centering
    \includegraphics[scale=0.35, clip, trim=10 10 10 10]{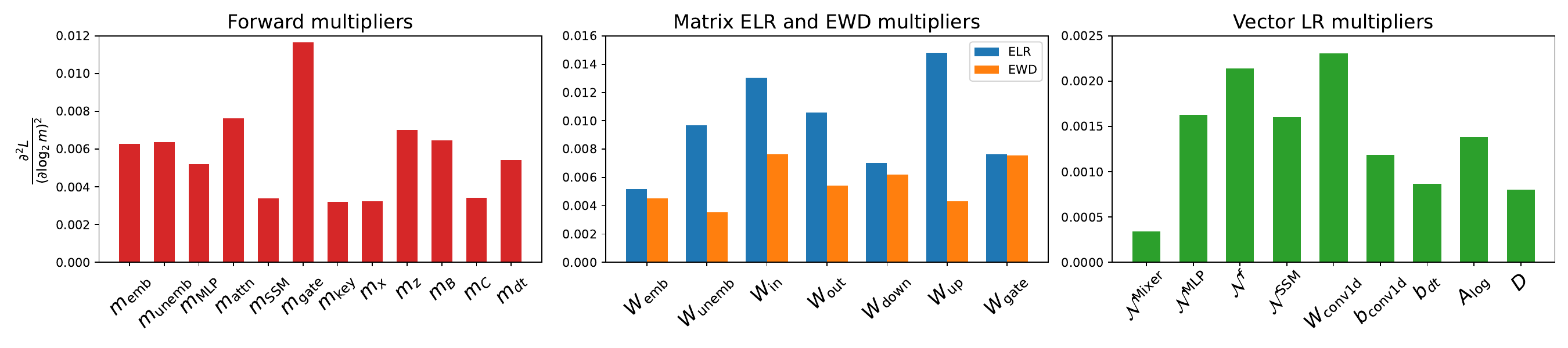}
    \caption{Sensitivity of the loss with respect to all 35 $\mu$P multipliers we have tuned. The multipliers are organized into 4 groups as in table \ref{tab:tuned_multipliers}, with matrix layers ELR and EWD multipliers are shown in the middle plot side-by-side for comparison. All 4 multiplier groups have different magnitudes of the effect on the loss, and the y-axis limits were adjusted accordingly. ELR multipliers have the strongest impact, followed by foraward multipliers, then EWD multipliers, and lastly LR multipliers of vector-like layers. Although the sensitivities w.r.t. to the last group are low, note that the final multiplier values are quite far from those of the matrix layers (see table \ref{tab:tuned_multipliers}), and, therefore, separating LRs of vector-like and matrix-like layers (as was done in our tuning) still yields significant performance improvement.   
    }
    \label{fig:mup_multipliers}
\end{figure}

The final values of tuned multipliers, as well as all base model shapes and global LR/WD, can be found in table \ref{tab:tuned_multipliers}, and the location of forward multipliers is detailed in table \ref{tab:forward_multipliers}. We describe our procedure to tune multipliers in section \ref{sec:mup_tuning}. It naturally leads to the estimation of the sensitivity of the loss on each multiplier around the optimum $\frac{\partial^2}{(\partial \log_2 m)^2}L$ that we display on figure \ref{fig:mup_multipliers} to provide intuition behind the impact of each of the 35 tuned multipliers.

\subsubsection{Other Aspects: Batch Scaling, Rampup, Warmup}
\begin{figure}[h!]
    \centering
    \includegraphics[scale=0.55, clip, trim=10 10 10 10]{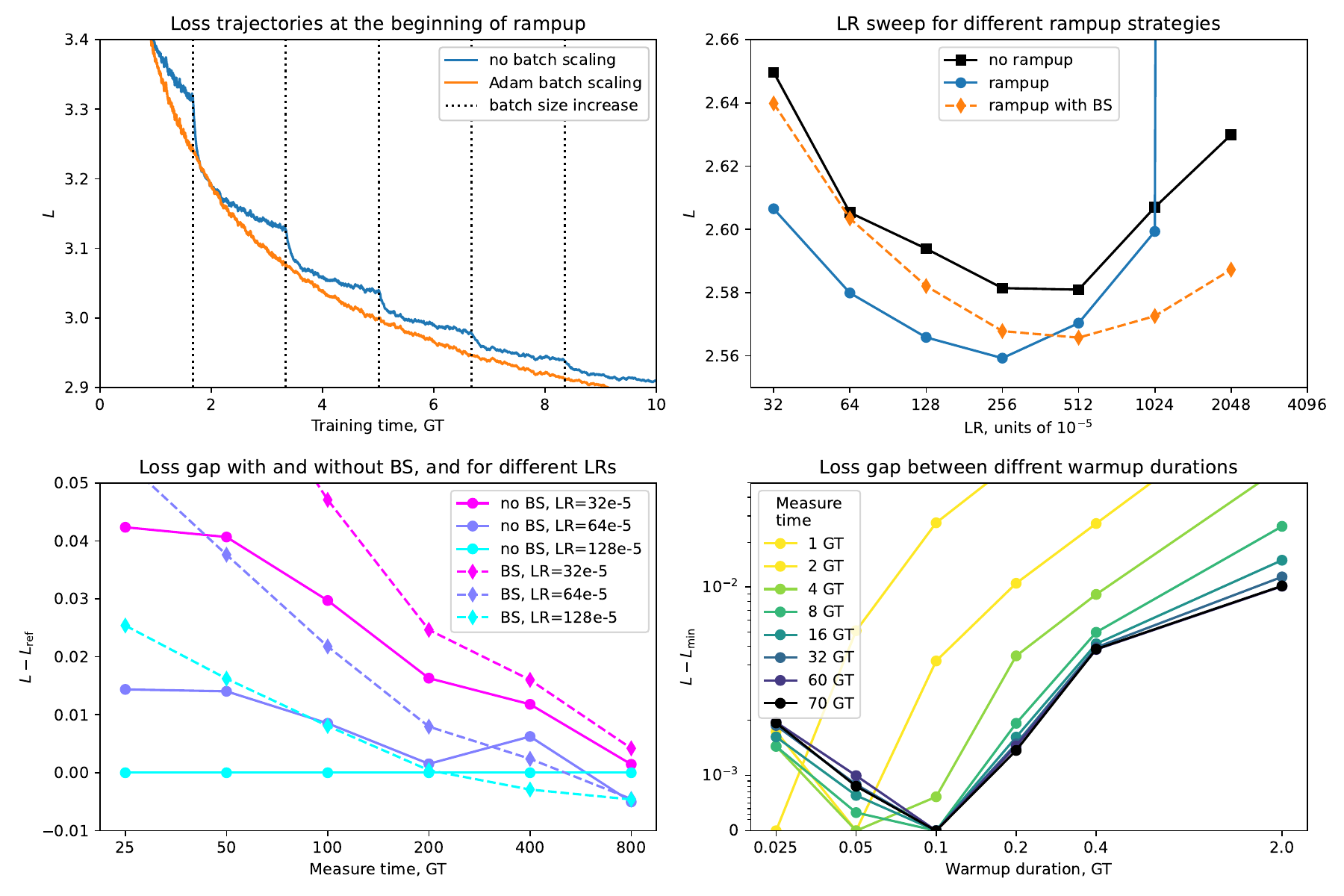}
    \caption{Rampup and warmup related observations on a 1.5B pure Mamba2 model. \textbf{(Top Left)} Train loss trajectories during rampup with and without batch scaling. No batch scaling run exhibits loss jumps at the moments of batch size increase. These jumps could be associated with a decrease in the noise level. \textbf{(Top Right)} Loss after learning rate decay at 70GT on a LR sweep for different rampup strategies. \textbf{(Bottom Left)} Evolution of the loss for runs with and without BS scaling and 3 different learning rates. To take into account the noise level, we measure the loss after LR decay, which is performed at several positions of the training trajectory. To be able to display losses at very distant moments of training, we subtract the loss of the reference run. We see that, for all considered learning rates, the runs with batch scaling have a better trend with longer training duration. \textbf{(Bottom Right)} Evolution of the loss gap between runs with different warmup durations. Unlike the previous plot, the loss is measured directly (without LR decay) because warmup duration does not seem to affect noise level, as can be seen from the last predecay measurement at 60GT and after decay measurement at 70GT being very close. We see that at early stages ($\lesssim 16$GT) loss vs warmup duration curve shifts with measurement time, reflecting that short warmup runs simply had more high LR steps at the beginning of the training. However, at later stages ($\gtrsim 16$GT) the curve stabilizes, suggesting a well-defined optimal warmup duration with long-lasting effect on the training.}
    \label{fig:rampup_warmup}
\end{figure}
In this section, we briefly report interesting observations related to training dynamics that we have incorporated into Falcon-H1 training.  
\paragraph{Batch scaling.} First, following~\citep{zuo2024falconmamba} we have used batch scaling that scales the learning rate if the batch size is changed  
\begin{equation}\label{eq:batch_scaling}
    \eta(b) = \eta_\mathrm{ref}\sqrt{\frac{b}{b_\mathrm{ref}}},
\end{equation}
where square root is a suitable scaling for Adam optimizer~\citep{malladi2022on}. We have observed that scaling \eqref{eq:batch_scaling} better preservers optimal learning than no batch scaling at all. However, more careful studies are required for robust transfer of HPs with batch size, taking into account, for example, scaling of parameter norms with batch size to combine it with ELR/EWD picture discussed in section \ref{sec:param_norms_and_schedules}.

\paragraph{Rampup.} At the beginning of the training, we have used the batch size rampup that linearly increases batch size over the specified duration, which we set around $50$GT for different Falcon-H1 model sizes. 

In figure \ref{fig:rampup_warmup} we considered 3 strategies and their impact on the training loss: no rampup, rampup without batch scaling, rampup with batch scaling \eqref{eq:batch_scaling}. No rampup was observed to have the worst loss, while also amplifying training instabilities. Shortly after the rampup period, batch scaling showed worse loss than classical rampup without batch scaling, see figure \ref{fig:rampup_warmup} (top right). However, the gap between them closes at long training durations, and rampup batch scaling eventually outperforms its no batch scaling version, see figure \ref{fig:rampup_warmup} (bottom left). This later trend is especially interesting because BS and no BS runs have exactly the same hyperparameters after the ramp-up period. One interpretation would be that batch scaling during rampup directs the training trajectory to a better region of parameter space, and the model continues to learn in this region for a very long time.

\paragraph{Warmup.} Another classical technique used at the beginning of training is to linearly increase LR to the target value in the earliest stages of the training, typically for a few gigatokens. In figure \ref{fig:rampup_warmup} (bottom right) we have tested the impact of warmup duration on the loss at a later training stage. Similarly to the rampup batch scaling, we observe that warmup duration has a long-lasting impact on the loss, with a relatively short optimal duration of $0.1$GT.

\subsection{Pretraining Infrastructure}
Pretraining was conducted using \textit{Mambatron}, our in-house distributed training framework. This framework is an evolution of \textit{Gigatron}, the codebase previously used to train the Falcon model series, and has been specifically optimized to support the unique hybrid architecture of Falcon-H1.
Building upon a foundation of standard 3-dimensional parallelism (3DP), we introduced two key innovations tailored for this hybrid design. First, we redesigned Context Parallelism (CP) to efficiently manage and scale the long sequence lengths inherent to the hybrid attention-SSM architecture. Second, we developed a novel strategy termed Mixer Parallelism (MP), which is specifically designed to parallelize computations across the distinct attention and SSM heads. This approach significantly accelerates both training and inference throughput.
The Falcon-H1 model series was trained on a large-scale infrastructure comprising 4,096 NVIDIA H100 GPUs. To optimize resource utilization, we employed a dynamic node allocation strategy, enabling the simultaneous training of six Falcon-H1 models. 
The parallelism configurations for each model are detailed in Table~\ref{tab:parallelism_config}.

\begin{table}[!htbp]
\centering
\begin{adjustbox}{max width=0.8\textwidth}
\begin{tabular}{lccccccc}
\toprule
Models & Batch Size & Context Len. Stage & DP & TP & PP & CP & MP \\
\midrule
\multirow{1}{*}{Falcon-H1-0.5B}
  & \multirow{1}{*}{4M} & 4K, 16K  & 64  & 1 & 1 & 1 & \purplecross \\
\midrule
\multirow{2}{*}{Falcon-H1-1.5B (1.5B‑Deep)}
  & \multirow{2}{*}{4M} & 16K, 32K   & 256 & 1 & 1 & 1 & \purplecross \\
  &                     & 131K  & 64  & 1  & 1  & 4  &     \purplecross          \\
\midrule
\multirow{2}{*}{Falcon-H1-3B}
  & \multirow{2}{*}{8M} & 16K, 32K   & 256 & 1 & 1 & 1 & \purplecross \\
  &                     & 131K  & 64  & 1  & 1  & 4  &     \purplecross        \\
\midrule
\multirow{3}{*}{Falcon-H1-7B}
  & \multirow{3}{*}{8M} & 16K, 32K   & 256 & 2 & 1 & 1 & \purplecheck   \\
  &                     & 131K  &   128  & 2  & 1  & 4  &    \purplecheck          \\
  &                     & 262K  &   64  & 2  & 1  & 8  &      \purplecheck        \\
\midrule
\multirow{4}{*}{Falcon-H1-34B}
  & \multirow{4}{*}{26M} & 16K   & 448 & 4 & 2 & 1 & \purplecheck   \\
  &                      & 32K  & 192    &  4 & 2  & 2  &    \purplecheck            \\
  &                      & 131K  & 48    &  4 & 2  & 8  & \purplecheck \\
  &                      & 262K  & 24    &  4 & 2  & 16  &    \purplecheck          \\
\bottomrule
\end{tabular}
\end{adjustbox}
\caption{5D Parallelism Configurations for Falcon-H1's Training}
\label{tab:parallelism_config}
\end{table}

\subsubsection{Scaling Dynamics of Data Parallelism}

While Data Parallelism (DP) is a fundamental technique for distributed training, its throughput scaling is not without limits. When scaling the number of DP workers ($N_{\text{DP}}$) while keeping the global batch size ($B_{\text{g}}$) constant, throughput gains diminish significantly once communication overhead outweighs computation. 
This occurs because, to maintain a fixed $B_{\text{g}}$, the number of gradient accumulation steps ($K$) per optimizer update must decrease inversely with $N_{\text{DP}}$, where $B_{\text{g}} = N_{\text{DP}} \cdot K \cdot B_{\mu}$ for a micro-batch size $B_{\mu}$.

To formally analyze this behavior, we model the time for a single optimizer step as:
\[
T_{\text{step}}(N_{\text{DP}}) \approx K \cdot t_{\mu} + t_{\text{sync}}(N_{\text{DP}})
\tag{1}
\]
Here, $t_{\mu}$ is the constant time for a forward-backward pass on a single micro-batch, and $t_{\text{sync}}$ is the latency of the gradient all-reduce operation, which can grow with $N_{\text{DP}}$ due to network complexity. As $N_{\text{DP}}$ increases, the computation term ($K \cdot t_{\mu}$) shrinks, while the communication term ($t_{\text{sync}}$) becomes the dominant component.
Consequently, the overall throughput, given by $B_{\text{g}} / T_{\text{step}}$, deviates from ideal linear scaling. Substituting the terms, we get:
\[
\text{Throughput}(N_{\text{DP}}) \approx \frac{B_{\text{g}}}{\frac{B_{\text{g}}}{N_{\text{DP}}B_{\mu}} \cdot t_{\mu} + t_{\text{sync}}(N_{\text{DP}})}
\tag{2}
\]
Theoretically, linear scaling could be maintained by increasing $B_{\text{g}}$ proportionally with $N_{\text{DP}}$, keeping $K$ constant. However, significantly changing the global batch size from the value determined during hyperparameter tuning can destabilize training dynamics, thus impacting model convergence and final performance.
Therefore, our strategy involves a pragmatic trade-off. We cap the DP size at a value where communication overhead remains manageable, and increase the global batch size only up to a critical point that balances high throughput with stable model convergence. This ensures efficient hardware utilization without compromising the integrity of the training regime.

\subsubsection{Mixer Parallelism (MP)}
\label{subsubsec:mixer_parallelism}
During pre-training, we developed a novel distributed training paradigm to boost the efficiency of training our largest models. Leveraging the parallel architecture of our decoder layers, where attention and Mamba layers are executed sequentially, we partitioned the Tensor Parallel (TP) world into two distinct groups: one dedicated to Mamba operations, the other to attention operations. This design allows these computations to run concurrently, followed by an all-reduce operation to synchronize their outputs.
We refer to this strategy as \textit{Mixer Parallelism} (MP). It significantly improves training throughput by optimizing both computational speed and memory efficiency while also boosting inference efficiency particularly for scenarios involving small batch sizes and sequence lengths.

\begin{figure}[ht] 
    \centering 
    \includegraphics[width=1.0\textwidth]{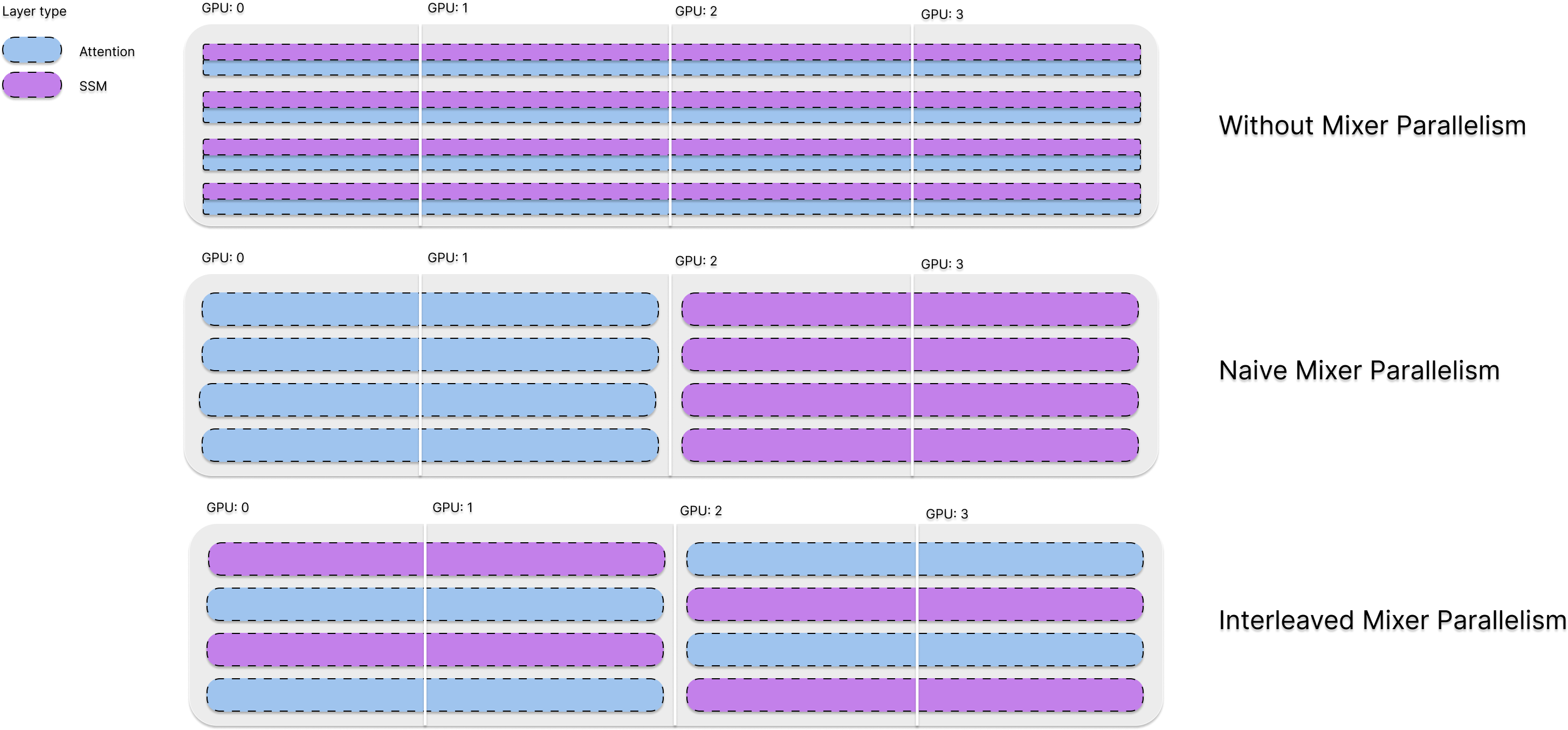} 
    \caption{Diagram illustrating the Mixer Parallelism (MP) strategies. Each row represents a full decoder layer. Output projection and all reduce operations are applied at the end of each layer. \textit{(Top)} Without MP, all GPUs compute both mixer types sequentially. \textit{(Middle)} Naive MP statically assigns GPUs to one mixer type for all layers. \textit{(Bottom)} Interleaved MP alternates assignments per layer, achieving better load balancing.}
    \label{fig:mp-diag} 
\end{figure}

Mixer Parallelism can be implemented using two distinct approaches. In \textit{naive} Mixer Parallelism, predefined TP groups are assigned exclusively to a single mixer type (Attention or Mamba).  In contrast, \textit{interleaved} Mixer Parallelism distributes different mixer types across TP groups in an alternating fashion, achieving a more balanced distribution of computational overhead from the slower mixer layers. Figure~\ref{fig:mp-diag} shows how different Mixer Parallelisms are implemented.

\noindent \textbf{Training Efficiency with MP.} To evaluate the efficacy of Mixer Parallelism for the model's training, we conducted experiments using a 2B hybrid model, configured with a data parallelism of 4, a tensor parallelism of 4, and a context length of 2048. We measured the training throughput for a baseline without MP against both the naive and interleaved variants.


The results, summarized in Table~\ref{tab:mp_optimized}, demonstrate the clear superiority of the interleaved MP strategy. By effectively balancing the computational load, interleaved Mixer Parallelism achieves a substantial 1.43x speedup over the baseline.

\begin{table}[ht!]
\centering
\begin{tabular}{l S[table-format=1.4] S[table-format=1.2]}
\toprule
\textbf{MP Variant} & {\textbf{Throughput (Gtok/hr)}} & {\textbf{Speedup (ratio)}} \\
\midrule
None (Baseline)     & 0.2339  & 1.00 \\
Naive MP              & 0.2640  & 1.13 \\ 
Interleaved MP        & \textbf{0.3343}  & \textbf{1.43} \\
\bottomrule
\end{tabular}
\caption{Training throughput and speedup comparison for different Mixer Parallelism variants.}
\label{tab:mp_optimized}
\end{table}




\noindent \textbf{Inference Efficiency with MP.} To evaluate the efficacy of Mixer Parallelism for inference scenarios, we conducted a comprehensive throughput analysis across a 3B and 7B parameters models. Based on its superior load-balancing properties, we focused exclusively on the \textit{interleaved} MP variant for these experiments. 
Our evaluation was performed on a single node equipped with two NVIDIA H100 GPUs, configured with a Tensor Parallelism size of 2. We systematically varied two key parameters to simulate diverse workloads: the batch size (from 1 to 128) and the number of generated output tokens (from 4096 to 32768). A constant prefill size of 8 tokens was used for all runs.



\begin{figure}[h!] 
    \begin{subfigure}[b]{0.95\textwidth}
        \includegraphics[width=\linewidth]{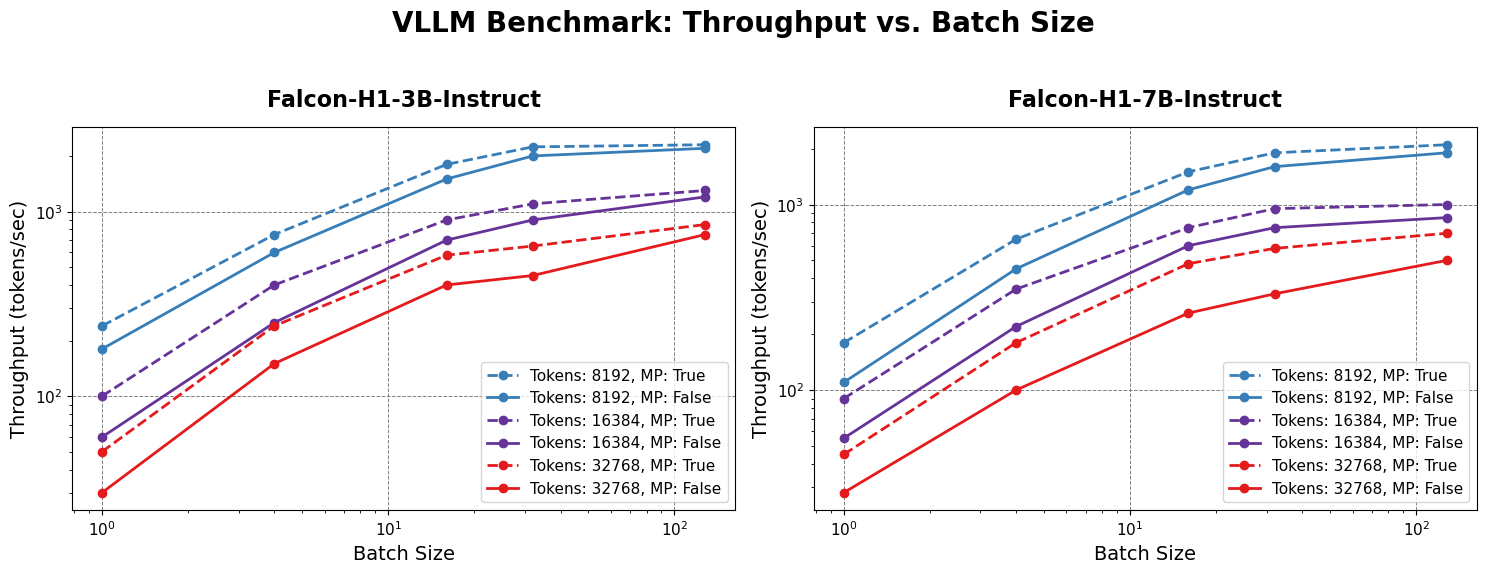}
        \label{fig:mp-7b}
    \end{subfigure}
    
    \caption{Throughput comparison for Mixer Parallelism across different model sizes (3B and 7B).} 
    \label{fig:multi-mp-vertical-3} 
    
\end{figure}

The implementation was integrated into a custom fork of the vLLM library~\citep{kwon2023efficient}. 
We report in figure \ref{fig:multi-mp-vertical-3} throughput results of this experiment on all our model sizes. The dashed curves represent configurations using Mixer Parallelism, while matching colors indicate experiments with the same number of generated tokens.

The results show that Mixer parallelism significantly accelerates inference for low-latency scenarios characterized by small batch sizes and short generated sequences. Though this advantage diminishes and reverses for larger batches and longer generation sequences.
This observation is consistent across all model sizes. We provide the community an easy way to test the implementation of Mixer parallelism in a vLLM fork ~\footnote{https://tiiuae.github.io/Falcon-H1/deployment/}. 


\subsubsection{Context Parallelism (CP)}
\label{sec:cp}

To keep GPU memory flat while scaling to long contexts, the model shards
each sequence horizontally across the \texttt{cp\_world} of $J$
devices and lets every device work on one contiguous \emph{chunk} of
length~$Q$.

\paragraph{Self‑Attention chunks.}
For the attention blocks we reuse \textbf{RingAttention}
\citep{liu2023ringattentionblockwisetransformers}: each rank holds its local
query–key–value slice and circulates \texttt{K/V} tensors around the
ring so that the full score matrix is produced without ever materialising
the whole sequence on one GPU.  Memory is therefore
$O(Q)$ per rank instead of $O(T)$ for the full length~$T$.

\paragraph{SSM (Mamba‑2) chunks.}
For the SSM layers we follow the chunk‑wise \textbf{state‑passing
schedule} described in Section 8.2 of Mamba‑2
\citep{dao2024transformers} and illustrated in their Figure 7
(right).  In words:

\begin{enumerate}
  \item \textbf{Initial state.}  
        Rank~$j$ waits for the final hidden state
        produced by rank~$j-1$ (rank 0 starts from zeros or an optional
        user‑supplied context state).
  \item \textbf{Local work.}  
        Using that state and its own input slice~$\mathbf{x}_{j}$,
        the rank runs the SSM kernel on its private chunk and produces the chunk’s output tokens~$\mathbf{y}_{j}$ and the \emph{next} hidden state~$h_{j+1}$. The computation is entirely local and no cross‑rank dependency
        exists.
  \item \textbf{State hand‑off.}  
        If a following rank exists, $h_{j+1}$ is sent
        asynchronously to rank $j+1$; otherwise the pipeline finishes.
\end{enumerate}

The only communication is a single tensor of shape
\texttt{[B $\times$ H $\times$ d\_state]} per boundary—%
where \(B\) is the micro‑batch size on the rank, \(H\) the number of
SSM heads, and
\(d_{\text{state}}\) the width of each state vector—so bandwidth
remains \emph{linear} in the number of GPUs.

\paragraph{CausalConv1D chunks.}
The preceding causal convolution stage is split in the same fashion:
each rank gets the last $k{-}1$ timesteps from its left neighbour,
performs its local depth‑wise convolution, and forwards the
$k{-}1$ boundary activations to the next rank.

\newpage
\section{Post-trainining}
\label{sec:post_training}
In this section, we detail the two-stage post-training procedure for our Falcon-H1 models, which consists of Supervised Fine-Tuning (SFT) followed by Offline Reinforcement Learning.  

\subsection{Post-training Data}
\label{ssec:post_training_data}

Compared to the previous Falcon model series, we have significantly expanded the scope and quality of the supervised fine-tuning data. This expansion spans multiple domains and includes millions of high-quality samples. Notably, our efforts focus on improving the model’s capabilities in complex reasoning, mathematical and scientific problem-solving, instruction-following, and function calling. Some of the key improvements include:

\begin{itemize}
    \item \textbf{Mathematical Problem-Solving:} We enhanced both the breadth and difficulty of mathematical problems by curating and rewriting high-quality solutions. Inspired by OpenMathInstruct-2~\citep{toshniwal2024openmathinstruct} and AceMath~\citep{liu2024acemath}, we synthesized math problems across diverse sub-domains, carefully filtering out incorrect solutions. Additional synthetic examples were generated using data from the pretraining math corpus, ensuring high correctness and varying difficulty levels.

    \item \textbf{Scientific Problem-Solving:} We improved the model’s performance in scientific reasoning, particularly across STEM domains. Problem-solution pairs were extracted or synthesized from existing pretraining corpora, with subsequent refinements to ensure consistent formatting and data quality.

    \item \textbf{Conversational and Instruction-Following:} We enhanced the model’s conversational abilities and instruction-following by improving personalization, stylistic variety and multi-round long conversation capability.
\end{itemize}

The post-training corpus is based on both license-permissive open datasets and proprietary data sources. From open data, we partially adopted and refined the datasets such as OpenMathInstruct-2~\citep{toshniwal2024openmathinstruct}, Tulu3~\citep{lambert2024tulu}, 
Smoltalk~\citep{benallal2024smollmcorpus}, and hermes-function-calling-v1~\citep{Hermes-Function-Calling-Dataset-V1}. In many cases, questions and solutions were further optimized to improve clarity and correctness. 
Proprietary data includes high-quality extracted, refined, and synthetic sample pairs from internal high-quality sources on both STEM and non-STEM domains, along with human-annotated examples targeting specific skills. This hybrid strategy ensures diverse, accurate, and skill-aligned training data. We find that data quality and structure have a much greater impact on post-training performance than data volume alone. We carefully decontaminated the post-training data against popular benchmarks to prevent unintended data leakage and ensure fair evaluation.

\subsection{Supervised Fine-Tuning (SFT)}
\label{ssec:sft}
We performed extensive data mixture and duration testing during SFT to balance the model's performance across different domains. Compared to pretraining, SFT is more sensitive to data mixture. The resulting checkpoints that we released offer strong general performance but can be further adjusted for specific use cases.

Our SFT process is divided into two distinct stages: a primary stage with a 16k context length, followed by a long-context stage extending to 128k. The initial 16k stage was conducted for 3 GT, succeeded by an additional 3GT for the long-context stage. During the total 6GT of SFT, we repeated different data sources for different amount of times (epochs) depending on their volume and weight in the mixture. The most repeated data source was Tulu3, with around 50\% weight in the data mixture and around 3.5 epochs over the SFT duration. The other data sources were repeated for fewer than 2 epochs. 

For the main 16k stage, we employed a Warmup-Stable-Decay (WSD) learning rate schedule. The decay phase followed an exponential profile, reducing the learning rate by a factor of eight to a minimum value of $\eta_{\min} = \eta/8$. The subsequent 128k long-context stage proceeded with a constant learning rate equal to this minimal value, $\eta_{\min}$. The 128k stage was omitted for our smallest 0.5B parameter model due to the inherent limitations in processing long sequences by models of that scale. The key hyperparameters for our SFT setup are summarized in Table~\ref{tab:sft_hyperparams}.

\begin{table}[h!]
\centering
\begin{tabular}{ll}
\hline
\textbf{Hyperparameter / Setting} & \textbf{Value / Details} \\
\hline
Sequence Length & 16k for the main stage; 128k for long context stage \\
Batch Size ($b$) & 1 million tokens (MT) \\
Learning Rate ($\eta$) & $128 \times 10^{-6}$ \\
AdamW Parameters & $\beta_1=0.9$; $\beta_2=0.95$; no weight decay \\
Learning Rate Schedule & WSD: 50MT warmup, 1.5GT stable, 1.5GT decay \\
LR Decay & Exponential schedule from $\eta$ to $\eta_{\min}=\eta/8$ \\
Long Context Stage & +3GT with constant learning rate $\eta_{\min}$ \\
Epochs per Data Source & $\lesssim 3.5$ \\
\hline
\end{tabular}
\caption{Hyperparameters for the Supervised Fine-Tuning (SFT) stage.}
\label{tab:sft_hyperparams}
\end{table}

Finally, we note that we have used slightly different batch sizes, in the range of 0.25MT to 4MT, for different model sizes to meet the desired GPU allocation on the cluster and the job run time\footnote{For instance, the 128k long-context stage, being computationally intensive due to context parallelism, necessitated larger batch sizes to enable higher degrees of data parallelism (DP) for quicker job execution.}. Then, for a given batch size $b$, we have used square root batch scaling \eqref{eq:batch_scaling}, taking the values from table \ref{tab:sft_hyperparams} as reference values $b_\mathrm{ref},\eta_\mathrm{ref}$. With such scaling, we have observed minimal impact of batch size on the final performance of the SFT model. 

\subsection{Direct Preference Optimization (DPO)}
\label{ssec:dpo}

For the DPO stage, we utilized a fork of the AllenAI open-instruct repository \footnote{https://github.com/allenai/open-instruct}. Similar to the SFT stage, our data mixture was built upon Tulu3~\citep{lambert2024tulu}, supplemented by several other open-source and in-house preference datasets. We employed the standard DPO loss function (\texttt{dpo\_loss\_type: dpo\_norm}). The hyperparameters for the DPO stage are outlined in Table~\ref{tab:dpo_hyperparams}.

\begin{table}[h!]
\centering
\begin{tabular}{ll}
\hline
\textbf{Hyperparameter / Setting} & \textbf{Value / Details} \\
\hline
Batch Size & 256 \\
Learning Rate & $5 \times 10^{-6}$ \\
Learning Rate Schedule & Linear decay to zero over 2 epochs, with a warmup ratio of 0.1 \\
DPO Loss Parameter ($\beta$) & 5 \\
AdamW Parameters & PyTorch default ($\beta_1=0.9$, $\beta_2=0.999$) \\
\hline
\end{tabular}
\caption{Hyperparameters for the Direct Preference Optimization (DPO) stage.}
\label{tab:dpo_hyperparams}
\end{table}

An important aspect of our DPO strategy was the stopping criterion. Instead of training for the full two epochs, at which point the learning rate schedule concludes, we found that stopping at approximately one epoch yielded superior results. This approach was empirically determined to be more effective than either completing the full two-epoch schedule or using a linear scheduler that terminates after a single epoch.

\section{Evaluation}
To provide a rigorous and reproducible comparison, we benchmarked the Falcon-H1 series against a comprehensive suite of leading models, including Qwen3~\citep{qwen3}, Qwen2.5~\citep{yang2024qwen2_5}, Gemma3~\citep{team2025gemma3}, Falcon3~\citep{Falcon3}, Llama3~\citep{grattafiori2024llama} and Llama4~\footnote{https://huggingface.co/meta-llama/Llama-4-Scout-17B-16E}. Our evaluation pipeline is built upon a foundation of established open-source frameworks: \texttt{lm-evaluation-harness}~\citep{eval-harness}, \texttt{evalchemy}~\citep{evalchemy}, \texttt{evalplus}~\citep{evalplus} and \texttt{helmet}~\citep{yen2025helmet}. 

\noindent\textbf{Standardization and Reproducibility.} Our methodology was standardized across all models to ensure a fair comparison. All evaluations were conducted within the same Docker environment to eliminate system-level variance. For the Qwen3 series, we disabled the "thinking mode" on all benchmarks to align its inference process with that of other models.

\noindent\textbf{Framework-Specific Settings.} To ensure stability, all \texttt{evalchemy} evaluations were pinned to a specific commit hash (\texttt{f735e77}). For relevant mathematical benchmarks within this framework, we standardized the number of generation turns to 16 and applied an identical system prompt across all models. Furthermore, all final math results were post-processed using Math-Verify~\citep{Kydlicek_Math-Verify_Math_Verification} for consistent verification. For all other frameworks, we adhered to their default settings to maintain comparability with established results in the literature.

\subsection{Base Models}
For base models, we report in Table \ref{tab:eval_setting_base} the benchmarks and settings used for the evaluations. Basically, we check the model's capabilities in five main domains: General, Math, Science, Code, and Multilingual. Noting that some benchmarks contains cross-domain tasks and do not fall neatly into a single category.

\begin{table}[htbp]
\centering
\footnotesize
\begin{tabularx}{\textwidth}{l >{\raggedright\arraybackslash}X l}
\toprule
\textbf{Benchmark} & \textbf{Settings} & \textbf{Framework}\\
\midrule
\multicolumn{3}{l}{\textbf{General}} \\
\quad BBH~\citep{suzgun2022challenging} & logprobs, 3-shot & \texttt{lm-eval-harness}\\
\quad ARC-C~\citep{clark2018think} & logprobs, 25-shot & \texttt{lm-eval-harness} \\
\quad HellaSwag~\citep{zellers2019hellaswag} & logprobs, 10-shot & \texttt{lm-eval-harness}\\
\quad Winogrande~\citep{sakaguchi2021winogrande} & logprobs, 5-shot & \texttt{lm-eval-harness}\\
\quad MMLU~\citep{hendrycks2020measuring} & logprobs, 5-shot & \texttt{lm-eval-harness}\\
\midrule
\multicolumn{3}{l}{\textbf{Math}} \\
\quad GSM8k~\citep{cobbe2021training} & strict match, 5-shot & \texttt{lm-eval-harness} \\
\quad MATH lvl5~\citep{hendrycks2021measuring} & math verify, logprobs, 4-shot & \texttt{lm-eval-harness}\\
\midrule
\multicolumn{3}{l}{\textbf{Science}} \\
\quad GPQA~\citep{rein2023gpqa} & logprobs, 5-shot & \texttt{lm-eval-harness}  \\
\quad MMLU-Pro~\citep{wang2024mmlu} & logprobs, 5-shot & \texttt{lm-eval-harness}  \\
\quad MMLU-stem~\citep{hendrycks2020measuring} & logprobs, 5-shot & \texttt{lm-eval-harness} \\
\midrule
\multicolumn{3}{l}{\textbf{Code}} \\
\quad HumanEval~\citep{chen2021codex}  & pass@1 & \texttt{evalplus} \\
\quad HumanEval+~\citep{evalplus} & pass@1 & \texttt{evalplus}\\
\quad MBPP~\citep{austin2021program} & pass@1 & \texttt{evalplus} \\
\quad MBPP+~\citep{evalplus} & pass@1 & \texttt{evalplus}\\
\midrule
\multicolumn{3}{l}{\textbf{Multilingual}} \\
\quad Multi-Hellaswag~\citep{dac2023okapi} & logprobs, 0-shot & \texttt{lm-eval-harness} \\
\quad MGSM~\citep{shi2022language} & flexible extract, 8-shot native CoT & \texttt{lm-eval-harness} \\
\quad Multi-MMLU~\citep{dac2023okapi} & logprobs, 5-shot & \texttt{lm-eval-harness} \\
\bottomrule
\end{tabularx}
\caption{Evaluation settings and benchmark sources for base models.}
\label{tab:eval_setting_base}
\end{table}

\noindent \textbf{Falcon-H1-0.5B-Base.} 
The results presented in Table~\ref{tab:05b-base} establish Falcon-H1-0.5B as a new benchmark for sub-1B parameter base models. Despite possessing the smallest footprint in the comparison, it leads on every Math, Science, and Code benchmark, often by substantial margins (e.g., GSM8k: 60.20 vs. 50.04; MATH-lvl5: 15.18 vs. 9.29). The model further confirms its strength in structured, knowledge-intensive reasoning by securing the highest scores on BBH and MMLU. Its performance on commonsense benchmarks like HellaSwag and Winogrande is competitive but surpassed by larger models in the 1B-1.6B class, suggesting a design that prioritizes deep reasoning capabilities over broad world knowledge. Ultimately, Falcon-H1-0.5B's results demonstrate that with targeted design choices, even a 0.5B model can achieve highly competitive, non-trivial performance on complex tasks.

\begin{table}[htbp]
\centering
\footnotesize
\begin{tabular*}{\textwidth}{l @{\extracolsep{\fill}} *{6}{S[table-format=2.2]}}
\toprule
\textbf{Tasks} & {\makecell{Falcon-H1- \\ 0.5B}} & {\makecell{Qwen3- \\ 0.6B}} & {\makecell{Qwen2.5- \\ 0.5B}} & {\makecell{Gemma3- \\ 1B}} & {\makecell{Llama3.2- \\ 1.2B}} & {\makecell{Falcon3- \\ 1.6B}} \\
\midrule
\multicolumn{7}{l}{\textbf{General}} \\
\quad BBH & \textbf{40.22} & \underline{36.07} & 32.62 & 30.26 & 30.72 & 35.24 \\
\quad MMLU & \textbf{55.04} & \underline{52.64} & 47.61 & 26.33 & 32.39 & 45.14 \\
\quad ARC-C & \underline{46.93} & 44.80 & 35.32 & 39.33 & 39.42 & \textbf{47.87} \\
\quad HellaSwag & 56.30 & 53.51 & 51.79 & \underline{62.94} & \textbf{65.73} & 62.30 \\
\quad Winogrande & 59.43 & 60.54 & 56.83 & \underline{62.59} & \textbf{62.75} & 61.17 \\
\midrule
\multicolumn{7}{l}{\textbf{Math}} \\
\quad GSM8k & \textbf{60.20} & \underline{50.04} & 34.80 & 2.20 & 7.05 & 34.95 \\
\quad MATH lvl5 & \textbf{15.18} & \underline{9.29} & 4.23 & 1.21 & 0.98 & 3.40 \\
\midrule
\multicolumn{7}{l}{\textbf{Science}} \\
\quad GPQA & \textbf{29.70} & \underline{29.11} & 27.94 & 24.66 & 23.57 & 27.85 \\
\quad MMLU-Pro & \textbf{30.04} & \underline{22.99} & 18.98 & 11.31 & 11.80 & 16.11 \\
\quad MMLU-stem & \textbf{57.12} & \underline{50.11} & 43.74 & 27.59 & 30.19 & 40.06 \\
\midrule
\multicolumn{7}{l}{\textbf{Code}} \\
\quad HumanEval & \textbf{35.98} & \underline{31.71} & 29.27 & 6.71 & 18.90 & 10.37 \\
\quad HumanEval+ & \textbf{31.10} & \underline{27.44} & 25.00 & 5.49 & 16.46 & 9.15 \\
\quad MBPP & \textbf{52.12} & \underline{51.06} & 40.74 & 12.70 & 35.98 & 12.43 \\
\quad MBPP+ & \textbf{43.39} & \underline{42.33} & 34.66 & 9.52 & 29.89 & 9.52 \\
\bottomrule
\end{tabular*}
\caption{Performance of the 0.5B+ \textbf{Base} models.}
\label{tab:05b-base}
\end{table}

\noindent \textbf{Falcon-H1-1.5B-Base and Falcon-H1-1.5B-Deep-Base.} 
The evaluation results for our 1.5B-scale models, presented in Table~\ref{tab:1b-base}, highlight the significant benefits of increased model depth. As discussed in Section~\ref{sec:width_depth_tradeoff}, the \texttt{Falcon-H1-1.5B-Deep} variant, which features more layers while maintaining a similar parameter count, establishes itself as the clear state-of-the-art model in its class. Its performance is highly competitive, often rivaling that of current leading 7B to 10B models such as Qwen2.5-7B~\citep{yang2024qwen2_5} and Falcon3-7B/10B~\citep{Falcon3}. This architectural choice proves particularly impactful for reasoning-intensive tasks; the deep model substantially outperforms its shallower counterpart and other peers on Science and Math benchmarks like MATH-lvl5 (+4.38 points) and MMLU-Pro (+5.54 points).

This strong performance continues into the Code and Multilingual domains, where Falcon-H1-1.5B-Deep consistently leads or is highly competitive. Although it is surpassed by Qwen3-1.7B~\citep{qwen3} on specific benchmarks like HumanEval and GSM8k, its dominant overall profile underscores the efficacy of our deep architecture. The result also suggests that while benchmark categories are distinct, the underlying skills required to solve them are often shared. A strong, generalizable reasoning capability, which our deeper model demonstrates, appears to be a critical polyvalent skill that confers benefits across multiple domains.

\begin{table}[htbp]
\centering
\footnotesize
\begin{tabular*}{\textwidth}{l @{\extracolsep{\fill}} *{7}{S[table-format=2.2]}}
\toprule
\textbf{Tasks} & {\makecell{Falcon-H1- \\ 1.5B-Deep}} & {\makecell{Falcon-H1- \\ 1.5B}} & {\makecell{Qwen3- \\ 1.7B}} & {\makecell{Qwen2.5- \\ 1.5B}} & {\makecell{Gemma3- \\ 1B}} & {\makecell{Llama3.2- \\ 1.2B}} & {\makecell{Falcon3- \\ 1.6B}} \\
\midrule
\multicolumn{8}{l}{\textbf{General}} \\
\quad BBH & \textbf{52.37} & \underline{46.57} & 43.05 & 40.55 & 30.26 & 30.72 & 35.24 \\
\quad MMLU & \textbf{66.29} & 61.81 & \underline{62.46} & 61.13 & 26.33 & 32.39 & 45.14 \\
\quad ARC-C & \textbf{55.89} & 53.24 & \underline{55.72} & 54.27 & 39.33 & 39.42 & 47.87 \\
\quad HellaSwag & \textbf{69.72} & 66.76 & 67.09 & \underline{67.86} & 62.94 & 65.73 & 62.30 \\
\quad Winogrande & \textbf{67.09} & 65.59 & \underline{66.30} & 64.56 & 62.59 & 62.75 & 61.17 \\
\midrule
\multicolumn{8}{l}{\textbf{Math}} \\
\quad GSM8k & \underline{68.69} & 52.01 & \textbf{70.74} & 63.00 & 2.20 & 7.05 & 34.95 \\
\quad MATH lvl5 & \textbf{24.77} & \underline{20.39} & 16.39 & 8.84 & 1.21 & 0.98 & 3.40 \\
\midrule
\multicolumn{8}{l}{\textbf{Science}} \\
\quad GPQA & \textbf{32.80} & 29.11 & \underline{29.45} & 28.36 & 24.66 & 23.57 & 27.85 \\
\quad MMLU-Pro & \textbf{41.07} & \underline{35.53} & 33.81 & 28.72 & 11.31 & 11.80 & 16.11 \\
\quad MMLU-stem & \textbf{67.43} & \underline{63.37} & 61.53 & 54.93 & 27.59 & 30.19 & 40.06 \\
\midrule
\multicolumn{8}{l}{\textbf{Code}} \\
\quad HumanEval & \underline{52.44} & 50.00 & \textbf{67.68} & 35.37 & 6.71 & 18.90 & 10.37 \\
\quad HumanEval+ & \underline{46.34} & 42.68 & \textbf{60.98} & 29.27 & 5.49 & 16.46 & 9.15 \\
\quad MBPP & \textbf{70.90} & 65.08 & \underline{67.72} & 60.05 & 12.70 & 35.98 & 12.43 \\
\quad MBPP+ & \textbf{60.32} & 55.03 & \underline{58.99} & 49.47 & 9.52 & 29.89 & 9.52 \\
\midrule
\multicolumn{8}{l}{\textbf{Multilingual}} \\
\quad Multi-Hellaswag & \textbf{50.36} & \underline{46.62} & 46.47 & 42.89 & 46.14 & 41.61 & 31.42 \\
\quad Multi-MMLU & \textbf{52.00} & 46.51 & {-} & \underline{48.09} & 26.50 & 28.22 & 31.56 \\
\quad MGSM & \textbf{60.33} & \underline{50.80} & {-} & 45.13 & {-} & 4.73 & 9.40 \\
\bottomrule
\end{tabular*}
\caption{Performance of the 1B+ \textbf{Base} models.}
\label{tab:1b-base}
\end{table}

\noindent \textbf{Falcon-H1-3B-Base.}
At the 3B-4B parameter scale, Falcon-H1-3B showcases exceptional training efficiency. Despite being trained on only 2.5T tokens—an order of magnitude less data than the 36T tokens reportedly used for Qwen3~\citep{qwen3}—our model delivers a highly competitive performance profile, as shown in Table~\ref{tab:3b-base}. While Qwen3-4B's extensive training confers an advantage on many general, science, and coding benchmarks, Falcon-H1-3B's resource-efficient approach enables it to achieve state-of-the-art capabilities in tasks like advanced mathematical reasoning. It secures leading scores on the challenging MATH-lvl5 benchmark (25.83) and the multilingual MGSM (64.00). Crucially, its performance had not yet plateaued at the conclusion of training, suggesting that its already strong results represent a conservative estimate of its full potential and that our targeted data strategy can achieve specialized excellence with significantly less computational cost.
\begin{table}[htbp]
\centering
\footnotesize
\begin{tabular*}{\textwidth}{l @{\extracolsep{\fill}} *{6}{S[table-format=2.2]}}
\toprule
\textbf{Tasks} & {\makecell{Falcon-H1- \\ 3B}} & {\makecell{Qwen3- \\ 4B}} & {\makecell{Qwen2.5- \\ 3B}} & {\makecell{Gemma3- \\ 4B}} & {\makecell{Llama3.2- \\ 3B}} & {\makecell{Falcon3- \\ 3B}} \\
\midrule
\multicolumn{7}{l}{\textbf{General}} \\
\quad BBH & \underline{53.17} & \textbf{56.88} & 46.40 & 40.41 & 39.45 & 44.02 \\
\quad MMLU & \underline{68.39} & \textbf{72.92} & 65.56 & 59.41 & 55.94 & 56.77 \\
\quad ARC-C & \underline{61.35} & \textbf{64.33} & 56.57 & 58.36 & 51.02 & 55.12 \\
\quad HellaSwag & 73.85 & 75.74 & 74.60 & \textbf{77.62} & \underline{76.39} & 67.13 \\
\quad Winogrande & 68.11 & \underline{72.30} & 71.03 & \textbf{72.77} & 72.22 & 65.11 \\
\midrule
\multicolumn{7}{l}{\textbf{Math}} \\
\quad GSM8k & 68.31 & \textbf{81.65} & \underline{74.60} & 37.60 & 27.82 & 64.67 \\
\quad MATH lvl5 & \textbf{25.83} & \underline{24.47} & 16.09 & 6.95 & 1.74 & 11.56 \\
\midrule
\multicolumn{7}{l}{\textbf{Science}} \\
\quad GPQA & \underline{32.63} & \textbf{34.90} & 28.44 & 29.78 & 28.78 & 29.78 \\
\quad MMLU-Pro & \underline{40.58} & \textbf{46.18} & 32.12 & 28.34 & 25.08 & 29.03 \\
\quad MMLU-stem & \underline{69.55} & \textbf{75.58} & 62.23 & 51.70 & 47.67 & 55.34 \\
\midrule
\multicolumn{7}{l}{\textbf{Code}} \\
\quad HumanEval & \underline{59.15} & \textbf{74.39} & 42.68 & 33.54 & 29.27 & 36.59 \\
\quad HumanEval+ & \underline{53.66} & \textbf{68.90} & 35.37 & 28.05 & 26.22 & 31.71 \\
\quad MBPP & \underline{71.43} & \textbf{74.60} & 59.52 & 60.05 & 48.94 & 51.85 \\
\quad MBPP+ & \underline{57.94} & \textbf{63.76} & 50.53 & 51.32 & 39.42 & 42.06 \\
\midrule
\multicolumn{7}{l}{\textbf{Multilingual}} \\
\quad Multi-Hellaswag & \underline{55.15} & 56.69 & 54.71 & \textbf{61.03} & 53.89 & 36.30 \\
\quad Multi-MMLU & \underline{54.78} & \textbf{64.91} & 55.13 & 52.57 & 45.45 & 38.31 \\
\quad MGSM & \textbf{64.00} & {-} & \underline{59.93} & {-} & 20.33 & 31.80 \\
\bottomrule
\end{tabular*}
\caption{Performance of the 3B+ \textbf{Base} models.}
\label{tab:3b-base}
\end{table}

\noindent \textbf{Falcon-H1-7B-Base.} 
At the 7B+ parameter scale, as detailed in Table~\ref{tab:7b-base}, Falcon-H1-7B establishes a new state-of-the-art benchmark, particularly in complex, knowledge-intensive domains. It demonstrates clear leadership in advanced reasoning by securing top scores on MMLU (77.38), MATH-lvl5 (34.67), and the challenging GPQA science benchmark (36.58). Furthermore, it excels in code generation tasks, leading on MBPP and MBPP+, and also achieves the best performance in multilingual mathematics on MGSM. This consistent top-tier performance underscores its exceptional capabilities in structured reasoning and specialized knowledge application.
The competitive landscape at this scale also reveals distinct strengths among other models. The Qwen models, for instance, exhibit a strong aptitude for coding, with Qwen3-8B leading on HumanEval benchmarks. Meanwhile, Gemma3-12B, despite its larger size, primarily excels on commonsense reasoning tasks like HellaSwag and Winogrande. This distribution of results highlights a key finding: while larger models may show advantages in general commonsense tasks, the architectural and data choices of Falcon-H1-7B make it a superior model for high-value, reasoning-focused applications in science, math, and code, while also maintaining an advantage over its counterpart, Qwen3-8B, on general tasks requiring more world knowledge.

\begin{table}[htbp]
\centering
\footnotesize
\begin{tabular*}{\textwidth}{l @{\extracolsep{\fill}} *{7}{S[table-format=2.2]}}
\toprule
\textbf{Tasks} & {\makecell{Falcon-H1- \\ 7B}} & {\makecell{Qwen3- \\ 8B}} & {\makecell{Qwen2.5- \\ 7B}} & {\makecell{Gemma3- \\ 12B}} & {\makecell{Llama3.1- \\ 8B}} & {\makecell{Falcon3- \\ 7B}} & {\makecell{Falcon3- \\ 10B}} \\
\midrule
\multicolumn{8}{l}{\textbf{General}} \\
\quad BBH & \textbf{60.61} & 58.44 & 53.72 & 54.33 & 46.52 & 50.88 & \underline{59.30} \\
\quad MMLU & \textbf{77.38} & \underline{76.63} & 74.17 & 74.23 & 65.17 & 69.98 & 73.22 \\
\quad ARC-C & 65.19 & \textbf{67.75} & 63.91 & \underline{67.58} & 57.68 & 62.71 & 67.49 \\
\quad HellaSwag & 81.26 & 79.60 & 80.20 & \textbf{84.22} & \underline{81.97} & 76.69 & 79.64 \\
\quad Winogrande & \underline{79.01} & 76.80 & 76.01 & \textbf{79.79} & 77.11 & 73.64 & \underline{79.01} \\
\midrule
\multicolumn{8}{l}{\textbf{Math}} \\
\quad GSM8k & 73.46 & \underline{83.02} & \textbf{83.09} & 71.19 & 49.51 & 76.95 & 82.11 \\
\quad MATH lvl5 & \textbf{34.67} & \underline{28.85} & 22.58 & 17.22 & 6.57 & 20.09 & 25.38 \\
\midrule
\multicolumn{8}{l}{\textbf{Science}} \\
\quad GPQA & \textbf{36.58} & \underline{35.65} & 32.30 & 34.56 & 31.46 & 35.07 & 35.40 \\
\quad MMLU-Pro & \textbf{48.38} & \underline{48.25} & 43.55 & 42.72 & 32.71 & 39.23 & 42.45 \\
\quad MMLU-stem & \underline{77.20} & \textbf{78.53} & 71.04 & 68.51 & 55.72 & 67.71 & 70.85 \\
\midrule
\multicolumn{8}{l}{\textbf{Code}} \\
\quad HumanEval & \underline{67.68} & \textbf{87.80} & 57.32 & 45.12 & 39.02 & 50.00 & 51.83 \\
\quad HumanEval+ & \underline{63.41} & \textbf{82.32} & 48.78 & 36.59 & 31.71 & 43.29 & 44.51 \\
\quad MBPP & \textbf{78.57} & 75.13 & \underline{76.72} & 73.02 & 61.38 & 67.99 & 73.54 \\
\quad MBPP+ & \textbf{67.20} & \underline{64.02} & 63.49 & 59.79 & 51.32 & 57.14 & 61.38 \\
\midrule
\multicolumn{8}{l}{\textbf{Multilingual}} \\
\quad Multi-Hellaswag & \underline{65.16} & 62.13 & 58.74 & \textbf{70.62} & 61.72 & 46.58 & 50.91 \\
\quad Multi-MMLU & \underline{67.55} & \textbf{68.71} & 64.07 & 67.14 & 53.58 & {-} & 53.17 \\
\quad MGSM & \textbf{74.53} & 67.87 & \underline{71.07} & {-} & 41.53 & 52.20 & 59.00 \\
\bottomrule
\end{tabular*}
\caption{Performance of the 7B+ \textbf{Base} models.}
\label{tab:7b-base}
\end{table}

\noindent \textbf{Falcon-H1-34B-Base.} 
In the 34B parameter class, we evaluate Falcon-H1-34B against a highly competitive field that includes models up to the 70B scale, as well as the Llama4-scout-17B MoE model (109B). Qwen3-32B is not included since no base model checkpoint was released. As shown in Table~\ref{tab:34b-base}, despite this challenging comparison, Falcon-H1-34B demonstrates state-of-the-art, parameter-efficient performance, distinguishing itself in specialized, complex domains, even when compared against significantly larger models. 

It secures the top position on challenging reasoning benchmarks such as BBH (69.36) and MATH-lvl5 (40.71). Furthermore, its leadership on the GPQA benchmark (42.70), all code generation tasks (HumanEval, HumanEval+), and multilingual mathematics (MGSM) underscores its exceptional and well-rounded capabilities in both reasoning and knowledge-intensive applications. The performance of other models at this scale reveals interesting trade-offs. The larger Qwen2.5-72B and Llama3.1-70B models show an advantage on general knowledge and commonsense reasoning tasks like MMLU, ARC-C, and HellaSwag. However, Falcon-H1-34B remains highly competitive, often securing the second-best score. This pattern reinforces a key finding from our smaller models: while increased scale can confer advantages on broad-knowledge tasks, the specialized architecture and data strategy of Falcon-H1 enable it to deliver superior, parameter-efficient performance on complex reasoning and code generation tasks. This positions Falcon-H1-34B as a leading model in its class and establishes it as a far more cost-efficient alternative to 70B+ models for developers seeking a powerful base for fine-tuning or specialized reasoning applications.

\begin{table}[htbp]
\centering
\footnotesize
\begin{tabular*}{\textwidth}{l @{\extracolsep{\fill}} *{6}{S[table-format=2.2]}}
\toprule
\textbf{Tasks} & {\makecell{Falcon-H1- \\ 34B}} & {\makecell{Qwen2.5- \\ 72B}} & {\makecell{Qwen2.5- \\ 32B}} & {\makecell{Gemma3- \\ 27B}} & {\makecell{Llama3.1- \\ 70B}} & {\makecell{Llama4- \\ scout}} \\
\midrule
\multicolumn{7}{l}{\textbf{General}} \\
\quad BBH & \textbf{69.36} & \underline{67.77} & 67.45 & 61.60 & 62.78 & 61.71 \\
\quad MMLU & \underline{83.46} & \textbf{85.96} & 83.18 & 78.32 & 78.49 & 77.98 \\
\quad ARC-C & \underline{71.25} & \textbf{72.44} & 70.48 & 70.31 & 69.20 & 62.97 \\
\quad HellaSwag & 85.68 & \underline{87.57} & 85.13 & 86.19 & \textbf{87.78} & 84.01 \\
\quad Winogrande & 82.72 & \underline{83.74} & 82.32 & 82.40 & \textbf{85.32} & 78.93 \\
\midrule
\multicolumn{7}{l}{\textbf{Math}} \\
\quad GSM8k & 76.50 & \underline{89.76} & \textbf{90.14} & 81.35 & 80.52 & 83.24 \\
\quad MATH lvl5 & \textbf{40.71} & \underline{38.14} & 36.40 & 25.38 & 18.81 & 27.19 \\
\midrule
\multicolumn{7}{l}{\textbf{Science}} \\
\quad GPQA & \textbf{42.70} & \underline{42.28} & 39.68 & 35.82 & 36.49 & 35.99 \\
\quad MMLU-Pro & 57.18 & \textbf{60.22} & \underline{58.05} & 49.64 & 47.07 & 50.16 \\
\quad MMLU-stem & \underline{83.82} & \textbf{84.81} & 82.81 & 76.59 & 70.35 & 72.57 \\
\midrule
\multicolumn{7}{l}{\textbf{Code}} \\
\quad HumanEval & \textbf{70.12} & 59.15 & \underline{59.76} & 48.78 & 57.32 & 57.32 \\
\quad HumanEval+ & \textbf{64.63} & 51.22 & \underline{51.83} & 40.85 & 50.61 & 48.78 \\
\quad MBPP & \underline{83.33} & \textbf{87.04} & 83.07 & 76.19 & 78.84 & 77.78 \\
\quad MBPP+ & \underline{70.37} & \textbf{70.63} & 68.78 & 61.64 & 66.67 & 64.29 \\
\midrule
\multicolumn{7}{l}{\textbf{Multilingual}} \\
\quad Multi-Hellaswag & 72.62 & 71.20 & {-} & \underline{74.01} & \textbf{74.65} & 72.02 \\
\quad Multi-MMLU & \underline{76.76} & \textbf{78.54} & {-} & 72.41 & 71.10 & 72.38 \\
\quad MGSM & \textbf{82.40} & \underline{82.20} & {-} & {-} & 70.73 & 75.80 \\
\bottomrule
\end{tabular*}
\caption{Performance of the 34B+ \textbf{Base} models.}
\label{tab:34b-base}
\end{table}

\noindent \textbf{Overall Remarks of Falcon-H1 Base models.}
A key finding across our suite of base models is the achievement of state-of-the-art performance with remarkable training efficiency. Despite being trained on a modest 2.5T to 18T tokens, the Falcon-H1 series consistently challenges and often surpasses competitor models trained on substantially larger datasets. Our models establish clear leadership in complex, reasoning-intensive domains such as mathematics, science, and code generation across all evaluated scales. Notably, the performance of these models, particularly the smaller variants, had not yet plateaued at the conclusion of pretraining, indicating significant headroom for further optimizations with extended training. The resulting models exhibit a well-balanced performance profile, positioning them as powerful and efficient foundations for fine-tuning on specialized downstream applications.

For the full evaluation results on the model’s multilingual capabilities, please refer to the dedicated Appendix section \ref{eval:tables_ml_base}, where we report each model’s performance by language.

\subsection{Instruct Models}
For instruction-tuned models, we report in Table \ref{tab:eval_setting_ins} the benchmarks and settings used for the evaluations. For these models, we expanded the evaluation scope to cover a broader range of domains and tasks. While we provide a general categorization for clarity, it is important to note that some benchmarks are cross-domain and do not fall neatly into a single category. \texttt{Falcon-H1-0.5B-Instruct} model was evaluated exclusively on non-multilingual tasks, as it was trained only on English data. Long context benchmarks are only reported on 34B scale models for simplicity.
For both the multilingual and long-context evaluations, this section reports the average scores for each task category. A detailed, per-language and per-task breakdown of these results is available in Appendix~\ref{eval:tables_ml_ins} and Appendix~\ref{eval:tables_long_context}, respectively.




\begin{table}[htbp]
\centering
\footnotesize
\begin{tabularx}{\textwidth}{l >{\raggedright\arraybackslash}X l}
\toprule
\textbf{Benchmark} & \textbf{Settings} & \textbf{Framework}\\
\midrule
\multicolumn{3}{l}{\textbf{General}} \\
\quad BBH~\citep{suzgun2022challenging} & logprobs, 3-shot & \texttt{lm-eval-harness}\\
\quad ARC-C~\citep{clark2018think} & logprobs, 0-shot & \texttt{lm-eval-harness} \\
\quad TruthfulQA~\citep{lin2021truthfulqa} & logprobs, 0-shot & \texttt{lm-eval-harness} \\
\quad HellaSwag~\citep{zellers2019hellaswag} & logprobs, 0-shot & \texttt{lm-eval-harness}\\
\quad MMLU~\citep{hendrycks2020measuring} & logprobs, 5-shot & \texttt{lm-eval-harness}\\
\midrule
\multicolumn{3}{l}{\textbf{Math}} \\
\quad GSM8k~\citep{cobbe2021training} & strict match, 5-shot & \texttt{lm-eval-harness} \\
\quad MATH-500~\citep{lightman2023lets} & accuracy & \texttt{evalchemy} \\
\quad AMC-23 & average accuracy, 16 repetitions & \texttt{evalchemy} \\
\quad AIME-24~\citep{AIME25} & average accuracy, 16 repetitions & \texttt{evalchemy}\\
\quad AIME-25~\citep{AIME25} & average accuracy, 16 repetitions & \texttt{evalchemy} \\
\midrule
\multicolumn{3}{l}{\textbf{Science}} \\
\quad GPQA~\citep{rein2023gpqa} & logprobs, 5-shot & \texttt{lm-eval-harness}  \\
\quad GPQA\_Diamond~\citep{rein2023gpqa} & average accuracy, 3 repetitions & \texttt{evalchemy}\\
\quad MMLU-Pro~\citep{wang2024mmlu} & logprobs, 5-shot & \texttt{lm-eval-harness}  \\
\quad MMLU-stem~\citep{hendrycks2020measuring} & logprobs, 5-shot & \texttt{lm-eval-harness} \\
\midrule
\multicolumn{3}{l}{\textbf{Code}} \\
\quad HumanEval~\citep{chen2021codex}  & pass@1 & \texttt{evalplus} \\
\quad HumanEval+~\citep{evalplus} & pass@1 & \texttt{evalplus}\\
\quad MBPP~\citep{austin2021program} & pass@1 & \texttt{evalplus} \\
\quad MBPP+~\citep{evalplus} & pass@1 & \texttt{evalplus}\\
\quad LiveCodeBench~\citep{jain2024livecodebench} & accuracy & \texttt{evalchemy} \\
\quad CRUXEval~\citep{gu2024cruxeval} & pass@1, input \& output average & \texttt{evalchemy} \\
\midrule
\multicolumn{3}{l}{\textbf{Instruction Following \& Others}} \\
\quad IFEval~\citep{zhou2023instruction} & inst \& prompt avg accuracy & \texttt{lm-eval-harness} \\
\quad Alpaca-Eval~\citep{alpaca_eval} & LC winrate & \texttt{evalchemy}\\
\quad MTBench~\citep{bai2024mt} & turn 1 \& 2 average & \texttt{evalchemy} \\
\quad LiveBench~\citep{white2024livebench} & global\_average & \texttt{evalchemy}\\
\midrule
\multicolumn{3}{l}{\textbf{Multilingual}} \\
\quad Multi-Hellaswag~\citep{dac2023okapi} & logprobs, 0-shot & \texttt{lm-eval-harness} \\
\quad MGSM~\citep{shi2022language} & flexible extract, 8-shot native CoT & \texttt{lm-eval-harness} \\
\quad Multi-MMLU~\citep{dac2023okapi} & logprobs, 5-shot & \texttt{lm-eval-harness} \\
\midrule
\multicolumn{3}{l}{\textbf{Long Context}} \\
\quad HELMET-LongQA~\citep{yen2025helmet} & default & \texttt{helmet} \\
\quad HELMET-RAG~\citep{yen2025helmet} & default & \texttt{helmet} \\
\quad HELMET-Recall~\citep{yen2025helmet} & default & \texttt{helmet} \\
\bottomrule
\end{tabularx}
\caption{Evaluation settings and benchmark sources for instruction-tuned models.}
\label{tab:eval_setting_ins}
\end{table}

\noindent \textbf{Falcon-H1-0.5B-Instruct.} 
At the sub-1B parameter scale, the Falcon-H1-0.5B-Instruct model sets a new state-of-the-art benchmark, demonstrating a clear and consistent advantage in complex, reasoning-intensive domains as shown in Table~\ref{tab:05b-ins}. The model's superiority is most pronounced in Math, where it achieves a sweeping dominance across all five benchmarks. Its performance on GSM8k (68.39) and MATH-500 (58.40) is particularly notable, substantially outperforming all competitors. This strength in structured reasoning extends to the Science and Code categories, where it secures top scores on the majority of tasks, including MMLU-Pro, HumanEval, and CRUXEval. Furthermore, its leading performance on IFEval (72.07) confirms its exceptional ability to adhere to complex instructions.
However, the competitive landscape is not uniform. Competitors like Gemma3-1B and Qwen3-0.6B show an edge on certain coding (MBPP, LiveCodeBench) and preference-based instruction following (Alpaca-Eval, LiveBench) benchmarks. Similarly, on general commonsense tasks like HellaSwag and ARC-C, larger models in the comparison group hold an advantage. This specialized profile suggests that our fine-tuning strategy for Falcon-H1-0.5B-Instruct prioritizes deep, multi-step reasoning and precise instruction execution over performance on conversational or broad-knowledge benchmarks. This trade-off firmly establishes the model as the leading choice for applications requiring robust, complex problem-solving at an efficient scale.

\begin{table}[htbp]
\centering
\footnotesize
\begin{tabular*}{\textwidth}{l @{\extracolsep{\fill}} *{6}{S[table-format=2.2]}}
\toprule
\textbf{Tasks} & {Falcon-H1-0.5B} & {Qwen3-0.6B} & {Qwen2.5-0.5B} & {Gemma3-1B} & {Llama3.2-1.2B} & {Falcon3-1.6B} \\
\midrule
\multicolumn{7}{l}{\textbf{General}} \\
\quad BBH & \textbf{42.91} & 32.95 & 33.26 & \underline{35.86} & 33.21 & 34.47 \\
\quad ARC-C & \underline{37.80} & 31.06 & 33.28 & 34.13 & 34.64 & \textbf{43.09} \\
\quad TruthfulQA & 44.12 & \textbf{51.65} & \underline{46.19} & 42.17 & 42.08 & 42.31 \\
\quad HellaSwag & 51.93 & 42.17 & 52.38 & 42.24 & \underline{55.30} & \textbf{58.53} \\
\quad MMLU & \textbf{53.40} & 42.98 & \underline{46.07} & 40.87 & 45.93 & 46.10 \\
\midrule
\multicolumn{7}{l}{\textbf{Math}} \\
\quad GSM8k & \textbf{68.39} & 42.61 & 38.51 & 42.38 & \underline{44.28} & 44.05 \\
\quad MATH-500 & \textbf{58.40} & \underline{46.00} & 27.80 & 45.40 & 13.20 & 19.80 \\
\quad AMC-23 & \textbf{33.13} & \underline{27.97} & 12.50 & 19.22 & 7.19 & 6.87 \\
\quad AIME-24 & \textbf{3.75} & \underline{2.71} & 0.62 & 0.42 & 1.46 & 0.41 \\
\quad AIME-25 & \textbf{4.38} & \underline{1.67} & 0.21 & 1.25 & 0.00 & 0.21 \\
\midrule
\multicolumn{7}{l}{\textbf{Science}} \\
\quad GPQA & \textbf{29.95} & 26.09 & 26.85 & \underline{28.19} & 26.59 & 26.76 \\
\quad GPQA\_Diamond & 27.95 & 25.08 & 24.24 & 21.55 & \underline{25.08} & \textbf{31.31} \\
\quad MMLU-Pro & \textbf{31.03} & 16.95 & \underline{18.73} & 14.46 & 16.20 & 18.49 \\
\quad MMLU-stem & \textbf{54.55} & 39.30 & \underline{39.83} & 35.39 & 39.16 & 39.64 \\
\midrule
\multicolumn{7}{l}{\textbf{Code}} \\
\quad HumanEval & \textbf{51.83} & \underline{41.46} & 36.59 & 40.85 & 34.15 & 22.56 \\
\quad HumanEval+ & \textbf{45.12} & \underline{37.19} & 32.32 & 37.20 & 29.88 & 20.73 \\
\quad MBPP & 42.59 & \underline{56.08} & 46.83 & \textbf{57.67} & 33.60 & 20.63 \\
\quad MBPP+ & 33.07 & \underline{47.08} & 39.68 & \textbf{50.00} & 29.37 & 17.20 \\
\quad LiveCodeBench & \underline{7.05} & \textbf{9.78} & 2.94 & 5.09 & 2.35 & 0.78 \\
\quad CRUXEval & \textbf{25.75} & \underline{23.63} & 14.88 & 12.70 & 0.06 & 15.58 \\
\midrule
\multicolumn{7}{l}{\textbf{Instruction Following}} \\
\quad IFEval & \textbf{72.07} & \underline{62.16} & 32.11 & 61.48 & 55.34 & 54.26 \\
\quad Alpaca-Eval & 10.79 & 9.59 & 3.26 & \textbf{17.87} & \underline{9.38} & 6.98 \\
\quad MTBench & \textbf{7.06} & 5.75 & 4.71 & \underline{7.03} & 6.37 & 6.03 \\
\quad LiveBench & 20.80 & \textbf{27.78} & 14.27 & \underline{18.79} & 14.97 & 14.10 \\
\bottomrule
\end{tabular*}
\caption{Performance of the 0.5B+ \textbf{Instruct} models.}
\label{tab:05b-ins}
\end{table}

\noindent \textbf{Falcon-H1-1.5B-Instruct and Falcon-H1-1.5B-Deep-Instruct.}
At the 1.5B parameter scale, the Falcon-H1 instruct models establish clear dominance, with both the deep and shallow architecture variants setting new state-of-the-art benchmarks. As shown in Table~\ref{tab:1b-ins}, Falcon-H1-1.5B-Deep-Instruct achieves comprehensive leadership across nearly all evaluated domains, securing the top position on the vast majority of General, Math, Science, Code, and Multilingual tasks, reaching a level of performance competitive with current state-of-the-art 7B models, like Qwen3-8B~\citep{qwen3}, Qwen2.5-7B~\citep{yang2024qwen2} (check Table \ref{tab:7b-ins}). This sweeping success highlights the profound impact of our deep architectural design combined with instruction tuning. The performance gap is particularly striking in complex reasoning, where the model substantially outperforms all peers on benchmarks like GSM8k (82.34) and MATH-500 (77.80).
The shallower Falcon-H1-1.5B-Instruct model also delivers an exceptionally strong performance, consistently securing the second-best score across most benchmarks and often outperforming larger models like Qwen3-1.7B. Its leadership on Alpaca-Eval further underscores its well-rounded instruction-following capabilities. The overwhelming evidence from these evaluations confirms that the Falcon-H1-1.5B-Instruct models, particularly the deep variant, are leading choices for sophisticated, reasoning-driven tasks, delivering performance that often transcends their parameter scale.
\begin{table}[htbp]
\centering
\footnotesize
\begin{tabular*}{\textwidth}{l @{\extracolsep{\fill}} *{7}{S[table-format=2.2]}}
\toprule
\textbf{Tasks} & {\makecell{Falcon-H1- \\ 1.5B-Deep}} & {\makecell{Falcon-H1- \\ 1.5B}} & {\makecell{Qwen3- \\ 1.7B}} & {\makecell{Qwen2.5- \\ 1.5B}} & {\makecell{Gemma3- \\ 1B}} & {\makecell{Llama3.2- \\ 1B}} & {\makecell{Falcon3- \\ 1.6B}} \\
\midrule
\multicolumn{8}{l}{\textbf{General}} \\
\quad BBH & \textbf{54.43} & \underline{46.47} & 35.18 & 42.41 & 35.86 & 33.21 & 34.47 \\
\quad ARC-C & \textbf{43.86} & 42.06 & 34.81 & 40.53 & 34.13 & 34.64 & \underline{43.09} \\
\quad TruthfulQA & \textbf{50.48} & \underline{49.39} & 45.98 & 47.05 & 42.17 & 42.08 & 42.31 \\
\quad HellaSwag & \textbf{65.54} & \underline{63.33} & 49.27 & 62.23 & 42.24 & 55.30 & 58.53 \\
\quad MMLU & \textbf{66.11} & \underline{62.03} & 57.04 & 59.76 & 40.87 & 45.93 & 46.10 \\
\midrule
\multicolumn{8}{l}{\textbf{Math}} \\
\quad GSM8k & \textbf{82.34} & \underline{74.98} & 69.83 & 57.47 & 42.38 & 44.28 & 44.05 \\
\quad MATH-500 & \textbf{77.80} & \underline{74.00} & 73.00 & 48.40 & 45.40 & 13.20 & 19.80 \\
\quad AMC-23 & \textbf{56.56} & \underline{46.09} & 43.59 & 24.06 & 19.22 & 7.19 & 6.87 \\
\quad AIME-24 & \textbf{14.37} & \underline{12.50} & 11.25 & 2.29 & 0.42 & 1.46 & 0.41 \\
\quad AIME-25 & \textbf{11.04} & \underline{9.58} & 8.12 & 1.25 & 1.25 & 0.00 & 0.21 \\
\midrule
\multicolumn{8}{l}{\textbf{Science}} \\
\quad GPQA & \textbf{33.22} & 26.34 & 27.68 & 26.26 & \underline{28.19} & 26.59 & 26.76 \\
\quad GPQA\_Diamond & \textbf{40.57} & \underline{35.19} & 33.33 & 25.59 & 21.55 & 25.08 & 31.31 \\
\quad MMLU-Pro & \textbf{41.89} & \underline{37.80} & 23.54 & 28.35 & 14.46 & 16.20 & 18.49 \\
\quad MMLU-stem & \textbf{67.30} & \underline{64.13} & 54.30 & 54.04 & 35.39 & 39.16 & 39.64 \\
\midrule
\multicolumn{8}{l}{\textbf{Code}} \\
\quad HumanEval & \textbf{73.78} & \underline{68.29} & 67.68 & 56.10 & 40.85 & 34.15 & 22.56 \\
\quad HumanEval+ & \textbf{68.90} & \underline{61.59} & 60.96 & 50.61 & 37.20 & 29.88 & 20.73 \\
\quad MBPP & \textbf{68.25} & \underline{64.81} & 58.73 & \underline{64.81} & 57.67 & 33.60 & 20.63 \\
\quad MBPP+ & \textbf{56.61} & \underline{56.35} & 49.74 & 56.08 & 50.00 & 29.37 & 17.20 \\
\quad LiveCodeBench & \textbf{23.87} & \underline{17.61} & 14.87 & 12.52 & 5.09 & 2.35 & 0.78 \\
\quad CRUXEval & \textbf{52.32} & \underline{39.57} & 18.88 & 34.76 & 12.70 & 0.06 & 15.58 \\
\midrule
\multicolumn{8}{l}{\textbf{Instruction Following}} \\
\quad IFEval & \textbf{83.50} & \underline{80.66} & 70.77 & 45.33 & 61.48 & 55.34 & 54.26 \\
\quad Alpaca-Eval & \underline{27.12} & \textbf{28.18} & 21.89 & 9.54 & 17.87 & 9.38 & 6.98 \\
\quad MTBench & \textbf{8.53} & \underline{8.46} & 7.61 & 7.10 & 7.03 & 6.37 & 6.03 \\
\quad LiveBench & \underline{36.83} & 34.13 & \textbf{40.73} & 21.65 & 18.79 & 14.97 & 14.10 \\
\midrule
\multicolumn{8}{l}{\textbf{Multilingual}} \\
\quad Multi-Hellaswag & \textbf{53.14} & \underline{49.38} & 37.89 & 42.93 & 41.77 & 39.78 & 32.04 \\
\quad Multi-MMLU & \textbf{53.00} & \underline{48.06} & 39.60 & 45.90 & 34.91 & 35.24 & 32.25 \\
\quad MGSM & \textbf{60.00} & \underline{58.00} & 52.40 & 45.20 & {-} & 29.73 & 15.33 \\
\bottomrule
\end{tabular*}
\caption{Performance of the 1B+ \textbf{Instruct} models.}
\label{tab:1b-ins}
\end{table}

\noindent \textbf{Falcon-H1-3B-Instruct.}
At the 3B-4B scale, Falcon-H1-3B-Instruct emerges as a top-performing and highly versatile model, demonstrating clear strengths in reasoning, science, and instruction following (Table~\ref{tab:3b-ins}). It leads on the majority of General knowledge benchmarks like MMLU and BBH, and dominates the Science category. This shows that after instruction tuning, the model excels at applying its knowledge. Furthermore, its state-of-the-art scores on IFEval and MTBench highlight its superior ability to understand and follow complex instructions.
While other models show specialized strengths, such as Qwen3-4B in mathematics, Falcon-H1-3B remains highly competitive across all areas. It also shows a distinct advantage in Code generation on MBPP and excels in Multilingual tasks. This balanced and powerful performance across many different domains establishes Falcon-H1-3B-Instruct as a premier, all-around model in its class.
\begin{table}[htbp]
\centering
\footnotesize
\begin{tabular*}{\textwidth}{l @{\extracolsep{\fill}} *{6}{S[table-format=2.2]}}
\toprule
\textbf{Tasks} & {\makecell{Falcon-H1- \\ 3B}} & {\makecell{Qwen3- \\ 4B}} & {\makecell{Qwen2.5- \\ 3B}} & {\makecell{Gemma3- \\ 4B}} & {\makecell{Llama3.2- \\ 3B}} & {\makecell{Falcon3- \\ 3B}} \\
\midrule
\multicolumn{7}{l}{\textbf{General}} \\
\quad BBH & \textbf{53.69} & \underline{51.07} & 46.55 & 50.01 & 41.47 & 45.02 \\
\quad ARC-C & \textbf{49.57} & 37.71 & 43.77 & 44.88 & 44.88 & \underline{48.21} \\
\quad TruthfulQA & \underline{53.19} & 51.75 & \textbf{58.11} & 51.68 & 50.27 & 50.06 \\
\quad HellaSwag & \textbf{69.85} & 55.31 & \underline{64.21} & 47.68 & 63.74 & 64.24 \\
\quad MMLU & \textbf{68.30} & \underline{67.01} & 65.09 & 59.53 & 61.74 & 56.76 \\
\midrule
\multicolumn{7}{l}{\textbf{Math}} \\
\quad GSM8k & \textbf{84.76} & \underline{80.44} & 57.54 & 77.41 & 77.26 & 74.68 \\
\quad MATH-500 & 74.20 & \textbf{85.00} & 64.20 & \underline{76.40} & 41.20 & 54.20 \\
\quad AMC-23 & \underline{55.63} & \textbf{66.88} & 39.84 & 48.12 & 22.66 & 29.69 \\
\quad AIME-24 & 11.88 & \textbf{22.29} & 6.25 & 6.67 & \underline{11.67} & 3.96 \\
\quad AIME-25 & \underline{13.33} & \textbf{18.96} & 3.96 & \underline{13.33} & 0.21 & 2.29 \\
\midrule
\multicolumn{7}{l}{\textbf{Science}} \\
\quad GPQA & \textbf{33.89} & 28.02 & 28.69 & \underline{29.19} & 28.94 & 28.69 \\
\quad GPQA\_Diamond & \underline{38.72} & \textbf{40.74} & 35.69 & 28.62 & 29.97 & 29.29 \\
\quad MMLU-Pro & \textbf{43.69} & 29.75 & \underline{32.76} & 29.71 & 27.44 & 29.71 \\
\quad MMLU-stem & \textbf{69.93} & \underline{67.46} & 59.78 & 52.17 & 51.92 & 56.11 \\
\midrule
\multicolumn{7}{l}{\textbf{Code}} \\
\quad HumanEval & \underline{76.83} & \textbf{84.15} & 73.78 & 67.07 & 54.27 & 52.44 \\
\quad HumanEval+ & \underline{70.73} & \textbf{76.83} & 68.29 & 61.59 & 50.00 & 45.73 \\
\quad MBPP & \textbf{79.63} & 68.78 & 72.75 & \underline{77.78} & 62.17 & 61.90 \\
\quad MBPP+ & \textbf{67.46} & 59.79 & 60.85 & \underline{66.93} & 50.53 & 55.29 \\
\quad LiveCodeBench & 26.81 & \textbf{39.92} & 11.74 & \underline{21.14} & 2.74 & 3.13 \\
\quad CRUXEval & \underline{56.25} & \textbf{69.63} & 43.26 & 52.13 & 17.75 & 44.38 \\
\midrule
\multicolumn{7}{l}{\textbf{Instruction Following}} \\
\quad IFEval & \textbf{85.05} & \underline{84.01} & 64.26 & 77.01 & 74.00 & 69.10 \\
\quad Alpaca-Eval & 31.09 & \underline{36.51} & 17.37 & \textbf{39.64} & 19.69 & 14.82 \\
\quad MTBench & \textbf{8.72} & \underline{8.45} & 7.79 & 8.24 & 7.96 & 7.79 \\
\quad LiveBench & 36.86 & \textbf{51.34} & 27.32 & \underline{36.70} & 26.37 & 26.01 \\
\midrule
\multicolumn{7}{l}{\textbf{Multilingual}} \\
\quad Multi-Hellaswag & \textbf{58.34} & 43.12 & 50.81 & \underline{54.48} & 50.93 & 37.51 \\
\quad Multi-MMLU & \textbf{54.90} & 50.70 & \underline{52.90} & 51.10 & 48.40 & 38.90 \\
\quad MGSM & \underline{63.90} & \textbf{68.90} & 57.30 & {-} & 62.20 & 42.10 \\
\bottomrule
\end{tabular*}
\caption{Performance of the 3B+ \textbf{Instruct} models.}
\label{tab:3b-ins}
\end{table}

\noindent \textbf{Falcon-H1-7B-Instruct.}
At the 7B-12B parameter scale, Falcon-H1-7B-Instruct demonstrates a highly competitive and well-rounded performance profile, outperforming the larger Gemma3-12B model~\citep{team2025gemma3} on a majority of benchmarks (Table~\ref{tab:7b-ins}). When compared to its direct counterparts like Qwen3-8B~\citep{qwen3}, its strengths in knowledge-intensive domains become particularly evident. The model leads on all four Science benchmarks and on key General reasoning tasks such as MMLU and ARC-C. It also exhibits strong practical skills, achieving top scores on HumanEval and HumanEval+ for code generation and leading across all three Multilingual benchmarks.
While competitors like Qwen3-8B and Gemma3-12B show advantages in specific areas, particularly on several Math and preference-based benchmarks (e.g., Alpaca-Eval, LiveBench), the collective results highlight a key distinction. The broad and deep capabilities of Falcon-H1-7B-Instruct across science, reasoning, code, and multilingualism make it an exceptionally effective and versatile model for a wide range of sophisticated, instruction-driven applications.

\begin{table}[htbp]
\centering
\footnotesize
\begin{tabular*}{\textwidth}{l @{\extracolsep{\fill}} *{7}{S[table-format=2.2]}}
\toprule
\textbf{Tasks} & {\makecell{Falcon-H1- \\ 7B}} & {\makecell{Qwen3- \\ 8B}} & {\makecell{Qwen2.5- \\ 7B}} & {\makecell{Gemma3- \\ 12B}} & {\makecell{Llama3.1- \\ 8B}} & {\makecell{Falcon3- \\ 7B}} & {\makecell{Falcon3- \\ 10B}} \\
\midrule
\multicolumn{8}{l}{\textbf{General}} \\
\quad BBH & \underline{62.28} & 47.47 & 53.76 & \textbf{63.36} & 48.58 & 52.12 & 58.09 \\
\quad ARC-C & \textbf{59.98} & 42.06 & 41.38 & 51.96 & 52.39 & \underline{54.35} & 54.44 \\
\quad TruthfulQA & 59.91 & 53.19 & \textbf{62.41} & \underline{61.02} & 52.99 & 55.58 & 55.05 \\
\quad HellaSwag & \textbf{75.92} & 60.56 & 63.40 & 55.63 & 71.28 & 71.81 & \underline{75.57} \\
\quad MMLU & \textbf{76.83} & 71.56 & 73.64 & \underline{72.50} & 68.67 & 70.81 & 74.01 \\
\midrule
\multicolumn{8}{l}{\textbf{Math}} \\
\quad GSM8k & 81.65 & 78.92 & 71.95 & \textbf{87.49} & 82.49 & 81.05 & \underline{85.06} \\
\quad MATH-500 & 73.40 & \underline{83.80} & 75.80 & \textbf{86.20} & 45.80 & 69.00 & 68.60 \\
\quad AMC-23 & \underline{56.72} & \textbf{70.78} & 53.91 & 66.88 & 22.81 & 40.00 & 45.78 \\
\quad AIME-24 & 16.04 & \textbf{28.33} & 12.29 & \underline{22.50} & 5.42 & 8.75 & 9.79 \\
\quad AIME-25 & 13.96 & \textbf{19.17} & 9.58 & \underline{18.75} & 0.42 & 6.25 & 5.42 \\
\midrule
\multicolumn{8}{l}{\textbf{Science}} \\
\quad GPQA & \textbf{36.33} & 25.84 & 31.79 & \underline{33.98} & 32.72 & 31.21 & 33.39 \\
\quad GPQA\_Diamond & \textbf{56.90} & \underline{43.10} & 33.00 & 37.71 & 31.31 & 37.21 & 34.68 \\
\quad MMLU-Pro & \textbf{51.75} & 34.64 & 43.23 & 39.88 & 36.42 & 40.73 & \underline{44.05} \\
\quad MMLU-stem & \textbf{77.61} & 66.89 & \underline{69.36} & 66.54 & 59.31 & 67.43 & 70.57 \\
\midrule
\multicolumn{8}{l}{\textbf{Code}} \\
\quad HumanEval & \textbf{86.59} & \underline{84.75} & 82.32 & 84.76 & 68.29 & 71.95 & 82.32 \\
\quad HumanEval+ & \textbf{81.10} & \underline{79.27} & 73.78 & 75.61 & 61.59 & 65.85 & 75.00 \\
\quad MBPP & 80.69 & 71.96 & 79.63 & \textbf{85.71} & 68.25 & \underline{77.25} & 73.28 \\
\quad MBPP+ & 68.78 & 62.70 & 68.25 & \textbf{72.22} & 55.03 & \underline{65.87} & 64.02 \\
\quad LiveCodeBench & \underline{35.03} & \textbf{45.60} & 32.68 & 30.92 & 15.85 & 12.72 & 19.77 \\
\quad CRUXEval & \underline{66.51} & \textbf{72.70} & 56.90 & 67.67 & 21.57 & 55.00 & 59.57 \\
\midrule
\multicolumn{8}{l}{\textbf{Instruction Following}} \\
\quad IFEval & \textbf{85.35} & \underline{83.43} & 75.25 & 81.51 & 77.04 & 76.59 & 78.84 \\
\quad Alpaca-Eval & 40.23 & \textbf{46.13} & 29.48 & \underline{43.55} & 25.48 & 27.56 & 24.31 \\
\quad MTBench & \textbf{8.85} & \underline{8.74} & 8.45 & 8.69 & 8.29 & 8.73 & 8.46 \\
\quad LiveBench & \underline{45.74} & \textbf{56.19} & 37.13 & 49.23 & 31.73 & 32.35 & 34.30 \\
\midrule
\multicolumn{8}{l}{\textbf{Multilingual}} \\
\quad Multi-Hellaswag & \textbf{67.75} & 47.30 & 58.74 & \underline{66.53} & 60.74 & 47.81 & 52.77 \\
\quad Multi-MMLU & \textbf{67.83} & 50.44 & 61.20 & \underline{65.22} & 55.53 & 50.62 & 53.67 \\
\quad MGSM & \textbf{73.50} & 65.20 & 66.10 & {-} & \underline{70.70} & 56.30 & 64.80 \\
\bottomrule
\end{tabular*}
\caption{Performance of the 7B+ \textbf{Instruct} models.}
\label{tab:7b-ins}
\end{table}

\noindent \textbf{Falcon-H1-34B-Instruct.}
At the 34B scale, Falcon-H1-34B-Instruct demonstrates that exceptional performance does not require massive parameter counts. As shown in Table~\ref{tab:34b-ins}, our 34B model consistently competes with and often outperforms models twice its size. Its primary strength lies in its deep knowledge and reasoning, leading across the entire Science category and on general reasoning benchmarks like HellaSwag. Furthermore, its top score on MTBench confirms its high-quality conversational abilities. While larger models leverage their scale to gain an edge in mathematics and some coding benchmarks, this profile highlights Falcon-H1-34B’s remarkable parameter efficiency. It delivers state-of-the-art results in science and general reasoning while being significantly smaller, making it a powerful and cost-effective choice for knowledge-intensive applications.
\begin{table}[htbp]
\centering
\footnotesize
\begin{tabular*}{\textwidth}{l @{\extracolsep{\fill}} *{7}{S[table-format=2.2]}}
\toprule
\textbf{Tasks} & {\makecell{Falcon-H1- \\ 34B}} & {\makecell{Qwen3- \\ 32B}} & {\makecell{Qwen2.5- \\ 72B}} & {\makecell{Qwen2.5- \\ 32B}} & {\makecell{Gemma3- \\ 27B}} & {\makecell{Llama3.3- \\ 70B}} & {\makecell{Llama4- \\ scout}} \\
\midrule
\multicolumn{8}{l}{\textbf{General}} \\
\quad BBH & \underline{70.68} & 62.47 & \textbf{72.52} & 68.72 & 67.28 & 69.15 & 64.90 \\
\quad ARC-C & 61.01 & 48.98 & 46.59 & 44.54 & 54.52 & \textbf{63.65} & \underline{56.14} \\
\quad TruthfulQA & 65.27 & 58.58 & \underline{69.80} & \textbf{70.28} & 64.26 & 66.15 & 62.74 \\
\quad HellaSwag & \textbf{81.94} & 68.89 & 68.79 & \underline{73.95} & 57.25 & 70.24 & 65.03 \\
\quad MMLU & \underline{84.05} & 80.89 & \textbf{84.42} & 82.80 & 78.01 & 82.08 & 80.40 \\
\midrule
\multicolumn{8}{l}{\textbf{Math}} \\
\quad GSM8k & 83.62 & 88.78 & 82.26 & 78.47 & \underline{90.37} & \textbf{93.71} & \underline{90.37} \\
\quad MATH-500 & \underline{83.80} & 82.00 & 83.60 & 82.20 & \textbf{90.00} & 70.60 & 83.20 \\
\quad AMC-23 & \underline{69.38} & 67.34 & 67.34 & 68.75 & \textbf{77.81} & 39.38 & 69.06 \\
\quad AIME-24 & 23.75 & \underline{27.71} & 17.29 & 17.92 & 27.50 & 12.92 & \textbf{27.92} \\
\quad AIME-25 & 16.67 & \underline{19.79} & 15.21 & 11.46 & \textbf{22.71} & 1.25 & 8.96 \\
\midrule
\multicolumn{8}{l}{\textbf{Science}} \\
\quad GPQA & \textbf{41.53} & 30.20 & \underline{37.67} & 34.31 & 36.49 & 31.99 & 31.80 \\
\quad GPQA\_Diamond & 49.66 & \underline{49.49} & 44.95 & 40.74 & 47.47 & 42.09 & \textbf{51.18} \\
\quad MMLU-Pro & \textbf{58.73} & 54.68 & \underline{56.63} & 56.35 & 47.81 & 53.29 & 55.58 \\
\quad MMLU-stem & \textbf{83.57} & 81.64 & \underline{82.59} & 82.37 & 73.55 & 74.88 & 75.20 \\
\midrule
\multicolumn{8}{l}{\textbf{Code}} \\
\quad HumanEval & 87.20 & \textbf{90.85} & 87.20 & \underline{90.24} & 86.59 & 83.53 & 85.40 \\
\quad HumanEval+ & 81.71 & \textbf{85.37} & 80.49 & \underline{82.32} & 78.05 & 79.87 & 78.70 \\
\quad MBPP & 83.86 & 86.24 & \textbf{89.68} & 87.83 & \underline{88.36} & 88.09 & 81.50 \\
\quad MBPP+ & 71.43 & 71.96 & \textbf{75.40} & \underline{74.07} & \underline{74.07} & 73.81 & 64.80 \\
\quad LiveCodeBench & \underline{49.71} & 45.01 & \textbf{54.60} & 49.12 & 39.53 & 40.31 & 40.12 \\
\quad CRUXEval & \underline{73.07} & \textbf{78.45} & 75.63 & 73.50 & 74.82 & 69.53 & 68.32 \\
\midrule
\multicolumn{8}{l}{\textbf{Instruction Following}} \\
\quad IFEval & \underline{89.37} & 86.97 & 86.35 & 81.79 & 83.19 & \textbf{89.94} & 86.32 \\
\quad Alpaca-Eval & 48.32 & \textbf{64.21} & 49.29 & 39.26 & \underline{56.16} & 38.27 & 36.26 \\
\quad MTBench & \textbf{9.20} & 9.05 & \underline{9.16} & 9.09 & 8.75 & 8.98 & 8.98 \\
\quad LiveBench & 46.26 & \textbf{63.05} & \underline{54.03} & 52.92 & 55.41 & 53.11 & 54.21 \\
\midrule
\multicolumn{8}{l}{\textbf{Multilingual}} \\
\quad Multi-Hellaswag & \textbf{74.55} & 58.39 & \underline{69.48} & 65.90 & 69.64 & 62.30 & 64.64 \\
\quad Multi-MMLU & \underline{77.76} & 66.20 & \textbf{78.26} & 73.56 & 71.60 & 74.58 & 76.67 \\
\quad MGSM & 76.33 & 71.80 & 72.33 & 73.60 & \underline{77.87} & 83.87 & \textbf{86.87} \\
\bottomrule
\end{tabular*}
\caption{Performance of the 34B-scale \textbf{Instruct} models.}
\label{tab:34b-ins}
\end{table}

\noindent \textbf{Long-Context Capabilities of Falcon-H1.}
Given the efficiency benefits of Mamba-based architectures in long-context scenarios, a systematic evaluation of Falcon-H1's capabilities in this area is crucial. We benchmarked Falcon-H1-34B-Instruct against several larger models using the HELMET suite~\citep{yen2025helmet}, with category-level results presented in Table~\ref{tab:helmet-avg-styled} and a full breakdown in Appendix~\ref{eval:tables_long_context}.
The findings highlight Falcon-H1's strong, parameter-efficient performance, particularly in the demanding Retrieval-Augmented Generation (RAG) task at extreme context lengths. While remaining competitive with 70B-class models at shorter sequences, our model uniquely achieves the top score on RAG at 131k tokens (62.21), surpassing all competitors, including those more than twice its size. This suggests a superior ability to synthesize information from retrieved documents, a key capability for practical RAG systems.
The results also reveal a nuanced performance profile. On tasks reliant on pure recall and long-form question answering (longQA), Falcon-H1 is surpassed by competitors like Qwen3-32B and Llama-3.3-70B at extreme context lengths. We attribute this performance gap not to architectural limitations but to our training data composition, which indicates substantial room for improvement with more curated long-context data. Crucially, despite this trade-off, Falcon-H1 still broadly outperforms the much larger Qwen2.5-72B-Instruct model across most long-context tasks. This, combined with its state-of-the-art RAG performance, positions Falcon-H1-34B as a highly effective and parameter-efficient choice for real-world, long-context systems.

\begin{table}[htbp]
\centering
\footnotesize
\begin{tabular*}{\textwidth}{l @{\extracolsep{\fill}} *{4}{S[table-format=3.2]}}
\toprule
\textbf{Seq. Length} & {\makecell{Falcon-H1- \\ 34B-Instruct}} & {\makecell{Qwen2.5-72B- \\ Instruct}} & {\makecell{Qwen3- \\ 32B}} & {\makecell{Llama-3.3-70B- \\ Instruct}} \\
\midrule
\multicolumn{5}{l}{\textbf{HELMET-RAG}} \\
\quad 8k    & 72.17 & \underline{72.21} & 69.25 & \textbf{74.29} \\
\quad 16k   & \underline{81.46} & 80.42 & 77.92 & \textbf{82.33} \\
\quad 32k   & 67.96 & \underline{70.08} & 64.83 & \textbf{70.21} \\
\quad 65k   & \underline{67.08} & 63.25 & 61.96 & \textbf{69.08} \\
\quad 131k  & \textbf{62.21} & 42.33 & \underline{57.08} & 55.38 \\
\midrule
\multicolumn{5}{l}{\textbf{HELMET-Recall}} \\
\quad 8k    & \bfseries 100.00 & \bfseries 100.00 & \bfseries 100.00 & \bfseries 100.00 \\
\quad 16k   & \bfseries 100.00 & \bfseries 100.00 & \bfseries 100.00 & \bfseries 100.00 \\
\quad 32k   & 97.50  & 98.38 & \textbf{100.00} & \underline{99.63} \\
\quad 65k   & 80.69  & 71.75 & \underline{96.50} & \textbf{98.81} \\
\quad 131k  & 56.63  & 38.81 & \textbf{86.13} & \underline{82.19} \\
\midrule
\multicolumn{5}{l}{\textbf{HELMET-longQA}} \\
\quad 8k    & 32.87 & \textbf{35.20} & 31.63 & \underline{33.67} \\
\quad 16k   & 34.64 & \underline{39.13} & 35.68 & \textbf{39.75} \\
\quad 32k   & 35.09 & 39.22 & \underline{41.15} & \textbf{47.53} \\
\quad 65k   & 32.45 & 36.71 & \underline{47.47} & \textbf{48.57} \\
\quad 131k  & 33.81 & 32.94 & \textbf{53.52} & \underline{46.06} \\
\bottomrule
\end{tabular*}
\caption{HELMET performance metrics at various sequence lengths. The best result in each row is in bold, and the second-best is underlined.}
\label{tab:helmet-avg-styled}
\end{table}

\noindent \textbf{Overall Remarks on Falcon-H1 Instruct models.}
The Falcon-H1 instruct series demonstrates a consistent pattern of state-of-the-art performance and exceptional parameter efficiency across all evaluated scales. Our smaller models, particularly the Falcon-H1-1.5B-Deep-Instruct, redefine the performance baseline in their class, delivering reasoning capabilities in math and science that are competitive with leading 7B and 10B models. This makes them powerful and efficient alternatives for resource-constrained and edge environments.
As we scale up, the Falcon-H1-7B and 34B models establish themselves as leaders in knowledge-intensive and reasoning-focused domains. They consistently excel in science, code generation, and multilingual understanding, often outperforming competitor models with more than double their parameter count. Furthermore, the Falcon-H1-34B shows a distinct advantage in practical long-context applications, leading on the 131k RAG benchmark. While larger competitor models show an edge in certain math and preference-based tasks, the Falcon-H1 series consistently provides a superior balance of deep reasoning, broad knowledge, and high efficiency, positioning it as a premier choice for sophisticated, real-world applications.

\subsection{Model Efficiency}
\label{sec:model_efficiency}
We conducted a comparative evaluation of prefill (input) and generation (output) throughput between Falcon-H1-34B and Qwen2.5-32B\footnote{Experiments were conducted prior to the release of Qwen3; however, we anticipate no significant efficiency differences between Qwen2.5-32B and Qwen3-32B.}. All experiments were conducted on H100 GPUs using the vLLM framework~\citep{kwon2023efficient} with a tensor parallel size of 2. The performance of each phase was measured as follows:
\begin{itemize}
    \item \textbf{Prefill throughput test}: Input sequence length was varied (2k to 262k tokens), while the output was fixed at 2,048 generated tokens per sequence with a batch size of 32.
    \item \textbf{Generation Throughput Test}: Input sequence length was fixed at 4,096 tokens with a batch size of 32, while the output generation length was varied (2k to 262k tokens).
\end{itemize}

\begin{figure}[ht] 
    \centering 
    \includegraphics[width=0.9\textwidth]{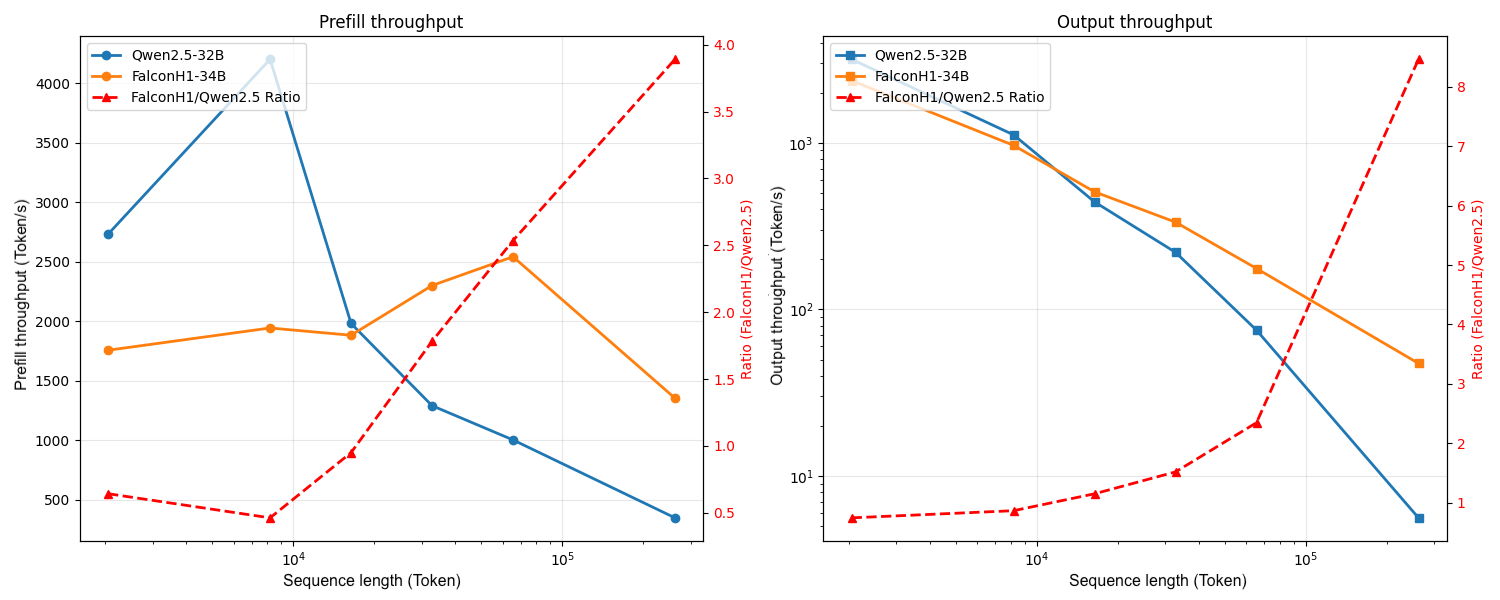}
    \caption{Model efficiency comparison between Falcon-H1-34B and Qwen2.5-32B.}
    \label{fig:throughput_comparison}
\end{figure}


As shown in Table \ref{fig:throughput_comparison}, the results demonstrate the superior scalability of the Falcon-H1 hybrid architecture. While Qwen2.5-32B exhibits a marginal throughput advantage at shorter context lengths, Falcon-H1-34B becomes significantly more efficient as the context grows. At the longest sequence lengths tested, Falcon-H1-34B achieves up to a \textbf{4x improvement in input throughput} and an \textbf{8x speedup in output throughput}. This performance profile makes the model exceptionally well-suited for long-context input and generation use cases.
The initial advantage of the Transformer-based model at short contexts is likely attributable to the highly mature optimizations of attention mechanisms within modern inference frameworks compared to current State-Space Model (SSM) implementations. As theoretically, Mamba-based or hybrid architectures are more efficient, we believe this gap highlights a promising direction for future work. We invite the community to contribute to optimizing SSM implementations, which we see as a critical step in advancing the next generation of efficient large language models.

\section{Model Integrations}
\label{sec:model_integrations}

To ensure broad accessibility and facilitate immediate adoption, Falcon-H1 is deeply integrated into the open-source AI ecosystem. Table~\ref{tab:integrations} summarizes the key platforms and tools supported at the time of this report's release. This list is continually expanding, with the most up-to-date information available on our project website\footnote{\url{https://tiiuae.github.io/Falcon-H1/}}.

\begin{table}[htbp]
\centering
\small
\begin{threeparttable}
\begin{tabularx}{\textwidth}{l >{\raggedright\arraybackslash}X}
\toprule
\textbf{Category} & \textbf{Supported Tools and Platforms} \\
\midrule
\textbf{General Usage} & \texttt{vLLM}~\citep{kwon2023efficient}, Hugging Face (\texttt{transformers}~\citep{Wolf_Transformers_State-of-the-Art_Natural_2020}, \texttt{PEFT}~\citep{peft}, \texttt{TRL}~\citep{von_Werra_TRL_Transformer_Reinforcement}) \\
\addlinespace
\textbf{Fine-tuning} & \texttt{Llama-Factory}~\citep{Zheng_LlamaFactory_Unified_Efficient_2024}, \texttt{OUMI}~\citep{Oumi_Community_Oumi_an_Open}, \texttt{Axolotl}\tnote{a}, \texttt{Unsloth}~\citep{unsloth} \\
\addlinespace
\textbf{Local Deployment} & \texttt{llama.cpp}\tnote{b}, \texttt{LM-Studio}\tnote{c}, \texttt{Jan}\tnote{d}, \texttt{Docker Model API}\tnote{e}, \texttt{Ollama}\tnote{f, *}, \texttt{Apple MLX}\tnote{g, *} \\
\addlinespace
\textbf{Cloud Deployment} & \texttt{SkyPilot}\tnote{h} \\
\addlinespace
\textbf{Quantization} & \texttt{AutoGPTQ}~\citep{qubitium2024gptqmodel} \\
\bottomrule
\end{tabularx}
\begin{tablenotes}[para,flushleft]
    \footnotesize 
    \item[*] Integrations validated internally; official support pending merge into main libraries at the time of the report release.
    \item[a] \url{https://github.com/axolotl-ai-cloud/axolotl}
    \item[b] \url{https://github.com/ggml-org/llama.cpp}
    \item[c] \url{https://lmstudio.ai/}
    \item[d] \url{https://github.com/menloresearch/jan}
    \item[e] \url{https://docs.docker.com/ai/model-runner/}
    \item[f] \url{https://github.com/ollama/ollama}
    \item[g] \url{https://github.com/ml-explore/mlx-lm}
    \item[h] \url{https://github.com/skypilot-org/skypilot}
\end{tablenotes}
\end{threeparttable}
\caption{Key Ecosystem Integrations for the Falcon-H1 Series.}
\label{tab:integrations}
\end{table}








\section{Conclusion}
In this report, we introduced the Falcon-H1 series, a new family of models built on an innovative hybrid Mamba-Transformer architecture. Our design goal was to achieve state-of-the-art performance with exceptional resource efficiency, and our comprehensive evaluations confirm the success of this approach.
A key advantage of Falcon-H1 is its ability to deliver superior performance while using significantly less training data—only 2.5T to 18T tokens—and offering up to 8x faster inference in long-context scenarios. This performance gain is particularly impactful at smaller scales, where our 1.5B-Deep model delivers capabilities competitive with leading 7B-10B models, making it ideal for edge deployments. At the larger end, our flagship 34B model challenges and often surpasses 70B+ competitors, particularly on knowledge-intensive tasks and practical applications like RAG at extreme context lengths.
The success of this series is rooted in several key innovations: a flexible hybrid architecture allowing for an optimal attention-SSM ratio, a robust multilingual design, and a customized training strategy that maximizes the value of high-quality data. We also observed that the models' performance had not yet saturated by the end of pretraining, indicating significant headroom for future gains. Ultimately, by providing a powerful, efficient, and versatile foundation for a wide range of applications, the Falcon-H1 series demonstrates a more sustainable and accessible path toward developing high-performance artificial intelligence.

Our future work will focus on several primary areas. First, we will prioritize data enhancement. We plan to iteratively refine both our base and instruction-tuned models by curating broader and more diverse high-quality datasets. 
Second, we will continue our architectural and algorithmic innovation, with a focus on effective knowledge compression and scaling to contexts beyond 256k. Finally, we will continue to scale our models strategically, exploring novel techniques to deepen their core reasoning capabilities. Through these efforts, we aim to continue developing models that are not only state-of-the-art but also fundamentally more efficient and accessible, thereby contributing to the sustainable advancement of artificial intelligence.

\section{Authors}

\noindent \textbf{Core Contributors}\\
Jingwei Zuo,  
Maksim Velikanov,  
Ilyas Chahed,  
Younes Belkada,  
Dhia Eddine Rhayem,  
Guillaume Kunsch\textsuperscript{*},  
Hakim Hacid  

\vspace{1em}
\noindent \textbf{Contributors}
\begin{itemize}
    \item \textbf{Quantization and Integration:}  
    Hamza Yous,  
    Brahim Farhat,  
    Ibrahim Khadraoui

    \item \textbf{Data \& Evaluation:}  
    Mugariya Farooq,  
    Giulia Campesan,  
    Ruxandra Cojocaru,  
    Yasser Djilali,  
    Shi Hu,
    Iheb Chaabane,  
    Puneesh Khanna,  
    Mohamed El Amine Seddik,  
    Ngoc Dung Huynh,
    Phuc Le Khac,
    Leen AlQadi,
    Billel Mokeddem,  
    Mohamed Chami,  
    Abdalgader Abubaker

    \item \textbf{Platform \& Infrastructure Support:}  
    Mikhail Lubinets,  
    Kacper Piskorski,  
    Slim Frikha
\end{itemize}

\vspace{0.5em}
\noindent \textsuperscript{*}\textit{Individual who has departed from our team; work was conducted at TII.}
\section{Acknowledgments}
We extend our sincere gratitude to the contributors and maintainers from the open-source AI community who were instrumental in integrating Falcon-H1 into the ecosystem. For their invaluable help with fine-tuning libraries, we thank Wing Lian for the support on Axolotl, Daniel Han for the support on Unsloth, Yaowei Zheng for the support on Llama-Factory, and the OUMI maintainers for their assistance.
For their support with local deployment tools, we are grateful to Prince Canuma and Awni Hannun for their work on the Apple MLX integration, and to Georgi Gerganov, \texttt{compilade}, and Gabe Goodhart for their extensive efforts on llama.cpp and Ollama integration. We also thank the Jan maintainers for their help.
Finally, for their crucial support on core libraries, we thank Arthur Zucker for his guidance on Hugging Face Transformers integration and the vLLM maintainers for their help in incorporating Falcon-H1 into their high-performance inference library.

\vspace{1em}
\bibliographystyle{iclr2025_conference}
\bibliography{references}
\newpage
\appendix
\section{Languages used for training Falcon-H1 tokenizers}

\begin{table}[htbp]
\centering
\scriptsize
\setlength{\tabcolsep}{4pt}
\begin{tabular}{|c|l|l||c|l|l||c|l|l|}
\hline
\textbf{\#} & \textbf{Code} & \textbf{Language} & 
\textbf{\#} & \textbf{Code} & \textbf{Language} & 
\textbf{\#} & \textbf{Code} & \textbf{Language} \\
\hline
1 & af & Afrikaans & 2 & als & Swiss German & 3 & am & Amharic \\
4 & an & Aragonese & 5 & ar & Arabic & 6 & arz & Egyptian Arabic \\
7 & as & Assamese & 8 & ast & Asturian & 9 & av & Avaric \\
10 & az & Azerbaijani & 11 & azb & South Azerbaijani & 12 & ba & Bashkir \\
13 & bar & Bavarian & 14 & bcl & Central Bikol & 15 & be & Belarusian \\
16 & bg & Bulgarian & 17 & bh & Bihari languages & 18 & bn & Bangla \\
19 & bo & Tibetan & 20 & bpy & Bishnupriya & 21 & bs & Bosnian \\
22 & bxr & Russia Buriat & 23 & ca & Catalan & 24 & ce & Chechen \\
25 & ckb & Central Kurdish & 26 & cs & Czech & 27 & cv & Chuvash \\
28 & cy & Welsh & 29 & da & Danish & 30 & de & German \\
31 & dsb & Lower Sorbian & 32 & dv & Divehi & 33 & el & Greek \\
34 & eml & Emiliano-Romagnol & 35 & eo & Esperanto & 36 & es & Spanish \\
37 & et & Estonian & 38 & eu & Basque & 39 & fa & Persian \\
40 & fi & Finnish & 41 & fr & French & 42 & fy & Western Frisian \\
43 & ga & Irish & 44 & gd & Scottish Gaelic & 45 & gl & Galician \\
46 & gn & Guarani & 47 & gom & Goan Konkani & 48 & gu & Gujarati \\
49 & he & Hebrew & 50 & hi & Hindi & 51 & hr & Croatian \\
52 & hsb & Upper Sorbian & 53 & ht & Haitian Creole & 54 & hu & Hungarian \\
55 & hy & Armenian & 56 & ia & Interlingua & 57 & id & Indonesian \\
58 & ie & Interlingue & 59 & ilo & Iloko & 60 & io & Ido \\
61 & is & Icelandic & 62 & it & Italian & 63 & ja & Japanese \\
64 & jbo & Lojban & 65 & jv & Javanese & 66 & ka & Georgian \\
67 & kk & Kazakh & 68 & km & Khmer & 69 & kn & Kannada \\
70 & ko & Korean & 71 & krc & Karachay-Balkar & 72 & ku & Kurdish \\
73 & kv & Komi & 74 & kw & Cornish & 75 & ky & Kyrgyz \\
76 & la & Latin & 77 & lb & Luxembourgish & 78 & lez & Lezghian \\
79 & li & Limburgish & 80 & lmo & Lombard & 81 & lo & Lao \\
82 & lt & Lithuanian & 83 & lv & Latvian & 84 & mai & Maithili \\
85 & mg & Malagasy & 86 & mk & Macedonian & 87 & ml & Malayalam \\
88 & mn & Mongolian & 89 & mr & Marathi & 90 & mrj & Western Mari \\
91 & ms & Malay & 92 & mt & Maltese & 93 & mwl & Mirandese \\
94 & my & Burmese & 95 & myv & Erzya & 96 & mzn & Mazanderani \\
97 & nah & Nahuatl languages & 98 & nap & Neapolitan & 99 & nds & Low German \\
100 & ne & Nepali & 101 & new & Newari & 102 & nl & Dutch \\
103 & nn & Norwegian Nynorsk & 104 & no & Norwegian & 105 & oc & Occitan \\
106 & or & Odia & 107 & os & Ossetic & 108 & pa & Punjabi \\
109 & pam & Pampanga & 110 & pl & Polish & 111 & pms & Piedmontese \\
112 & pnb & Western Panjabi & 113 & ps & Pashto & 114 & pt & Portuguese \\
115 & qu & Quechua & 116 & ro & Romanian & 117 & ru & Russian \\
118 & sv & Swedish & 119 & th & Thai & 120 & tr & Turkish \\
121 & vi & Vietnamese & & & & & &\\
\hline
\end{tabular}
\caption{Language Codes and Corresponding Languages}
\label{tab:language_codes}
\end{table}

\section{Scalar stochastic dynamics with weight decay}\label{sec:toy_model_for_param_norms}
To model the behavior of parameter norms $||W||^2=\sum_{ij} W_{ij}^2$ we look at a single entry $W_{ij}=x$ and consider its evolution during training. Then, behavior of the typical values of $x^2$ could serve as a proxy for behavior of the parameter norms $||W||^2$. 

The update of parameter $x$ at iteration $t$ using AdamW with learning rate $\eta$ and weight decay $\lambda$ can be written as
\begin{equation}\label{eq:single_parameter_update}
    x_{t+1} = x_t - \eta A_t - \eta \lambda x_t.
\end{equation}
Here $A_t = \frac{g_{1,t}}{\sqrt{g_{2,t}} + \varepsilon}$ is the Adam update rule that uses moving average of the gradient $g_{1}$ and moving average (element-wise) squared gradient $g_2$. 

Generally, updates $A_t$ depend on the dynamics of the rest of model beyond our selected parameter $x$, have stochastic nature w.r.t. randomness caused by sampling of the samples in a batch, and correlated across nearby iterations $t$ due to moving averages used in the update rule. Yet, to get an intuitive picture of the role of AdamW parameters $\eta,\lambda$ we consider a toy model of this update
\begin{equation}\label{eq:linear_iid_update}
    A_t = h(x_t-x^*) + \xi_t,
\end{equation}
where $\xi_t$ is i.i.d. noise with zero mean and variance $\mathbb{E}[\xi_t^2]=\sigma^2$, and $h$ describes steepness of the loss landscape for our chosen parameter $x$. Then, evolution \eqref{eq:single_parameter_update} of $x_t$ becomes a stationary linear stochastic equation. 

Let us now track the evolution of the first two moments of $x_t$. Taking the expectation of \eqref{eq:single_parameter_update} and its square we get the update of the moments
    \begin{align}
    \mathbb{E}[x_{t+1}] &= (1-\eta h -\eta \lambda)\mathbb{E}[x_{t}] + \eta h x^*\\
    \mathbb{E}[x_{t+1}^2] &= (1-\eta h-\eta\lambda)^2\mathbb{E}[x_{t}^2] + 2(1-\eta h-\eta\lambda)\eta h \mathbb{E}[x_{t}]x^* + (\eta h x^*)^2 + \eta^2 \sigma^2
    \end{align}
As $t\to\infty$, the dynamics of first and second moment converge to the stationary state $\mathbb{E}[x_t]\to x_\infty$, $\mathbb{E}[x_t^2]\to x_{2,\infty}$ that can be found by setting $\mathbb{E}[x_{t+1}]=\mathbb{E}[x_{t}]=x_\infty$ and $\mathbb{E}[x_{t+1}^2]=\mathbb{E}[x_{t}^2]=x_{2,\infty}$. With a direct calculation, we get
\begin{align}
    x_\infty &= \frac{h}{h+\lambda}x^*,\\
    x_{\infty,2} &= \frac{\eta \sigma^2}{(\lambda+h)(2-\eta\lambda-\eta h)} + x_\infty^2.
\end{align}

Now, let us try to map the toy model described above to the scenario of LLM training, where we find a substantial simplification of the second moment in the stationary state. Typical values of learning rate and weight decay used in pretraining are $\eta \lesssim 10^{-3}$ and $\lambda \approx 0.1$. These values imply that the product $\eta\lambda$ is very small, and we can drop it compared to the terms that are of the order of $1$. It is also reasonable to assume that $h\ll1$, which also implies $\frac{h}{\lambda}\ll1$ for practical values of $\lambda$. Indeed, the steepness $h$ roughly corresponds to quadratic approximation of the loss w.r.t. to the chosen parameter $L(x)\approx\frac{1}{2}h (x-x^*)^2$. Then, due to large number of parameters in the model, the sensitivity to a single parameter $x$ should be relatively small, implying $h\ll1$. Having two assumptions $\eta\lambda \ll 1$ and $\frac{h}{\lambda}\ll1$ the second moment simplifies to
\begin{equation}
    x_{\infty,2} \approx \frac{1}{2} \frac{\eta}{\lambda} \left(\sigma^2+ \frac{2(hx^*)^2}{\eta\lambda}\right).
\end{equation}
The obtained result for $x_{\infty,2}$ has a clear interpretation as it fully separates signal and noise contributions. The first term describes the balance between weight decay contraction and Brownian motion expansion due to the noise in the updates. The second term describes the balance between, again, weight decay contraction, and attraction of the parameter $x$ to its optimal value $x^*$. Importantly, these two terms have a different scaling w.r.t. learning rate $\eta$ and weight decay $\lambda$. 

\paragraph{Implication for the parameter norms.} Recall that our parameter $x$ was just a single entry of the parameter matrix $W$. Different entries would have its own values of $\sigma,h,x^*$ but the same $\eta,\lambda$ as those are global hyperparameters of the training algorithm. Then, we could write the dependence of parameters norm $||W||^2$ on learning rate and weight decay as
\begin{equation}
    ||W||^2 = \sum_{ij} W_{ij}^2 = \left(\frac{\sigma_{ij}^2}{2}\right)\frac{\eta}{\lambda} + \left(\sum_{ij}(h_{ij}x^*_{ij})^2\right)\frac{1}{\lambda^2} = C_1 \frac{\eta}{\lambda} + C_2 \frac{1}{\lambda^2}.
\end{equation}
As the dependence of the parameter norms on $\eta,\lambda$ can be accurately measured for the actual LLM training, we could check which of the two scalings more accurately describes the experimental data. As we have seen in section \ref{sec:param_norms_and_schedules}, the scaling $||W||^2\propto \frac{\eta}{\lambda}$ very well captures the experiments values, implying $C_1 \frac{\eta}{\lambda} \gg C_2 \frac{1}{\lambda^2}$ for typical values of $\eta,\lambda$. This suggest that, at least for the most of the parameters, the noise in the updates dominates the attraction to the optimal values.

\section{\texorpdfstring{Tuning $\mu P$ multipliers}{Tuning muP multipliers}}
\label{sec:mup_tuning}
To jointly tune all of our $M=35$ multipliers, we have used a stagewise procedure comprised of micro-sweeps over each multiplier on each stage. With a slight abuse of notations, denote $\mathcal{M}_n=\{m^{(1)}_n, \ldots, m^{(M)}_n\}$ the set of all $\mu$P multipliers at stage $n$. At each stage, we repeat the following steps:
\begin{enumerate}
    \item Start the stage by running a training job with $\mathcal{M}_n$, and measure the final loss $L_n$ that will serve as a baseline loss for the current stage.
    \item For each multiplier $i=1\ldots k$ run a log-scale microsweep: two training jobs with increased $i$'th multiplier $m^{(i)}_{n,+}=p m^{(i)}_{n}$ and decreased $m^{(i)}_{n,-}=p^{-1} m^{(i)}_{n}$, and measure the respective losses $L^{(i)}_{n,+}$ and $L^{(i)}_{n,-}$. The scaling factor $p$ is chosen to balance exploration and precision. We have used $p=2$ at the beginning of the tuning to converge faster to the vicinity of the optimal set of multipliers, while switching to $p=\sqrt{2}$ towards the end of the tuning for the increased precision.  
    \item For each multiplier $m^{(i)}_n$ we manually inspected decreased, baseline, and increased losses $(L^{(i)}_{n,-}, L_n, L^{(i)}_{n,+})$ to pick the value of the next stage multiplier $m^{(i)}_{n+1}$. Most of the time we pick this value from $\{m^{(i)}_{n, -}, m^{(i)}_{n}, m^{(i)}_{n, +}\}$ depending on which of the respective losses is lower. However, sometimes we have also picked other values, for example, at earlier stages we could pick the value beyond maximal $m^{(i)}_{n,+}$ or minimal $m^{(i)}_{n,-}$ to increase exploration. Also, sometimes we picked an intermediate value, for example, $\sqrt{m^{{(i)}}_nm^{{(i)}}_{n_+}}$ if two losses $L_n, L^{(i)}_{n,+}$ turned out roughly the same and significantly better than $L^{(i)}_{n,-}$.
    \item Construct the next stage set of multipliers $\mathcal{M}_{n+1}=\{m^{(1)}_{n+1},\ldots,m^{(M)}_{n+1}\}$ from the values picked on the step 3, then go to step 1 to start stage $n+1$. 
\end{enumerate}

The above procedure is quite simple, and we expect it can be improved in the future to achieve a faster convergence to the optimal set of multipliers. However, the simplicity and manual elements of our procedure was important to build an intuition behind the roles of different multipliers in the training process, as we were also checking the whole training curve for spikes, the noise level (the loss difference before and after LR decay), and any possible anomalies in the training. 

Another benefit of our procedure is the possibility to measure the sensitivity of the loss to each multiplier. Indeed, at the end of the procedure we obtain a set of multipliers that is close to optimality, and the dependence of the loss on the logarithm of each multiplier can be roughly approximated by a quadratic function $L(m)\approx \frac{a}{2} (\log m-\log m^*)^2 + L^*$. We fit our three loss measurements $(L^{(i)}_{n,-}, L_n, L^{(i)}_{n,+})$ with this quadratic dependence to estimate the sensitivity $a=\frac{\partial^2}{(\partial \log m)^2}L$, reported in Figure \ref{fig:mup_multipliers}.     

\paragraph{Effective learning rate and weight decay.} Recall that we have 4 types of $\mu$P multipliers, grouped as in table \ref{tab:tuned_multipliers}
\begin{enumerate}
    \item Forward multipliers $m^{(i)}$.
    \item LR multipliers $\eta^{(j)}/\eta$ for matrix-like layers with enabled weight decay. 
    \item WD multipliers $\lambda^{(j)}/\lambda$ for the same matrix-like layers with enabled weight decay.
    \item LR multipliers $\eta^{(k)}/\eta$ for vector-like layers with disabled weight decay. 
\end{enumerate}

Our tuning procedure in the space of multipliers has a coordinate-aligned structure, similar to coordinate descent. Therefore, it is beneficial for its convergence and interpretation to make different coordinates as independent as possible. We expect that forward multipliers and LR multipliers for vector-like layers are already quite independent from each other. However, as discussed in section \ref{sec:param_norms_and_schedules}, LR and WD multipliers for the same matrix-like layer are expected to be strongly coupled. 

We aim to decrease the coupling by changing the coordinate axes to ELR and EWD \eqref{eq:ELR_EWD_def}, and tuning ELR multiplier $\sqrt{\eta^{(j)}\lambda^{(j)}} / \sqrt{\eta\lambda}$ and EWD multiplier $\sqrt{\frac{\lambda^{(j)}}{\eta^{(j)}}}\big/\sqrt{\frac{\lambda}{\eta}}$ instead of plane LR and WD multipliers. Specifically, for ELR micro-sweep at stage $n$ we change  LR and WD values as $(\eta^{(j)}_n,\lambda^{(j)}_n)\to(p\eta^{(j)}_n,p\lambda^{(j)}_n)$, and for EWD micro-sweep as $(\eta^{(j)}_n,\lambda^{(j)}_n)\to(p\eta^{(j)}_n,p^{-1}\lambda^{(j)}_n)$. Accordingly, in figure \ref{fig:mup_multipliers} we also report sensitivities $\frac{\partial^2}{(\partial \log \eta_\mathrm{eff})^2}L$ and $\frac{\partial^2}{(\partial \log \lambda_\mathrm{eff})^2}L$ instead of the raw LR/WD sensitivities.       

\paragraph{Other settings, and applied scaling relations.} The shapes of the model used for multiplier tuning are reported in table \ref{tab:forward_multipliers}. As for the training job settings, we have used WSD learning rate schedule with 65GT of the stable stage, and 10GT of exponential decay that reduces LR $32$ times. The global LR and WD values are $\eta=256\times10^{-6}$ and $\lambda=0.1$, sequence length $2048$, and the global batch size of $4$ million tokens (MT). 

As we expect all the hyperparameters to be well-tuned by the end of multiplier tuning, it is essential to correctly transfer them to other model and training settings. Here are the scaling relations we have applied for the Falcon-H1 training. 
\begin{itemize}
    \item For the $\mu$P multipliers, we have the scaling version that scales forward multipliers, as described in table \ref{tab:forward_multipliers}, while keeping LR and WD values unchanged. We did not use scaling of multipliers with model depth $L$, for simplicity and due to time constraints, but it can be directly applied following recent work~\citep{yang2023tensorprogramsvifeature,dey2025dontlazycompletepenables}.
    \item When changing the batch size, we have used the square root scaling of the learning rate \eqref{eq:batch_scaling}.
    \item When changing the total training duration, we have applied our effective power scheduler (EPS) for the learning rate and weight decay of the stable stage of WSD, as described in section \ref{sec:param_norms_and_schedules}. The activation is the power scheduler should be set close to the duration of the stable stage our tuning jobs - 65GT. However, EPS can be activated later to ensure higher adaptivity in the later stages of training, for example, if curriculum learning or long context extension at the end of the training are used. 
\end{itemize}

\section{Detailed evaluation results}

\subsection{Multilingual Evaluations - Base Models}
\label{eval:tables_ml_base}

\begin{table}[htbp]
\centering
\footnotesize
\begin{tabular*}{\textwidth}{l @{\extracolsep{\fill}} *{7}{S[table-format=2.2]}}
\toprule
\textbf{Task} & {\makecell{Falcon-H1- \\ 1.5B-deep}} & {\makecell{Falcon-H1- \\ 1.5B}} & {\makecell{Qwen3- \\ 1.7B}} & {\makecell{Qwen2.5- \\ 1.5B}} & {\makecell{Gemma3- \\ 1B}} & {\makecell{Llama3.2- \\ 1.2B}} & {\makecell{Falcon3- \\ 1.6B}} \\
\midrule
\multicolumn{8}{l}{\textbf{Multilingual Hellaswag}} \\
\quad Arabic (ar) & 41.63 & 39.01 & 40.93 & 38.55 & 40.62 & 34.96 & 28.69 \\
\quad German (de) & 50.44 & 46.55 & 47.18 & 43.48 & 45.56 & 41.47 & 30.82 \\
\quad Spanish (es) & 58.52 & 54.69 & 54.20 & 51.98 & 52.22 & 48.13 & 37.49 \\
\quad French (fr) & 57.25 & 53.68 & 53.07 & 50.30 & 50.93 & 46.07 & 37.45 \\
\quad Hindi (hi) & 35.61 & 33.22 & 34.01 & 30.43 & 35.63 & 32.97 & 28.26 \\
\quad Italian (it) & 54.39 & 50.21 & 50.53 & 46.36 & 49.11 & 44.23 & 31.84 \\
\quad Dutch (nl) & 50.89 & 47.09 & 45.46 & 41.81 & 47.24 & 42.22 & 29.99 \\
\quad Portuguese (pt) & 56.57 & 51.99 & 52.41 & 50.60 & 50.62 & 45.84 & 33.69 \\
\quad Romanian (ro) & 48.17 & 43.74 & 42.86 & 35.71 & 43.46 & 39.04 & 29.63 \\
\quad Russian (ru) & 51.20 & 48.04 & 48.16 & 45.31 & 45.77 & 42.12 & 29.08 \\
\quad Swedish (sv) & 49.32 & 44.53 & 42.36 & 37.25 & 46.40 & 40.69 & 28.65 \\
\quad \textbf{Average} & \textbf{50.36} & \underline{46.62} & 46.47 & 42.89 & 46.14 & 41.61 & 31.42 \\
\midrule
\multicolumn{8}{l}{\textbf{Multilingual MMLU}} \\
\quad Arabic (ar) & 43.92 & 40.17 & {-} & 42.67 & 26.30 & 26.03 & 27.29 \\
\quad German (de) & 54.24 & 48.49 & {-} & 49.88 & 27.27 & 28.68 & 32.94 \\
\quad Spanish (es) & 57.33 & 51.73 & {-} & 53.65 & 26.92 & 29.00 & 35.98 \\
\quad French (fr) & 56.54 & 50.66 & {-} & 52.67 & 25.73 & 28.33 & 35.22 \\
\quad Hindi (hi) & 40.69 & 37.72 & {-} & 34.47 & 26.04 & 27.24 & 27.93 \\
\quad Italian (it) & 55.11 & 48.46 & {-} & 50.84 & 25.90 & 28.27 & 32.28 \\
\quad Dutch (nl) & 53.21 & 46.54 & {-} & 48.31 & 26.98 & 29.00 & 30.67 \\
\quad Portuguese (pt) & 57.04 & 51.53 & {-} & 53.23 & 27.09 & 28.61 & 33.99 \\
\quad Romanian (ro) & 52.49 & 46.67 & {-} & 44.74 & 27.03 & 27.03 & 30.29 \\
\quad Russian (ru) & 50.35 & 44.71 & {-} & 47.57 & 26.36 & 28.50 & 28.50 \\
\quad Swedish (sv) & 51.92 & 44.53 & {-} & 45.16 & 25.90 & 28.78 & 29.76 \\
\quad Chinese (zh) & 51.23 & 46.87 & {-} & 53.94 & 26.49 & 29.25 & 33.89 \\
\quad \textbf{Average} & \textbf{52.00} & 46.51 & {-} & \underline{48.09} & 26.50 & 28.22 & 31.56 \\
\midrule
\multicolumn{8}{l}{\textbf{Multilingual GSM (MGSM)}} \\
\quad German (de) & 64.40 & 48.40 & {-} & 41.60 & {-} & 6.00 & 8.40 \\
\quad Spanish (es) & 68.00 & 61.20 & {-} & 54.40 & {-} & 6.00 & 14.40 \\
\quad French (fr) & 61.20 & 57.60 & {-} & 44.00 & {-} & 4.80 & 14.40 \\
\quad Japanese (ja) & 42.80 & 31.60 & {-} & 34.00 & {-} & 2.80 & 2.00 \\
\quad Russian (ru) & 64.40 & 54.00 & {-} & 44.40 & {-} & 3.60 & 3.60 \\
\quad Chinese (zh) & 61.20 & 52.00 & {-} & 52.40 & {-} & 5.20 & 13.60 \\
\quad \textbf{Average} & \textbf{60.33} & \underline{50.80} & {-} & 45.13 & {-} & 4.73 & 9.40 \\
\bottomrule
\end{tabular*}
\caption{Performance comparison of \textbf{Base} models on multilingual tasks (1B-2B scale). All scores are percentages. The best average score is in bold, and the second-best is underlined.}
\label{tab:multi-1b-base-styled}
\end{table}

\begin{table}[htbp]
\centering
\footnotesize
\begin{tabular*}{\textwidth}{l @{\extracolsep{\fill}} *{6}{S[table-format=2.2]}}
\toprule
\textbf{Task} & {\makecell{Falcon-H1- \\ 3B}} & {\makecell{Qwen3- \\ 4B}} & {\makecell{Qwen2.5- \\ 3B}} & {\makecell{Gemma3- \\ 4B}} & {\makecell{Llama-3.2- \\ 3B}} & {\makecell{Falcon3- \\ 3B}} \\
\midrule
\multicolumn{7}{l}{\textbf{Multilingual Hellaswag}} \\
\quad Arabic (ar) & 45.42 & 49.63 & 49.50 & 52.14 & 43.59 & 29.52 \\
\quad German (de) & 56.29 & 58.09 & 57.29 & 61.36 & 54.74 & 35.01 \\
\quad Spanish (es) & 62.98 & 64.69 & 60.15 & 67.09 & 61.75 & 48.95 \\
\quad French (fr) & 62.52 & 63.76 & 59.32 & 66.62 & 59.99 & 48.33 \\
\quad Hindi (hi) & 39.35 & 41.12 & 40.40 & 46.26 & 41.03 & 28.71 \\
\quad Italian (it) & 58.90 & 61.86 & 58.34 & 64.44 & 57.78 & 36.45 \\
\quad Dutch (nl) & 56.43 & 56.52 & 56.33 & 63.17 & 55.59 & 31.60 \\
\quad Portuguese (pt) & 61.39 & 63.25 & 59.85 & 65.85 & 59.40 & 47.73 \\
\quad Romanian (ro) & 53.09 & 53.68 & 52.76 & 60.48 & 50.63 & 31.21 \\
\quad Russian (ru) & 55.48 & 57.39 & 55.47 & 59.76 & 53.99 & 31.26 \\
\quad Swedish (sv) & 54.79 & 53.67 & 52.50 & 64.14 & 54.39 & 30.60 \\
\quad \textbf{Average} & 55.15 & \underline{56.69} & 54.71 & \textbf{61.03} & 53.89 & 36.30 \\
\midrule
\multicolumn{7}{l}{\textbf{Multilingual MMLU}} \\
\quad Arabic (ar) & 46.98 & 57.70 & 49.50 & 47.14 & 39.24 & 29.77 \\
\quad German (de) & 57.82 & 66.33 & 57.29 & 54.22 & 47.33 & 39.98 \\
\quad Spanish (es) & 60.08 & 68.65 & 60.15 & 55.68 & 49.06 & 47.47 \\
\quad French (fr) & 58.81 & 68.28 & 59.32 & 54.92 & 48.44 & 46.89 \\
\quad Hindi (hi) & 42.53 & 52.64 & 40.40 & 45.28 & 37.50 & 29.95 \\
\quad Italian (it) & 58.17 & 68.10 & 58.34 & 54.42 & 47.84 & 39.84 \\
\quad Dutch (nl) & 56.48 & 65.99 & 56.33 & 53.69 & 46.96 & 36.11 \\
\quad Portuguese (pt) & 59.37 & 68.79 & 59.85 & 54.97 & 48.45 & 46.56 \\
\quad Romanian (ro) & 55.54 & 65.50 & 52.76 & 53.99 & 45.69 & 35.91 \\
\quad Russian (ru) & 53.53 & 65.01 & 55.47 & 51.58 & 44.56 & 32.51 \\
\quad Swedish (sv) & 54.98 & 64.78 & 52.50 & 53.61 & 45.76 & 34.70 \\
\quad Chinese (zh) & 53.05 & 67.22 & 59.63 & 51.36 & 44.54 & 40.07 \\
\quad \textbf{Average} & 54.78 & \textbf{64.91} & \underline{55.13} & 52.57 & 45.45 & 38.31 \\
\midrule
\multicolumn{7}{l}{\textbf{Multilingual GSM (MGSM)}} \\
\quad German (de) & 64.80 & {-} & 58.80 & {-} & 23.20 & 28.80 \\
\quad Spanish (es) & 72.00 & {-} & 68.00 & {-} & 24.40 & 46.80 \\
\quad French (fr) & 70.40 & {-} & 62.40 & {-} & 22.00 & 48.00 \\
\quad Japanese (ja) & 51.20 & {-} & 45.60 & {-} & 12.80 & 8.80 \\
\quad Russian (ru) & 64.00 & {-} & 62.00 & {-} & 16.80 & 14.80 \\
\quad Chinese (zh) & 61.60 & {-} & 62.80 & {-} & 22.80 & 43.60 \\
\quad \textbf{Average} & \textbf{64.00} & {-} & \underline{59.93} & {-} & 20.33 & 31.80 \\
\bottomrule
\end{tabular*}
\caption{Performance comparison of \textbf{Base} models on multilingual tasks (3B-4B scale). All scores are percentages. The best average score is in bold, and the second-best is underlined.}
\label{tab:multi-3b-base-styled}
\end{table}

\begin{table}[htbp]
\centering
\footnotesize
\begin{tabular*}{\textwidth}{l @{\extracolsep{\fill}} *{7}{S[table-format=2.2]}}
\toprule
\textbf{Task} & {\makecell{Falcon-H1- \\ 7B}} & {\makecell{Qwen3- \\ 8B}} & {\makecell{Qwen2.5- \\ 7B}} & {\makecell{Gemma3- \\ 12B}} & {\makecell{Llama3.1- \\ 8B}} & {\makecell{Falcon3- \\ 7B}} & {\makecell{Falcon3- \\ 10B}} \\
\midrule
\multicolumn{8}{l}{\textbf{Multilingual Hellaswag}} \\
\quad Arabic (ar) & 54.57 & 54.24 & 52.23 & 62.62 & 50.14 & 31.11 & 34.21 \\
\quad German (de) & 66.28 & 64.09 & 60.58 & 71.81 & 63.13 & 44.52 & 51.18 \\
\quad Spanish (es) & 72.71 & 70.30 & 68.82 & 76.82 & 69.71 & 67.57 & 71.36 \\
\quad French (fr) & 71.73 & 68.86 & 67.89 & 75.32 & 67.95 & 66.96 & 70.25 \\
\quad Hindi (hi) & 47.65 & 45.82 & 38.72 & 53.57 & 45.75 & 30.15 & 31.09 \\
\quad Italian (it) & 69.71 & 66.86 & 63.57 & 74.01 & 66.37 & 51.34 & 56.61 \\
\quad Dutch (nl) & 67.47 & 62.51 & 60.76 & 73.36 & 64.21 & 40.14 & 45.87 \\
\quad Portuguese (pt) & 70.92 & 68.44 & 67.94 & 75.28 & 68.06 & 65.98 & 69.49 \\
\quad Romanian (ro) & 64.65 & 59.64 & 49.32 & 70.74 & 59.48 & 38.09 & 42.91 \\
\quad Russian (ru) & 64.52 & 62.39 & 61.60 & 68.95 & 60.25 & 37.55 & 43.17 \\
\quad Swedish (sv) & 66.61 & 60.32 & 54.65 & 74.31 & 63.91 & 38.91 & 43.85 \\
\quad \textbf{Average} & \underline{65.16} & 62.13 & 58.74 & \textbf{70.62} & 61.72 & 46.58 & 50.91 \\
\midrule
\multicolumn{8}{l}{\textbf{Multilingual MMLU}} \\
\quad Arabic (ar) & 60.93 & 62.06 & 58.19 & 61.96 & 46.98 & {-} & 35.33 \\
\quad German (de) & 69.19 & 70.11 & 65.69 & 68.32 & 55.48 & {-} & 55.57 \\
\quad Spanish (es) & 71.75 & 71.82 & 68.19 & 69.24 & 57.31 & {-} & 66.99 \\
\quad French (fr) & 70.85 & 71.56 & 67.88 & 69.27 & 56.93 & {-} & 67.09 \\
\quad Hindi (hi) & 57.06 & 57.92 & 49.65 & 59.03 & 43.93 & {-} & 35.81 \\
\quad Italian (it) & 70.68 & 71.78 & 66.68 & 69.08 & 56.51 & {-} & 59.45 \\
\quad Dutch (nl) & 69.74 & 70.06 & 65.73 & 69.04 & 55.54 & {-} & 51.94 \\
\quad Portuguese (pt) & 70.93 & 71.67 & 68.48 & 68.96 & 57.30 & {-} & 66.98 \\
\quad Romanian (ro) & 68.48 & 69.52 & 63.02 & 69.09 & 53.87 & {-} & 51.06 \\
\quad Russian (ru) & 67.74 & 69.20 & 65.47 & 67.10 & 52.93 & {-} & 44.80 \\
\quad Swedish (sv) & 67.82 & 68.71 & 62.92 & 68.44 & 54.11 & {-} & 49.14 \\
\quad Chinese (zh) & 65.40 & 70.13 & 66.96 & 66.13 & 52.17 & {-} & 53.82 \\
\quad \textbf{Average} & \underline{67.55} & \textbf{68.71} & 64.07 & 67.14 & 53.58 & {-} & 53.17 \\
\midrule
\multicolumn{8}{l}{\textbf{Multilingual GSM (MGSM)}} \\
\quad German (de) & 75.20 & 75.60 & 74.00 & {-} & 41.20 & 51.20 & 60.00 \\
\quad Spanish (es) & 83.20 & 82.00 & 75.20 & {-} & 52.80 & 71.20 & 73.60 \\
\quad French (fr) & 77.20 & 77.60 & 70.80 & {-} & 40.00 & 65.60 & 69.20 \\
\quad Japanese (ja) & 62.00 & 58.00 & 59.60 & {-} & 28.00 & 24.00 & 34.80 \\
\quad Russian (ru) & 76.40 & 81.20 & 74.40 & {-} & 44.40 & 38.80 & 49.20 \\
\quad Chinese (zh) & 73.20 & 32.80 & 72.40 & {-} & 42.80 & 62.40 & 67.20 \\
\quad \textbf{Average} & \textbf{74.53} & 67.87 & \underline{71.07} & {-} & 41.53 & 52.20 & 59.00 \\
\bottomrule
\end{tabular*}
\caption{Performance comparison of \textbf{Base} models on multilingual tasks (7B-12B scale). All scores are percentages. The best average score is in bold, and the second-best is underlined.}
\label{tab:multi-7b-base-updated}
\end{table}

\begin{table}[htbp]
\centering
\footnotesize
\begin{tabular*}{\textwidth}{l @{\extracolsep{\fill}} *{5}{S[table-format=2.2]}}
\toprule
\textbf{Task} & {\makecell{Falcon-H1- \\ 34B}} & {\makecell{Qwen2.5- \\ 72B}} & {\makecell{Gemma3- \\ 27B}} & {\makecell{Llama4- \\ scout}} & {\makecell{Llama3.1- \\ 70B}} \\
\midrule
\multicolumn{6}{l}{\textbf{Multilingual Hellaswag}} \\
\quad Arabic (ar) & 62.95 & 63.59 & 66.36 & 64.08 & 65.19 \\
\quad German (de) & 74.30 & 73.06 & 75.46 & 73.55 & 76.53 \\
\quad Spanish (es) & 79.68 & 78.75 & 79.80 & 77.96 & 80.86 \\
\quad French (fr) & 78.18 & 77.19 & 78.52 & 76.84 & 78.92 \\
\quad Hindi (hi) & 54.53 & 54.08 & 57.31 & 55.99 & 58.46 \\
\quad Italian (it) & 76.86 & 76.06 & 77.58 & 75.17 & 78.36 \\
\quad Dutch (nl) & 75.54 & 73.93 & 76.77 & 74.71 & 77.65 \\
\quad Portuguese (pt) & 78.26 & 77.68 & 78.87 & 76.77 & 80.37 \\
\quad Romanian (ro) & 72.47 & 66.87 & 74.04 & 72.17 & 74.38 \\
\quad Russian (ru) & 71.10 & 70.88 & 71.84 & 70.87 & 72.66 \\
\quad Swedish (sv) & 74.93 & 71.08 & 77.60 & 74.10 & 77.78 \\
\quad \textbf{Average} & 72.62 & 71.20 & \underline{74.01} & 72.02 & \textbf{74.65} \\
\midrule
\multicolumn{6}{l}{\textbf{Multilingual MMLU}} \\
\quad Arabic (ar) & 70.83 & 73.52 & 68.10 & 67.43 & 65.46 \\
\quad German (de) & 78.23 & 80.28 & 73.31 & 73.78 & 72.10 \\
\quad Spanish (es) & 79.65 & 80.63 & 74.56 & 74.33 & 73.33 \\
\quad French (fr) & 79.67 & 81.03 & 74.13 & 74.60 & 73.45 \\
\quad Hindi (hi) & 67.16 & 68.94 & 65.46 & 64.68 & 63.38 \\
\quad Italian (it) & 79.46 & 81.03 & 74.12 & 74.22 & 73.79 \\
\quad Dutch (nl) & 78.69 & 80.32 & 74.33 & 74.02 & 72.73 \\
\quad Portuguese (pt) & 79.64 & 80.88 & 74.47 & 73.95 & 73.78 \\
\quad Romanian (ro) & 78.42 & 79.03 & 73.97 & 73.60 & 72.52 \\
\quad Russian (ru) & 76.65 & 79.08 & 71.96 & 72.55 & 71.28 \\
\quad Swedish (sv) & 78.13 & 79.03 & 73.54 & 74.23 & 72.08 \\
\quad Chinese (zh) & 74.60 & 78.78 & 70.94 & 71.27 & 69.39 \\
\quad \textbf{Average} & \underline{76.76} & \textbf{78.54} & 72.41 & 72.38 & 71.10 \\
\midrule
\multicolumn{6}{l}{\textbf{Multilingual GSM (MGSM)}} \\
\quad German (de) & 82.40 & 83.20 & {-} & 76.00 & 71.20 \\
\quad Spanish (es) & 85.60 & 86.00 & {-} & 77.60 & 79.60 \\
\quad French (fr) & 80.00 & 77.60 & {-} & 72.40 & 69.20 \\
\quad Japanese (ja) & 76.80 & 80.80 & {-} & 70.80 & 61.20 \\
\quad Russian (ru) & 88.00 & 83.60 & {-} & 79.20 & 74.40 \\
\quad Chinese (zh) & 81.60 & 82.00 & {-} & 78.80 & 68.80 \\
\quad \textbf{Average} & \textbf{82.40} & \underline{82.20} & {-} & 75.80 & 70.73 \\
\bottomrule
\end{tabular*}
\caption{Performance comparison of \textbf{Base} models on multilingual tasks (34B scale). All scores are percentages. The best average score is in bold, and the second-best is underlined.}
\label{tab:multi-30b-base}
\end{table}

\newpage
\subsection{Multilingual Evaluations - Instruct Models}
\label{eval:tables_ml_ins}
\begin{table}[htbp]
\centering
\footnotesize
\begin{tabular*}{\textwidth}{l @{\extracolsep{\fill}} *{7}{S[table-format=2.2]}}
\toprule
\textbf{Task} & {\makecell{Falcon-H1- \\ 1.5B-deep}} & {\makecell{Falcon-H1- \\ 1.5B}} & {\makecell{Qwen3- \\ 1.7B}} & {\makecell{Qwen2.5- \\ 1.5B}} & {\makecell{Gemma3- \\ 1B}} & {\makecell{Llama3.2- \\ 1.2B}} & {\makecell{Falcon3- \\ 1.6B}} \\
\midrule
\multicolumn{8}{l}{\textbf{Multilingual Hellaswag}} \\
\quad Arabic (ar) & 44.22 & 41.51 & 36.19 & 39.20 & 37.65 & 33.38 & 29.49 \\
\quad German (de) & 53.83 & 49.97 & 37.95 & 43.04 & 42.09 & 40.65 & 31.31 \\
\quad Spanish (es) & 61.50 & 57.67 & 41.92 & 52.07 & 47.22 & 46.02 & 38.79 \\
\quad French (fr) & 60.54 & 56.51 & 41.45 & 50.50 & 46.29 & 44.04 & 38.52 \\
\quad Hindi (hi) & 37.35 & 34.98 & 30.68 & 30.24 & 33.42 & 32.89 & 28.50 \\
\quad Italian (it) & 57.96 & 53.54 & 40.05 & 46.32 & 44.62 & 43.09 & 32.76 \\
\quad Dutch (nl) & 52.94 & 49.98 & 36.89 & 41.76 & 40.11 & 39.19 & 29.67 \\
\quad Portuguese (pt) & 60.01 & 55.80 & 41.64 & 50.89 & 45.66 & 44.22 & 35.22 \\
\quad Romanian (ro) & 50.44 & 45.81 & 36.01 & 35.70 & 39.81 & 37.10 & 30.07 \\
\quad Russian (ru) & 54.16 & 50.66 & 38.22 & 45.85 & 42.27 & 38.36 & 29.22 \\
\quad Swedish (sv) & 51.58 & 46.80 & 35.80 & 36.67 & 40.39 & 38.60 & 28.88 \\
\quad \textbf{Average} & \textbf{53.14} & \underline{49.38} & 37.89 & 42.93 & 41.77 & 39.78 & 32.04 \\
\midrule
\multicolumn{8}{l}{\textbf{Multilingual MMLU}} \\
\quad Arabic (ar) & 45.13 & 41.21 & 35.53 & 41.51 & 32.52 & 30.00 & 27.48 \\
\quad German (de) & 55.02 & 50.57 & 43.92 & 48.04 & 36.11 & 36.75 & 33.91 \\
\quad Spanish (es) & 57.85 & 53.09 & 45.40 & 51.42 & 36.25 & 38.29 & 37.49 \\
\quad French (fr) & 57.47 & 51.91 & 44.90 & 51.08 & 36.06 & 37.66 & 36.87 \\
\quad Hindi (hi) & 40.74 & 38.05 & 28.98 & 33.27 & 31.66 & 32.39 & 27.92 \\
\quad Italian (it) & 55.94 & 50.56 & 45.63 & 49.04 & 35.54 & 37.01 & 33.71 \\
\quad Dutch (nl) & 54.54 & 48.97 & 41.63 & 46.72 & 35.90 & 36.60 & 31.00 \\
\quad Portuguese (pt) & 57.55 & 52.09 & 30.76 & 51.82 & 34.94 & 36.04 & 35.80 \\
\quad Romanian (ro) & 53.83 & 48.32 & 40.63 & 43.32 & 34.92 & 32.81 & 30.97 \\
\quad Russian (ru) & 51.93 & 46.84 & 40.59 & 45.89 & 35.37 & 35.27 & 29.12 \\
\quad Swedish (sv) & 53.02 & 47.10 & 37.62 & 42.82 & 34.76 & 34.87 & 30.46 \\
\quad Chinese (zh) & 51.53 & 47.16 & 50.38 & 52.95 & 33.66 & 37.68 & 33.60 \\
\quad \textbf{Average} & \textbf{53.00} & \underline{48.06} & 39.60 & 45.90 & 34.91 & 35.24 & 32.25 \\
\midrule
\multicolumn{8}{l}{\textbf{Multilingual GSM (MGSM)}} \\
\quad German (de) & 61.20 & 61.20 & 50.80 & 42.00 & {-} & 32.40 & 15.20 \\
\quad Spanish (es) & 70.00 & 71.20 & 60.80 & 49.20 & {-} & 33.60 & 24.40 \\
\quad French (fr) & 50.40 & 57.60 & 45.20 & 46.80 & {-} & 34.00 & 24.00 \\
\quad Japanese (ja) & 54.40 & 41.20 & 44.80 & 34.40 & {-} & 19.60 & 3.20 \\
\quad Russian (ru) & 57.60 & 55.60 & 56.00 & 43.60 & {-} & 29.20 & 5.20 \\
\quad Chinese (zh) & 66.40 & 61.20 & 56.80 & 55.20 & {-} & 29.60 & 20.00 \\
\quad \textbf{Average} & \textbf{60.00} & \underline{58.00} & 52.40 & 45.20 & {-} & 29.73 & 15.33 \\
\bottomrule
\end{tabular*}
\caption{Performance comparison of \textbf{Instruct} models on multilingual tasks (1B-2B scale). All scores are percentages. The best average score is in bold, and the second-best is underlined.}
\label{tab:multi-1b-ins-updated}
\end{table}

\begin{table}[htbp]
\centering
\footnotesize
\begin{tabular*}{\textwidth}{l @{\extracolsep{\fill}} *{6}{S[table-format=2.2]}}
\toprule
\textbf{Task} & {\makecell{Falcon-H1- \\ 3B}} & {\makecell{Qwen3- \\ 4B}} & {\makecell{Qwen2.5- \\ 3B}} & {\makecell{Gemma3- \\ 4B}} & {\makecell{Llama3.2- \\ 3B}} & {\makecell{Falcon3- \\ 3B}} \\
\midrule
\multicolumn{7}{l}{\textbf{Multilingual Hellaswag}} \\
\quad Arabic (ar) & 49.37 & 40.48 & 46.57 & 48.69 & 40.52 & 29.83 \\
\quad German (de) & 59.27 & 44.51 & 51.41 & 54.22 & 52.90 & 36.47 \\
\quad Spanish (es) & 66.48 & 46.99 & 60.17 & 60.26 & 58.80 & 51.19 \\
\quad French (fr) & 65.88 & 46.82 & 60.09 & 59.55 & 56.92 & 50.98 \\
\quad Hindi (hi) & 42.32 & 33.66 & 33.33 & 43.19 & 40.32 & 28.47 \\
\quad Italian (it) & 62.44 & 45.45 & 55.11 & 58.01 & 55.04 & 38.47 \\
\quad Dutch (nl) & 58.83 & 43.19 & 51.36 & 53.96 & 51.96 & 32.35 \\
\quad Portuguese (pt) & 64.68 & 46.70 & 59.53 & 60.11 & 56.39 & 50.44 \\
\quad Romanian (ro) & 55.55 & 40.99 & 42.35 & 53.95 & 47.12 & 32.15 \\
\quad Russian (ru) & 59.04 & 42.83 & 53.42 & 54.26 & 49.12 & 31.22 \\
\quad Swedish (sv) & 57.83 & 42.63 & 45.54 & 53.08 & 51.17 & 31.01 \\
\quad \textbf{Average} & \textbf{58.34} & 43.12 & 50.81 & \underline{54.48} & 50.93 & 37.51 \\
\midrule
\multicolumn{7}{l}{\textbf{Multilingual MMLU}} \\
\quad Arabic (ar) & 47.57 & 41.65 & 48.21 & 46.24 & 41.03 & 30.39 \\
\quad German (de) & 56.87 & 58.70 & 55.35 & 52.27 & 50.76 & 40.85 \\
\quad Spanish (es) & 59.88 & 56.14 & 57.83 & 53.69 & 52.76 & 48.45 \\
\quad French (fr) & 58.99 & 59.38 & 57.61 & 53.07 & 51.52 & 47.32 \\
\quad Hindi (hi) & 42.42 & 36.39 & 38.45 & 44.27 & 40.09 & 30.17 \\
\quad Italian (it) & 57.89 & 57.34 & 56.35 & 53.49 & 51.24 & 41.55 \\
\quad Dutch (nl) & 55.99 & 56.59 & 54.34 & 52.30 & 50.31 & 37.47 \\
\quad Portuguese (pt) & 59.29 & 26.74 & 58.13 & 52.54 & 52.25 & 47.96 \\
\quad Romanian (ro) & 55.12 & 56.48 & 50.81 & 51.75 & 47.94 & 36.81 \\
\quad Russian (ru) & 54.14 & 54.39 & 54.39 & 50.48 & 46.22 & 33.32 \\
\quad Swedish (sv) & 55.30 & 53.83 & 50.23 & 52.03 & 48.60 & 33.88 \\
\quad Chinese (zh) & 53.83 & 55.88 & 58.34 & 50.13 & 47.77 & 39.80 \\
\quad \textbf{Average} & \textbf{54.90} & 50.70 & \underline{52.90} & 51.10 & 48.40 & 38.90 \\
\midrule
\multicolumn{7}{l}{\textbf{Multilingual GSM (MGSM)}} \\
\quad German (de) & 63.60 & 68.80 & 58.00 & {-} & 63.60 & 41.60 \\
\quad Spanish (es) & 71.60 & 73.20 & 64.80 & {-} & 71.60 & 63.20 \\
\quad French (fr) & 66.40 & 67.20 & 54.40 & {-} & 62.00 & 60.80 \\
\quad Japanese (ja) & 59.60 & 65.20 & 51.20 & {-} & 51.60 & 14.00 \\
\quad Russian (ru) & 54.00 & 65.60 & 54.40 & {-} & 62.80 & 20.40 \\
\quad Chinese (zh) & 68.40 & 73.20 & 62.40 & {-} & 61.60 & 52.40 \\
\quad \textbf{Average} & \underline{63.90} & \textbf{68.90} & 57.30 & {-} & 62.20 & 42.10 \\
\bottomrule
\end{tabular*}
\caption{Performance comparison of \textbf{Instruct} models on multilingual tasks (3B-4B scale). All scores are percentages. The best average score is in bold, and the second-best is underlined.}
\label{tab:multi-3b-ins-styled}
\end{table}

\begin{table}[htbp]
\centering
\footnotesize
\begin{tabular*}{\textwidth}{l @{\extracolsep{\fill}} *{7}{S[table-format=2.2]}}
\toprule
\textbf{Task} & {\makecell{Falcon-H1- \\ 7B}} & {\makecell{Qwen3- \\ 8B}} & {\makecell{Qwen2.5- \\ 7B}} & {\makecell{Gemma3- \\ 12B}} & {\makecell{Llama3.1- \\ 8B}} & {\makecell{Falcon3- \\ 7B}} & {\makecell{Falcon3- \\ 10B}} \\
\midrule
\multicolumn{8}{l}{\textbf{Multilingual Hellaswag}} \\
\quad Arabic (ar) & 56.73 & 44.19 & 52.23 & 59.19 & 49.44 & 32.19 & 35.51 \\
\quad German (de) & 69.51 & 48.34 & 60.57 & 67.64 & 62.29 & 46.58 & 52.95 \\
\quad Spanish (es) & 75.94 & 51.06 & 68.81 & 72.53 & 68.58 & 68.74 & 74.00 \\
\quad French (fr) & 75.08 & 52.35 & 67.89 & 70.86 & 66.79 & 68.14 & 72.71 \\
\quad Hindi (hi) & 49.26 & 37.34 & 38.72 & 51.93 & 45.66 & 29.92 & 31.90 \\
\quad Italian (it) & 72.32 & 49.67 & 63.57 & 69.86 & 64.55 & 52.72 & 59.14 \\
\quad Dutch (nl) & 70.52 & 47.61 & 60.75 & 67.64 & 63.18 & 41.54 & 47.52 \\
\quad Portuguese (pt) & 74.38 & 51.15 & 67.93 & 71.28 & 66.47 & 66.85 & 72.18 \\
\quad Romanian (ro) & 66.25 & 44.87 & 49.32 & 66.45 & 59.21 & 39.65 & 44.17 \\
\quad Russian (ru) & 66.26 & 47.25 & 61.60 & 65.17 & 59.17 & 39.60 & 44.80 \\
\quad Swedish (sv) & 68.97 & 46.45 & 54.64 & 69.21 & 62.73 & 39.98 & 45.54 \\
\quad \textbf{Average} & \textbf{67.75} & 47.30 & 58.74 & \underline{66.53} & 60.74 & 47.81 & 52.77 \\
\midrule
\multicolumn{8}{l}{\textbf{Multilingual MMLU}} \\
\quad Arabic (ar) & 60.44 & 49.11 & 56.23 & 59.40 & 47.17 & 35.34 & 36.26 \\
\quad German (de) & 69.13 & 58.63 & 63.65 & 66.48 & 57.62 & 52.44 & 55.68 \\
\quad Spanish (es) & 71.96 & 59.55 & 66.17 & 67.29 & 59.62 & 64.14 & 67.71 \\
\quad French (fr) & 71.04 & 53.14 & 65.30 & 67.35 & 59.51 & 63.78 & 67.45 \\
\quad Hindi (hi) & 56.52 & 36.61 & 46.15 & 57.41 & 45.31 & 34.33 & 35.24 \\
\quad Italian (it) & 70.23 & 58.99 & 65.19 & 67.35 & 58.27 & 56.02 & 60.26 \\
\quad Dutch (nl) & 69.83 & 57.66 & 63.05 & 67.02 & 57.01 & 49.28 & 52.79 \\
\quad Portuguese (pt) & 71.55 & 23.60 & 64.44 & 67.67 & 58.80 & 63.80 & 67.27 \\
\quad Romanian (ro) & 68.81 & 58.32 & 59.27 & 66.55 & 55.78 & 48.33 & 51.85 \\
\quad Russian (ru) & 67.96 & 55.51 & 63.29 & 64.58 & 55.21 & 42.48 & 45.85 \\
\quad Swedish (sv) & 68.65 & 43.71 & 60.43 & 66.31 & 56.57 & 46.79 & 50.02 \\
\quad Chinese (zh) & 65.82 & 65.66 & 65.89 & 63.94 & 55.09 & 50.12 & 53.64 \\
\quad \textbf{Average} & \textbf{67.83} & 50.44 & 61.20 & \underline{65.22} & 55.53 & 50.62 & 53.67 \\
\midrule
\multicolumn{8}{l}{\textbf{Multilingual GSM (MGSM)}} \\
\quad German (de) & 73.20 & 72.00 & 64.80 & {-} & 71.20 & 60.00 & 68.40 \\
\quad Spanish (es) & 82.00 & 68.40 & 69.20 & {-} & 79.60 & 80.40 & 84.80 \\
\quad French (fr) & 72.40 & 60.00 & 61.60 & {-} & 69.60 & 73.20 & 76.40 \\
\quad Japanese (ja) & 70.80 & 61.60 & 64.80 & {-} & 56.00 & 30.80 & 45.20 \\
\quad Russian (ru) & 67.60 & 60.80 & 61.60 & {-} & 76.40 & 36.00 & 46.80 \\
\quad Chinese (zh) & 74.80 & 68.40 & 74.80 & {-} & 71.60 & 57.60 & 67.20 \\
\quad \textbf{Average} & \textbf{73.50} & 65.20 & 66.10 & {-} & \underline{70.70} & 56.30 & 64.80 \\
\bottomrule
\end{tabular*}
\caption{Performance comparison of \textbf{Instruct} models on multilingual tasks (7B-12B scale). All scores are percentages. The best average score is in bold, and the second-best is underlined.}
\label{tab:multi-7b-ins-styled}
\end{table}

\begin{table}[htbp]
\centering
\footnotesize
\begin{tabular*}{\textwidth}{l @{\extracolsep{\fill}} *{7}{S[table-format=2.2]}}
\toprule
\textbf{Task} & {\makecell{Falcon-H1- \\ 34B}} & {\makecell{Qwen3- \\ 32B}} & {\makecell{Qwen2.5- \\ 72B}} & {\makecell{Qwen2.5- \\ 32B}} & {\makecell{Gemma3- \\ 27B}} & {\makecell{Llama4-Scout}} & {\makecell{Llama3.3- \\ 70B}} \\
\midrule
\multicolumn{8}{l}{\textbf{Multilingual Hellaswag}} \\
\quad Arabic (ar) & 64.33 & 52.39 & 63.20 & 60.75 & 62.40 & 55.39 & 58.44 \\
\quad German (de) & 76.96 & 59.48 & 70.93 & 67.63 & 70.52 & 62.57 & 64.68 \\
\quad Spanish (es) & 80.39 & 62.87 & 76.03 & 72.68 & 74.94 & 69.10 & 69.59 \\
\quad French (fr) & 80.56 & 62.45 & 75.19 & 72.60 & 73.97 & 68.13 & 68.62 \\
\quad Hindi (hi) & 57.01 & 49.25 & 53.40 & 48.50 & 55.37 & 45.25 & 50.47 \\
\quad Italian (it) & 79.07 & 61.38 & 73.61 & 70.21 & 73.64 & 65.44 & 66.97 \\
\quad Dutch (nl) & 77.08 & 59.56 & 70.84 & 67.75 & 70.72 & 65.53 & 68.05 \\
\quad Portuguese (pt) & 80.25 & 62.03 & 75.71 & 72.73 & 74.65 & 67.87 & 69.99 \\
\quad Romanian (ro) & 73.95 & 57.83 & 66.06 & 60.83 & 69.93 & 62.80 & 63.34 \\
\quad Russian (ru) & 73.92 & 56.58 & 69.53 & 66.75 & 68.31 & 59.93 & 64.55 \\
\quad Swedish (sv) & 76.47 & 58.50 & 69.83 & 64.46 & 71.61 & 63.30 & 66.39 \\
\quad \textbf{Average} & \textbf{74.55} & 58.39 & 69.48 & 65.90 & \underline{69.64} & 62.30 & 64.64 \\
\midrule
\multicolumn{8}{l}{\textbf{Multilingual MMLU}} \\
\quad Arabic (ar) & 72.08 & 65.71 & 73.67 & 68.17 & 66.30 & 68.70 & 70.51 \\
\quad German (de) & 78.97 & 71.32 & 78.57 & 75.71 & 71.96 & 75.61 & 77.89 \\
\quad Spanish (es) & 80.25 & 74.04 & 81.03 & 76.86 & 73.65 & 76.86 & 79.47 \\
\quad French (fr) & 80.38 & 72.49 & 81.05 & 76.88 & 73.86 & 76.50 & 79.08 \\
\quad Hindi (hi) & 68.34 & 56.77 & 69.07 & 61.39 & 63.95 & 67.37 & 68.22 \\
\quad Italian (it) & 79.97 & 74.71 & 81.04 & 76.88 & 73.70 & 76.32 & 78.82 \\
\quad Dutch (nl) & 79.56 & 67.34 & 79.74 & 75.58 & 72.54 & 76.00 & 78.70 \\
\quad Portuguese (pt) & 80.49 & 31.56 & 81.25 & 76.21 & 74.24 & 77.06 & 78.99 \\
\quad Romanian (ro) & 79.15 & 72.12 & 78.76 & 73.60 & 73.07 & 75.54 & 77.73 \\
\quad Russian (ru) & 77.31 & 73.06 & 78.20 & 74.41 & 71.52 & 74.31 & 75.88 \\
\quad Swedish (sv) & 78.85 & 69.09 & 78.51 & 73.54 & 72.82 & 76.11 & 78.10 \\
\quad Chinese (zh) & 75.41 & 73.61 & 78.22 & 74.94 & 69.54 & 73.50 & 74.97 \\
\quad \textbf{Average} & \underline{77.76} & 66.20 & \textbf{78.26} & 73.56 & 71.60 & 74.58 & 76.67 \\
\midrule
\multicolumn{8}{l}{\textbf{Multilingual GSM (MGSM)}} \\
\quad German (de) & 75.20 & 73.20 & 72.80 & 72.80 & 78.00 & 86.00 & 88.00 \\
\quad Spanish (es) & 80.80 & 76.00 & 79.20 & 76.80 & 84.80 & 88.40 & 89.60 \\
\quad French (fr) & 70.80 & 67.60 & 60.80 & 61.20 & 74.80 & 82.80 & 82.80 \\
\quad Japanese (ja) & 78.40 & 69.20 & 74.80 & 81.20 & 76.80 & 78.00 & 84.40 \\
\quad Russian (ru) & 68.00 & 68.00 & 63.20 & 68.40 & 72.40 & 84.40 & 88.40 \\
\quad Chinese (zh) & 84.80 & 76.80 & 83.20 & 81.20 & 80.40 & 83.60 & 88.00 \\
\quad \textbf{Average} & 76.33 & 71.80 & 72.33 & 73.60 & \underline{77.87} & 83.87 & \textbf{86.87} \\
\bottomrule
\end{tabular*}
\caption{Performance comparison of \textbf{Instruct} models on multilingual tasks (34B scale). All scores are percentages. The best average score is in bold, and the second-best is underlined.}
\label{tab:multi-34b-ins-styled}
\end{table}

\newpage
\subsection{Long-context Evaluations - Instruct Models}
\label{eval:tables_long_context}
\begin{table}[htbp]
\centering
\small
\scalebox{0.8}{
\begin{tabularx}{\textwidth}{|l|l|X|X|X|X|}
\hline
\textbf{Dataset} & \textbf{Metric} & \textbf{Falcon-H1-34B-Instruct} & \textbf{Qwen3-32B} & \textbf{Qwen2.5-72B-Instruct} & \textbf{Llama-3.3-70B-Instruct} \\
\hline
alce\_asqa & citation\_prec & 61.49 & 70.05 & 65.04 & 49.47 \\
alce\_asqa & citation\_rec & 61.90 & 63.37 & 69.52 & 66.00 \\
alce\_asqa & str\_em & 49.68 & 44.10 & 46.98 & 51.10 \\
alce\_qampari & citation\_prec & 23.67 & 29.01 & 28.36 & 20.48 \\
alce\_qampari & citation\_rec & 23.42 & 25.60 & 27.81 & 18.81 \\
alce\_qampari & qampari\_rec\_top5 & 28.80 & 29.20 & 29.40 & 37.00 \\
banking77 & exact\_match & 82.00 & 87.00 & 86.00 & 84.00 \\
clinic150 & exact\_match & 91.00 & 91.00 & 91.00 & 85.00 \\
hotpotqa & substring\_exact\_match & 66.67 & 69.33 & 67.67 & 71.33 \\
infbench\_choice & exact\_match & 55.00 & 46.00 & 61.00 & 50.00 \\
infbench\_qa & rougeL\_f1 & 20.75 & 24.01 & 20.39 & 22.14 \\
infbench\_sum & f1 & 25.02 & 33.12 & 31.65 & 30.25 \\
json\_kv & substring\_exact\_match & 100 & 100 & 100 & 100 \\
msmarco\_rerank\_psg & NDCG@10 & 76.94 & 81.92 & 85.40 & 84.32 \\
multi\_lexsum & f1 & 32.06 & 34.17 & 34.07 & 31.84 \\
narrativeqa & rougeL\_f1 & 22.85 & 24.89 & 24.21 & 28.86 \\
nlu & exact\_match & 74.00 & 79.00 & 83.00 & 83.00 \\
nq & substring\_exact\_match & 63.33 & 57.83 & 66.33 & 61.17 \\
popqa & substring\_exact\_match & 65.33 & 61.33 & 61.67 & 69.00 \\
ruler\_niah\_mk\_2 & ruler\_recall & 100 & 100 & 100 & 100 \\
ruler\_niah\_mk\_3 & ruler\_recall & 100 & 100 & 100 & 100 \\
ruler\_niah\_mv & ruler\_recall & 100 & 100 & 100 & 100 \\
trec\_coarse & exact\_match & 69.00 & 69.00 & 72.00 & 63.00 \\
trec\_fine & exact\_match & 34.00 & 42.00 & 46.00 & 34.00 \\
triviaqa & substring\_exact\_match & 93.33 & 88.50 & 93.17 & 95.67 \\
\hline
\end{tabularx}
}
\caption{Performance of HELMET tasks at sequence length 8192.}
\label{tab:helmet-8k}
\end{table}

\begin{table}[htbp]
\centering
\small
\scalebox{0.8}{
\begin{tabularx}{\textwidth}{|l|l|X|X|X|X|}
\hline
\textbf{Dataset} & \textbf{Metric} & \textbf{Falcon-H1-34B-Instruct} & \textbf{Qwen3-32B} & \textbf{Qwen2.5-72B-Instruct} & \textbf{Llama-3.3-70B-Instruct} \\
\hline
alce\_asqa & citation\_prec & 49.85 & 71.23 & 55.52 & 48.14 \\
alce\_asqa & citation\_rec & 47.75 & 65.46 & 62.08 & 62.54 \\
alce\_asqa & str\_em & 52.15 & 46.20 & 43.77 & 50.17 \\
alce\_qampari & citation\_prec & 21.90 & 28.08 & 28.85 & 17.34 \\
alce\_qampari & citation\_rec & 20.91 & 25.96 & 27.88 & 15.64 \\
alce\_qampari & qampari\_rec\_top5 & 29.40 & 31.40 & 30.60 & 42.40 \\
banking77 & exact\_match & 86.00 & 90.00 & 91.00 & 91.00 \\
clinic150 & exact\_match & 95.00 & 96.00 & 94.00 & 95.00 \\
hotpotqa & substring\_exact\_match & 65.33 & 64.33 & 65.67 & 69.33 \\
infbench\_choice & exact\_match & 55.00 & 49.00 & 56.00 & 55.00 \\
infbench\_qa & rougeL\_f1 & 28.40 & 31.48 & 32.13 & 32.46 \\
infbench\_sum & f1 & 24.92 & 34.91 & 34.08 & 32.11 \\
json\_kv & substring\_exact\_match & 100 & 100 & 100 & 100 \\
msmarco\_rerank\_psg & NDCG@10 & 66.15 & 70.24 & 74.89 & 76.01 \\
multi\_lexsum & f1 & 33.68 & 34.61 & 34.51 & 32.70 \\
narrativeqa & rougeL\_f1 & 20.52 & 26.57 & 29.27 & 31.78 \\
nlu & exact\_match & 77.00 & 76.00 & 86.00 & 85.00 \\
nq & substring\_exact\_match & 57.00 & 56.50 & 67.33 & 61.50 \\
popqa & substring\_exact\_match & 67.50 & 60.83 & 62.33 & 64.33 \\
ruler\_niah\_mk\_2 & ruler\_recall & 100 & 100 & 100 & 100 \\
ruler\_niah\_mk\_3 & ruler\_recall & 100 & 100 & 100 & 100 \\
ruler\_niah\_mv & ruler\_recall & 100 & 100 & 100 & 100 \\
trec\_coarse & exact\_match & 66.00 & 78.00 & 85.00 & 70.00 \\
trec\_fine & exact\_match & 43.00 & 51.00 & 51.00 & 39.00 \\
triviaqa & substring\_exact\_match & 93.00 & 86.50 & 93.67 & 95.67 \\
\hline
\end{tabularx}
}
\caption{Performance of HELMET tasks at sequence length 16384.}
\label{tab:helmet-16k}
\end{table}

\begin{table}[htbp]
\centering
\small
\scalebox{0.8}{
\begin{tabularx}{\textwidth}{|l|l|X|X|X|X|}
\hline
\textbf{Dataset} & \textbf{Metric} & \textbf{Falcon-H1-34B-Instruct} & \textbf{Qwen3-32B} & \textbf{Qwen2.5-72B-Instruct} & \textbf{Llama-3.3-70B-Instruct} \\
\hline
alce\_asqa & citation\_prec & 34.98 & 61.34 & 53.16 & 45.29 \\
alce\_asqa & citation\_rec & 25.52 & 56.04 & 57.80 & 55.78 \\
alce\_asqa & str\_em & 45.18 & 43.72 & 42.07 & 46.37 \\
alce\_qampari & citation\_prec & 12.13 & 25.43 & 24.18 & 11.12 \\
alce\_qampari & citation\_rec & 11.31 & 24.99 & 22.54 & 10.69 \\
alce\_qampari & qampari\_rec\_top5 & 16.80 & 31.40 & 27.60 & 45.00 \\
banking77 & exact\_match & 87.00 & 90.00 & 90.00 & 92.00 \\
clinic150 & exact\_match & 97.00 & 97.00 & 98.00 & 96.00 \\
hotpotqa & substring\_exact\_match & 60.67 & 57.67 & 64.33 & 66.00 \\
infbench\_choice & exact\_match & 56.00 & 62.00 & 57.00 & 68.00 \\
infbench\_qa & rougeL\_f1 & 29.26 & 36.47 & 32.36 & 41.08 \\
infbench\_sum & f1 & 22.87 & 35.47 & 35.21 & 31.58 \\
json\_kv & substring\_exact\_match & 99 & 100 & 98 & 100 \\
msmarco\_rerank\_psg & NDCG@10 & 50.35 & 54.98 & 60.66 & 65.71 \\
multi\_lexsum & f1 & 34.62 & 32.59 & 36.73 & 33.27 \\
narrativeqa & rougeL\_f1 & 19.99 & 24.97 & 28.29 & 33.50 \\
nlu & exact\_match & 84.00 & 80.00 & 82.00 & 87.00 \\
nq & substring\_exact\_match & 60.83 & 57.00 & 63.00 & 59.17 \\
popqa & substring\_exact\_match & 58.00 & 57.50 & 59.33 & 59.50 \\
ruler\_niah\_mk\_2 & ruler\_recall & 97.00 & 100 & 100 & 100 \\
ruler\_niah\_mk\_3 & ruler\_recall & 95.00 & 100 & 96.00 & 99.00 \\
ruler\_niah\_mv & ruler\_recall & 99.00 & 100 & 99.50 & 99.50 \\
trec\_coarse & exact\_match & 64.00 & 64.00 & 90.00 & 70.00 \\
trec\_fine & exact\_match & 44.00 & 50.00 & 47.00 & 46.00 \\
triviaqa & substring\_exact\_match & 92.33 & 87.17 & 93.67 & 96.17 \\
\hline
\end{tabularx}
}
\caption{Performance of HELMET tasks at sequence length 32768.}
\label{tab:helmet-32k}
\end{table}

\begin{table}[htbp]
\centering
\small
\scalebox{0.8}{
\begin{tabularx}{\textwidth}{|l|l|X|X|X|X|}
\hline
\textbf{Dataset} & \textbf{Metric} & \textbf{Falcon-H1-34B-Instruct} & \textbf{Qwen3-32B} & \textbf{Qwen2.5-72B-Instruct} & \textbf{Llama-3.3-70B-Instruct} \\
\hline
alce\_asqa & citation\_prec & 10.91 & 47.02 & 31.50 & 40.17 \\
alce\_asqa & citation\_rec & 4.09 & 45.37 & 34.93 & 43.70 \\
alce\_asqa & str\_em & 18.98 & 45.68 & 44.18 & 41.12 \\
alce\_qampari & citation\_prec & 2.82 & 16.15 & 7.76 & 8.57 \\
alce\_qampari & citation\_rec & 1.75 & 14.03 & 6.09 & 8.07 \\
alce\_qampari & qampari\_rec\_top5 & 0.40 & 29.20 & 19.60 & 37.60 \\
banking77 & exact\_match & 92.00 & 92.00 & 93.00 & 96.00 \\
clinic150 & exact\_match & 96.00 & 98.00 & 97.00 & 97.00 \\
hotpotqa & substring\_exact\_match & 63.00 & 55.33 & 54.67 & 61.00 \\
infbench\_choice & exact\_match & 56.00 & 65.00 & 62.00 & 74.00 \\
infbench\_qa & rougeL\_f1 & 23.32 & 42.56 & 18.15 & 38.74 \\
infbench\_sum & f1 & 21.59 & 38.51 & 33.99 & 30.11 \\
json\_kv & substring\_exact\_match & 87 & 99 & 54 & 100 \\
msmarco\_rerank\_psg & NDCG@10 & 30.60 & 37.83 & 40.71 & 41.04 \\
multi\_lexsum & f1 & 35.48 & 35.20 & 33.21 & 33.58 \\
narrativeqa & rougeL\_f1 & 18.03 & 34.85 & 29.98 & 32.97 \\
nlu & exact\_match & 83.00 & 83.00 & 82.00 & 86.00 \\
nq & substring\_exact\_match & 57.67 & 55.00 & 56.50 & 60.83 \\
popqa & substring\_exact\_match & 51.67 & 52.50 & 52.50 & 59.00 \\
ruler\_niah\_mk\_2 & ruler\_recall & 76.00 & 94.00 & 80.00 & 99.00 \\
ruler\_niah\_mk\_3 & ruler\_recall & 66.00 & 97.00 & 54.00 & 97.00 \\
ruler\_niah\_mv & ruler\_recall & 93.75 & 96.00 & 99.00 & 99.25 \\
trec\_coarse & exact\_match & 80.00 & 84.00 & 90.00 & 79.00 \\
trec\_fine & exact\_match & 60.00 & 64.00 & 55.00 & 52.00 \\
triviaqa & substring\_exact\_match & 96.00 & 85.00 & 89.33 & 95.50 \\
\hline
\end{tabularx}
}
\caption{Performance of HELMET tasks at sequence length 65536.}
\label{tab:helmet-65k}
\end{table}

\begin{table}[htbp]
\centering
\small
\scalebox{0.8}{
\begin{tabularx}{\textwidth}{|l|l|X|X|X|X|}
\hline
\textbf{Dataset} & \textbf{Metric} & \textbf{Falcon-H1-34B-Instruct} & \textbf{Qwen3-32B} & \textbf{Qwen2.5-72B-Instruct} & \textbf{Llama-3.3-70B-Instruct} \\
\hline
alce\_asqa & citation\_prec & 3.00 & 41.31 & 3.07 & 6.40 \\
alce\_asqa & citation\_rec & 0.53 & 37.40 & 2.76 & 4.80 \\
alce\_asqa & str\_em & 9.12 & 43.70 & 35.58 & 37.38 \\
alce\_qampari & citation\_prec & 2.77 & 11.31 & 0.00 & 1.53 \\
alce\_qampari & citation\_rec & 1.44 & 9.43 & 0.00 & 1.13 \\
alce\_qampari & qampari\_rec\_top5 & 0.40 & 21.00 & 4.40 & 18.20 \\
banking77 & exact\_match & 93.00 & 97.00 & 90.00 & 87.00 \\
clinic150 & exact\_match & 96.00 & 93.00 & 96.00 & 92.00 \\
hotpotqa & substring\_exact\_match & 56.00 & 47.67 & 30.67 & 43.67 \\
infbench\_choice & exact\_match & 62.00 & 75.00 & 61.00 & 58.00 \\
infbench\_qa & rougeL\_f1 & 24.82 & 46.94 & 13.16 & 45.26 \\
infbench\_sum & f1 & 19.44 & 38.90 & 30.67 & 32.79 \\
json\_kv & substring\_exact\_match & 49 & 82 & 15 & 87 \\
msmarco\_rerank\_psg & NDCG@10 & 19.63 & 29.75 & 24.86 & 25.47 \\
multi\_lexsum & f1 & 34.18 & 34.52 & 32.48 & 32.59 \\
narrativeqa & rougeL\_f1 & 14.60 & 38.61 & 24.67 & 34.93 \\
nlu & exact\_match & 87.00 & 89.00 & 85.00 & 82.00 \\
nq & substring\_exact\_match & 53.83 & 49.00 & 38.33 & 47.33 \\
popqa & substring\_exact\_match & 48.17 & 48.17 & 34.67 & 47.50 \\
ruler\_niah\_mk\_2 & ruler\_recall & 43.00 & 84.00 & 43.00 & 66.00 \\
ruler\_niah\_mk\_3 & ruler\_recall & 40.00 & 84.00 & 15.00 & 79.00 \\
ruler\_niah\_mv & ruler\_recall & 94.50 & 94.50 & 82.25 & 96.75 \\
trec\_coarse & exact\_match & 66.00 & 85.00 & 92.00 & 66.00 \\
trec\_fine & exact\_match & 46.00 & 67.00 & 48.00 & 63.00 \\
triviaqa & substring\_exact\_match & 90.83 & 83.50 & 65.67 & 83.00 \\
\hline
\end{tabularx}
}
\caption{Performance of HELMET tasks at sequence length 131072.}
\label{tab:helmet-131k}
\end{table}

\newpage
\section{Training Data}

\subsection{Synthetic English Data Topics}
\begin{table}[h!]
\centering
\begin{tabularx}{\textwidth}{ *{3}{>{\raggedright\arraybackslash}X} }
\toprule
Acoustics & Fields of finance & Patent law \\
Administrative law & Fields of history & Photonics \\
Algebra & Fields of mathematics & Physical chemistry \\
Analytical chemistry & Gases & Privacy law \\
Animal law & Genetics & Private law \\
Astrophysics & Geometry & Probability \\
Branches of biology & Health care & Property law \\
Branches of genetics & Health law & Public law \\
Branches of philosophy & Health research & Quantum mechanics \\
Branches of psychology & Health sciences & Religious law \\
Business law & Historical eras & Religious legal systems \\
Calculus & History by continent and topic & Sex laws \\
Chinese law & History by topic & Social law \\
Civil law (common law) & History of medical and surgical specialties & Specialist law enforcement agencies \\
Civil law legal systems & History of the United States & Statistics \\
Civil procedure & Housing law & Statutory law by topic \\
Civil rights case law & Human anatomy by organ & Subfields of chemistry \\
Combinatorics & Human anatomy by system & Subfields of computer science \\
Common law legal systems & Human physiology & Subfields of economics \\
Condensed matter physics & Immigration law & Subfields of physics \\
Contract law & Innovation economics & Tax law \\
Criminal law & Inorganic chemistry & Theory of relativity \\
Cultural history by period & Intellectual property law & Tort law \\
Decrees & International law & Types of accounting \\
Delegated legislation & Labour law & Types of marketing \\
Determinants of health & Legal terminology by type of law & Workplace \\
Edicts & Linear algebra & \\
Electromagnetism & Management theory & \\
Emergency laws & Marketing techniques & \\
Engineering disciplines & Medical diagnosis & \\
Entrepreneurship & Medical procedures & \\
Environmental law & Medical specialties & \\
Family law & Medical treatments & \\
\bottomrule
\end{tabularx}
\caption{List of topics crawled from Wikipedia category tree starting nodes.}
\label{tab:wikipedia-categories-synthetic}
\end{table}

\subsection{Programming Languages}
\label{appendix:programming_langs}
The 67 included programming languages are: Agda, Assembly, Batchfile, BibTex, C, C\#, C++, CMake, COBOL, Coq, CSS, Dart, Dockerfile, F\*, Fortran, Go, Haskell, HTML, Idris, Isabelle, Isabelle ROOT, Java, JavaScript, JSON, Julia, Kotlin, LabVIEW, Lean, Literate Agda, Lua, Makefile, Maple, Markdown, Mathematica, MATLAB, Nix, NumPy, Objective-C, Objective-C++, Octave, Pascal, Pep8, Perl, PHP, Pickle, PowerShell, Python, Q\#, R, Ruby, Rust, SAS, Scala, Scilab, Shell, Swift, SystemVerilog, TeX, TypeScript, VBA, Verilog, VHDL, VisualBasic.\_NET, XML, XSLT, YAML.

\subsection{Code Quality Classifier}
\label{appendix:code_classifier}
The code quality classifier we run over the multi-programming language corpus supports the following programming languages: Assembly, C, C\#, C++, CSS, Dart, Go, HTML, Java, JavaScript, Kotlin, Lua, PHP, PowerShell, Python, Ruby, Rust, Shell, SQL. Then, these are the only programming languages included in the HQ code corpus.

\end{document}